\documentclass[final,5p,times,twocolumn,authoryear]{elsarticle}
\DeclareUnicodeCharacter{2212}{-}
\PassOptionsToPackage{numbers}{natbib}

\usepackage{amssymb}
\usepackage{lipsum} 
\usepackage{hyperref}
\usepackage{amsmath}
\usepackage{graphicx}
\usepackage{subcaption}
\usepackage{booktabs} 
\usepackage{array}
\usepackage{lscape}
\usepackage{siunitx}
\usepackage{upgreek}

\journal{arXiv preprint}

\setcitestyle{numbers}
\begin{document}

\begin{frontmatter}



\title{Traditional and Recent Approaches for Infrared Small Target Detection : Overview, Challenges and Perspectives (Part 2)}


\author[first]{Thierry Bouwmans}
\author[first]{Anastasia Zakharova}

\affiliation[first]{
    departement={Laboratoire MIA},
    organization={La Rochelle Université},
    city={La Rochelle},
    postcode={17000}, 
    country={France}
}

\author[second]{Belmar Garcia}
\affiliation[second]{
    departement={ISEN Toulon},
    city={Toulon},
    postcode={83000}, 
   country={France}
}

\author[third]{Marwa Chendeb El Rai}
\affiliation[third]{
    departement={Mathematics Division},
    organization={American University in Dubai},
    city={Dubai},
    country={United Arab Emirates}
}

\author[fourth]{Meghna Kapoor}
\affiliation[fourth]{
    departement={Laboratoire L3i},
    organization={La Rochelle Université},
    city={La Rochelle},
    postcode={17000}, 
    country={France}
}

\author[fifth]{Badri Narayan Subudhi}
\affiliation[fifth]{
    departement={Department of Electrical Engineering},
    organization={IIT Jammu},
    city={Jagti},
    postcode={181221}, 
    country={India}
}

\begin{abstract}
Since the rise of supervised deep learning models, researchers have attempted to solve infrared small target detection using various models like convolutional neural networks, GANs, Transformers, Mixture of Experts and recent graph/hypergraph neural networks.  Moreover, these deep neural network methods have surpassed the traditional unsupervised models and the low-rank and sparse decomposition unsupervised models reviewed in "Traditional and Recent Approaches for Infrared Small Target Detection : Overview, Challenges and Perspectives (Part 1)". In this context, the aim of this paper is to provide a complete survey of the deep learning models in infrared small target detection. First, we categorize the different approaches found in the literature in terms of their modeling strategies. Then, we discuss them in terms of the critical situations that they claim to handle. Furthermore, we present the specific performance evaluation protocols. Then, we conclude with several promising directions for future research.
\end{abstract}




\end{frontmatter}



\section{Introduction}
\label{introduction}
Infrared small target detection is a crucial task in military and civilian fields and deep neural networks supervised methods are suitable for this field. Indeed, model-based methods require the manual design of complex regularization constraints to extract prior features from infrared images. But real scenes present complex with dramatic changes target size, shape, and clutter background and it is difficult to use handcrafted features and fixed hyper-parameters to handle such variations. In contrast, data-based methods can automatically learn feature representations through training without the need for hand-crafted constraints.  Note that the reader can refer to the introduction of the paper titled "Traditional and Recent Approaches for Infrared Small Target Detection: Overview, Challenges and Perspectives (Part 1)" to have more details about the context of IRSTD. However, these recent supervised methods are based on various deep neural network methods and they can be either single-frame or multi-frame according to the employed architecture. Thus, we present the different DNNs models to highlight the main new advances in their applications in IRSTD:
\begin{itemize}
\item \textbf{Convolutional Neural Networks}: Convolutional Neural Networks are the most employed for both detection and segmentation methods. In this context, we categorize the various architectures, employed loss functions, attention and guidance mechanisms developed in literature to highlight the main trends. We follow the same scheme for wavelets based methods, Mamba-based state space model, transformers-based neural networks, adversarial networks and Mixture-of-Experts (MoE). \\
\item \textbf{Language Models} that are applied for IRSTD since 2025 with Large Language Models (LLMs) and Vision-Language Models (VLMs). These methods allow to employ textual information with image modalities but also require specific datasets. \\
\item \textbf{Graph/Hypergraph Learning Methods:} GNNs are  deep learning methods that use graph data as input. Graph Neural Networks is divided into two branches:  Spectral-based GNNs where graph spectral theory was applied to learn node representations, and spatial-based GNNs based on local-dependence assumption of graph structured data that graph convolution is defined in the spatial domain as aggregating and transforming local information. Thus, spectral-based Spatial-based GNNs describe the pairwise relations between objects. Transductive and inductive learning schemes are possible \cite{GM-Theory-3}. \\
\end{itemize}
The rest of this paper is organized as follows:  First, Section \ref{Preliminaries} presents an overview for novices about key features of deep neural networks such as architectures, optimization schemes, the notion of loss-function and the training. Section \ref{ConventionalDNNs} reviews the convolutional neural networks methods employed in IRSTD. Section \ref{Wavelets} survey Wavelet based methods.  Section \ref{Mamba} and Section \ref{Transformers} review Mamba-based State Space Models and Transformers based methods, respectively. Section \ref{AdversarialNetworks} presents GANs based methods. Section \ref{MoE} presents Mixture-of-Experts (MoE) based methods. 
Then, we focus on recent applications of learning models based on large languages models (LLMs) and Vision Language Models (VLMs) in Section  \ref{LanguageM} and graph neural networks in Section \ref{GraphLearning}.  In Section \ref{RDC}, we present specific performance evaluation for deep learning based methods. Finally, we provide a conclusion and perspectives for future research. \\

Table \ref{OverviewDL-1} and Table \ref{OverviewDL-2} show an overview of diverse deep learning methods employed in infrared small target detection. For the various mechanisms employed in CNNs architectures, overview tables are provided in their corresponding sub-section. Note that methods (six in number) that address dense infrared small target detection are labeled with "[Dense]" after their acronym either they can be considered as methods addressing sparse infrared small target detection. Methods (thirty in number) for moving infrared small target detection are labeled with "[MIRST]" after their acronym either they can be considered as methods addressing SIRST.

\section{Preliminaries}
\label{Preliminaries}
With the progress of GPU and the storage of Big Data, deep neural networks gains attention in the area of 2000's with new deep learning concepts such as Deep Belief Networks in 2006 and Generative Adversarial Networks (GANs) in 2014. Liu et al. classified the deep neural network architectures in the following categories: restricted Boltzmann machines (RBMs), deep belief networks (DBNs), autoencoders(AEs) network and deep Convolutional Neural Network (CNNs). More recently, more efficient architectures have been developed for language and vision tasks such as Large Language Models (LLMs), Vision Language Models (VLMs), Foundation Models, Mamba  and Transformers. Moreover, graph representation have been also introduced into deep neural networks providing Graph Convolutional Networks (GCNs) and Graph Neural Networks (GNNs).  \\

Thus, Deep Neural Networks are characterized by their architecture that is more and more sophisticated over time. An architecture comprises different layers categorized into input layer, hidden layer and output layer. Each layers contains a number of neurons that are activated or not following an activation function. This activation function can be considered as the mapping of the input to the output via a nonlinear transform function at each node. In literature, different activation functions can be found such as the sigmoid function \cite{MOD-Theory-100}, Rectified Linear Unit (ReLU) \cite{MOD-Theory-101}, and Parametric ReLU (PReLU) \cite{MOD-Theory-101}. Once the architecture is determined and the activation function is chosen, the DNN need to be trained using a large-scale dataset. To this end, the architecture is exposed to the training dataset to learn the weights of each neurons in each layer. The parameters are learned via a cost function that are minimized with a loss function on the desired output and the predicted one. The most common method for training is the back-propagation. Commonly, the gradient of the error function computed on the correct output and the predicted one is propagated back to the beginning of the network in order to update its parameters. For this, it requires a gradient descent algorithm. Batch normalization which normalizes mini-batches can also be employed to accelerate learning because it uses higher learning rates, and also regularizes the learning. For vocabulary, an epoch is a complete pass through a given dataset, and thus is the number of time where the neural network has been exposed to every record of the dataset once. An epoch is not an iteration which corresponds to one update of the neural net models parameters. Many iterations can occur before an epoch is over. Epoch and iteration are only identical if the parameters are updated once for each pass through the whole dataset. \\

In summary, the key characteristics of the deep neural networks are
the following ones:
\begin{itemize}
\item \textbf{Architecture:} The number,the size and the type of the layers are key characteristics of an architecture as well as the classes of functions that can be approximated by a feed-forward neural network. The key issue is how the selected architecture impact expressiveness.  
\item \textbf{Optimization:} It concerns the way to train the DNNs. This issue contains two aspects which are the datasets used for the training, and mostly the algorithm to optimize the network.
\item \textbf{Loss Functions:} Most the time, a conventional loss function is employed such as the $l_2$ loss function but several loss functions have been developed to be more adapted to the target applications.
\item \textbf{Training:} The dataset should be sufficiently large to learn all the kinds of scenes to avoid discrepancy in the performance when unseen scenes appear. Data augmentation can be used to solve this problem.
\end{itemize}
In addition, several attention and guidance mechanisms can be added according to the target application to improve the capture of diverse characteristics. Deep neural networks may also present a multi-branch structure. Modules can also be plug and play modules that can be employed in existing methods to improve their performance.

\section{Application of Deep Neural Networks in Infrared Small Target Detection} 
The adequacy of deep neural networks with the infrared small target detection relies on its specific constraints. For example, these constraints are not present in the original detection methods such as YOLO based ones, DETR based ones and SAM based ones. These constraints concern robustness to noise (i.e heteroscedastic noise \cite{DL-246}), the spatial and temporal constraints, the need of real-time implementations, and the ability to deal with the challenges met in infrared imaging sequences. In practice, researchers have proposed dedicated methods which attempt to take into account these specific constraints by acting on its key characteristics (Architectures, loss functions, attention and guidance mechanisms, etc.). In infrared small target detection, the following dedicated adaptations have been developed in the literature.

\subsection{Conceptual Neural Networks}
About kinds of deep neural networks, convolutional neural networks  can  extract local features of infrared small targets but they have shortcomings in capturing long-distance global features between the target and its background.  CNNs expand the semantic receptive field through stacked convolutional layers to extract features, but they face difficulties to capture global context and low level detail features—both crucial for small target detection \cite{DL-95}. In summary, CNNs have the following limitations for infrared small target detection as indicated in the work of Li et al. \cite{DL-577}: 
\begin{itemize}
\item \textbf{Target Loss Risk:} CNNs typically employ successive downsampling operations. For small infrared targets occupying only a handful of pixels, this process easily cause target features to degrade or entirely disappear in deeper network layers.
\item \textbf{Lack of Long Range Dependencies:} The restricted receptive fields related to standard convolutions prevent the model from capturing global spatial relationships across the broader image context. This means the network faces difficulties to distinguish true targets from background noise with similar shapes unless it can reference global contextual information.
\end{itemize}
To address these limitations, some researchers employ variants of CNNs such as Pinwheel Convolution Network (PCNs) \cite{DL-104,DL-597} and Chinese Knot Convolutional Network (CKCNs )\cite{DL-572} or vision transformers to model global dependencies. Indeed, transformer-based methods can distinguish true targets from similar background noise by analyzing the background continuity across the entire image. The vision transformer (ViT) model can capture long-distance dependent and global features \cite{DL-94}. Even if transformers excel at capturing long range context through self-attention mechanisms, they still have the following limitations as indicated in the work of Li et al. \cite{DL-577}:
\begin{itemize}
\item \textbf{Low Pass Filtering Effect:} Self-attention inherently generates a low-pass filtering effect, tending to smooth feature maps. However, infrared small targets typically manifest as high frequency pulse signals and are easily suppressed as noise by the global smoothing mechanism.
\item \textbf{High Computational Complexity:} The computational complexity of transformers increases quadratically with image resolution. This causes a significant burden for small object detection tasks requiring high resolution feature retention.
 \end{itemize}
Thus, Transformer-based structures are insufficient in local feature extraction and are computationally expensive \cite{DL-263} with quadratic computational complexity. The reader can find in the work of Choi et al. \cite{DL-599-217} a comparative performance of CNN and Transformer in infrared small target detection. Nevertheless, Swin Transformers \cite{DL-239,DL-279,DL-350} appear to be more suitable for infrared small target detection. In this context, several researchers \cite{DL-94,DL-278,DL-590,DL-599-151,DL-599-240,DL-599-278,DL-599-316,DL-599-369,DL-599-401} provide combination of CNNs and ViT to compensate the shortcomings of CNNs in modeling long-distance global features. Nevertheless, these methods still have the following issues \cite{DL-95}: 
\begin{itemize}
\item \textbf{Detail vs. Noise Balance:} To address background noise and
expand semantic range in infrared images, continuous down-sampling
is often used. But this process can lead to significant
loss of target details due to their small and faint nature.
\item \textbf{Complexity vs. Locality Balance:} The self-attention mechanism in ViTs requires quadratic complexity leading to low computational efficiency.
\end{itemize}
However, different variations of the Mamba such as the integration of the 2D-Selective-Scan (SS2D) technique, appears as an alternative to CNN– and Transformer-based methods \cite{DL-263}. Several researchers also employ Kolmogorov–Arnold Networks (KAN) \cite{DL-121,DL-164,DL-343,DL-599-173,DL-599-210} because it explicitly models nonlinear mapping curves. This flexible function modeling mechanism is well suited to the localized and nonlinear saliency patterns of infrared small targets. Finally, several research works on the feature domain \cite{DL-298,DL-531,DL-599-13}.

\subsection{Architectures}
Several deep neural architectures  have been developed for image tasks such as AlexNet \cite{MOD-Theory-200} for image classification, VGG-Net \cite{MOD-Theory-201} for large-scale image recognition, U-Net  \cite{MOD-Theory-202} for image segmentation, GoogLeNet \cite{MOD-Theory-203} for computer vision, and Microsoft Residual Network (ResNet) \cite{MOD-Theory-204} for image recognition. Nevertheless, none of these architectures have been designed for infrared imaging. Nevertheless, several architectures have been used infrared small target detection namely U-Net \cite{DL-10,DL-13,DL-92,DL-41}, Siamese U-Net \cite{DL-139}, ifUNet++ \cite{DL-599-215}, U2-Net \cite{DL-47}, U-shaped \cite{DL-45,DL-542,DL-230,DL-161,DL-599-34}, Deformable Convolutional UNet \cite{DL-599-134}, ResNet \cite{DL-518,DL-373} and Res2Net \cite{DL-599-143}. \\

\begin{figure*}
\begin{center}
\includegraphics[width=15cm]{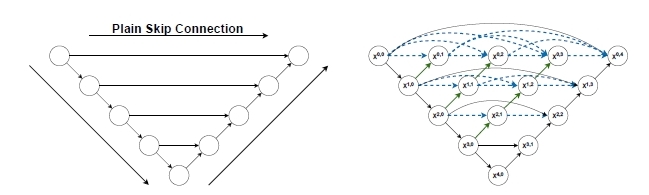} \\
\includegraphics[width=15cm]{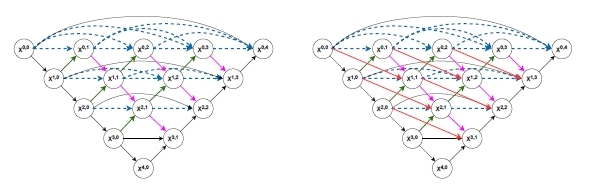} 
\caption{U-shape structures and a multi-scale nested interaction module. First row: UNet \cite{MOD-Theory-202} and UNet++ \cite{MOD-Theory-202-1} designed for biomedical images. Second row: DNA-Net \cite{DL-12} and iSmallNet \cite{DL-599-68} designed for infrared small target detection (Image from Hu et al. \cite{DL-599-68}).} 
\label{UNetAchitectures}
\end{center}
\end{figure*}

Even if U-shaped architectures have enhanced baseline performance, existing methods still suffer from two limitations: insufficient spatial perception for tiny targets leads to target location loss and edge degradation and semantic ambiguity in deep feature reconstruction. To address these limitations, several researchers \cite{DL-12,DL-13,DL-43,DL-61,DL-98,DL-221,DL-599-68} have developed dense nested attention architecture. For example, Figure \ref{UNetAchitectures} shows three U-shape structures and a multi-scale nested interaction module. The basic U-shaped structure consists of an encoder, a decoder, and plain skip connections. The encoder allows to expand the receptive area and extract useful information. The decoder helps to recover features to the same size as the input. The plain skip connection acts as a bridge to transfer these low-level and high-level features from the encoder to the decoder sub-networks. Nested and dense structures are represented for UNet++ \cite{MOD-Theory-202-1},  DNA-Net \cite{DL-12} Nested U-Structure \cite{DL-599-445} and iSmallNet \cite{DL-599-68}. Due to the optimal receptive domains of targets with sizes varying greatly, the U-shaped sub-networks with different depths are suitable for targets of different sizes. Indeed, each node can receive the characteristics of its own layer and adjacent layers resulting in repeated multi-layer feature fusion. Thus, the representation of small targets is kept in deep layers so that better results can be reached. 

\subsection{Learning/Training}
Deep neural networks needs to be trained on datasets requiring large amounts of costly and impractical annotated data. In addition, deep learning methods rely on supervised learning.  To address these deployable limitations, deep neural networks models for infrared small target detection attract investigation of few-shot learning mechanisms \cite{DL-599-106}, and zero-annotation strategies \cite{DL-599-119}. In the case of small scale datasets, data augmentation strategies have been developed using Random-Cloning models \cite{DL-67}, Random-Erasing models \cite{DL-397}, prior-guided models \cite{DL-599-136}, thermodynamics-Inspired models \cite{DL-599-140}, GANs \cite{DL-599-105,DL-599-137}, diffusion models
\cite{DL-235,DL-599-174} and hybrid models \cite{DL-599-130} . However, even if the training is achieved on large-scale datasets, the performance of deep neural networks methods decrease in presence of unseen scenes requiring generalization investigation \cite{DL-599-104}.

\subsection{Loss Functions}
For infrared small target detection, the most common loss function is the Soft-IoU loss function. Nevertheless, the Soft-IoU loss does not consider the difference in importance among individual pixels. It assigns the same weight to all pixels, which may lead to insufficient attention to critical parts of the target (such as the centroid) during training. Thus, it impacts the prediction accuracy of the central region of the target. Several researchers have developed dedicated loss functions to overcome this limitation. The three main alternatives are the weighted Mean Squared Error (MSE) \cite{DL-63} loss function, the Binary Cross-Entropy (BCE) loss, and the Dice loss function, defined respectively as:
\begin{itemize}
\item \textbf{Weighted Mean Squared Error (MSE) \cite{DL-63}:} 
Penalizes prediction errors with spatial weights that emphasize target pixels:
\begin{equation}
  \mathcal{L}_{\mathrm{MSE}} = \frac{1}{HW}\sum_{i,j} 
  w_{ij}(p_{ij} - g_{ij})^2,
\end{equation}
where $H$ and $W$ are the image height and width, $p_{ij} \in [0,1]$ 
is the predicted probability, $g_{ij} \in \{0,1\}$ the ground-truth mask at pixel $(i,j)$, and $w_{ij} \geq 0$ a spatial weight map emphasizing target pixels.
\item \textbf{Binary Cross-Entropy (BCE) \cite{DL-211}:} 
Measures pixel-wise classification error:
\begin{equation}
  \mathcal{L}_{\mathrm{BCE}} = -\frac{1}{HW}\sum_{i,j}
  \bigl[g_{ij}\log p_{ij} + (1-g_{ij})\log(1-p_{ij})\bigr],
\end{equation}
where $H$, $W$, $p_{ij}$, and $g_{ij}$ are defined above.
\item \textbf{Dice loss \cite{DL-41}:} Measures the overlap between predicted and ground-truth masks:
\begin{equation}
  \mathcal{L}_{\mathrm{Dice}} = 1 - 
  \frac{2\langle\mathbf{p},\mathbf{g}\rangle + \epsilon}
  {\|\mathbf{p}\|_1 + \|\mathbf{g}\|_1 + \epsilon},
\end{equation}
where $\mathbf{p}$ and $\mathbf{g}$ are the vectorized predicted and 
ground-truth masks, $\langle\mathbf{p},\mathbf{g}\rangle = 
\sum_{i,j}p_{ij}g_{ij}$ denotes their inner product, $\|\cdot\|_1$ 
denotes the element-wise sum, and $\epsilon > 0$ is a smoothing 
constant to avoid division by zero.
\end{itemize}

More elaborated loss functions have been recently developed such as the following ones: complete IoU (CIoU)  loss function \cite{DL-236}, adaptive loss function \cite{DL-599}, fuzzy perceptual loss function \cite{DL-269},  multi-task weighted hybrid loss function \cite{DL-96}, multi-stage loss  function \cite{DL-520}, multilevel cross-entropy loss function \cite{DL-201}, weighted-sigmoid-cross-entropy loss function \cite{DL-16}, mask-weighted heatmap loss \cite{DL-505}, DpT-k loss  function \cite{DL-97}, boundary difference loss function \cite{DL-91}), scale-based dynamic loss \cite{DL-104}, self-learning joint loss function \cite{DL-372}, self-supervised loss \cite{DL-246}, Physics-Constrained Loss (PCL) function \cite{DL-204}, Focal Loss function \cite{DL-75}, Adaptive Threshold Focal Loss (ATFL) function \cite{DL-568}, InfoNCE loss function \cite{DL-570}, Temperature Aware InfoNCE loss function (TAC) \cite{DL-569}, Boundary-aware Contrastive (BC) loss function \cite{DL-599-70}, basket-based hierarchical temporal consistency loss function \cite{DL-599-102}, Scale and Location Sensitive (SLS) loss function \cite{DL-41}, SCR-Guided Loss Reweighting function \cite{DL-599-176},  Smooth Wasserstein Loss (SWL) function \cite{DL-599-135},  Distribution-aware Pixel-wise Wasserstein Loss (DPWL) \cite{DL-564}, Structure–Confidence Hybrid (SCH) loss function \cite{DL-599-193}, SSQM loss function \cite{DL-599-406} and Hyperbolic Loss Function (HLF) \cite{DL-599-251} to be exhaustive as possible. The most elaborated loss function is the task-driven joint loss function \cite{DL-250} which combines intensity loss, Haar wavelet mask loss, and background residual loss.

\subsection{Attention and Guidance Mechanism}
Several attention and guidance mechanisms  are added to improve the capture of the various characteristics such as the local/global features, the semantic information, the motion information and the saliency information to cite a view. Most of the IRST methods based deep learning consist of three specific modules either in detection based methods \cite{Applications-S-33} and segmentation based methods \cite{DL-230,DL-242,DL-145,DL-165,DL-341,DL-134,DL-341,GM-2}. Ablation studies enable to assess the role and the importance of each module about the challenges they aim to handle. \\

However, several IRST methods based deep learning propose plug and play modules (RRCA-Net \cite{DL-97}, SDG-ResNet \cite{DL-501}, LLCEFA \cite{DL-190}, DGGENet \cite{DL-110}) that can be employed to consistent performance improvement in existing IRSTD methods. Plug and play modules are often available to provide a different learning mechanisms (PAL \cite{DL-368}, LESPS \cite{DL-364}, LSP \cite{DL-599-3}, DtoD \cite{DL-599-6}, TGPL \cite{DL-599-75}).

\subsection{Miscellaneous Mechanisms}
Several IRST methods based deep learning consist of multi-branch as dual branch mechanisms \cite{DL-92,DL-114,DL-233,DL-63,DL-165,DL-168,DL-210,DL-330,DL-396,DL-190} and triple branch mechanisms.\cite{DL-210}.  In addition, several researchers employ various synergistic frameworks \cite{DL-190,DL-201,DL-217,DL-233,DL-243,DL-296,DL-360,DL-550,DL-599-20}. Other researchers developed refinement mechanisms (OSCAR \cite{Dataset-200}, SAFARI-net \cite{DL-191}, CPMamba  \cite{DL-212}, AER-net \cite{DL-542}, MAPPR++-Net \cite{DL-599-44}).

\subsection{Computational Cost}
Deep neural networks require extensive computational cost. In this context, network quantization allows to reduce the model size, memory footprint, and computational cost. It is achieved by representing float weights and activations with lower bit counterparts, which is essential for model deployment on resource-limited devices. In 2024, Li et al. \cite{DL-599-115} study the quantization sensitivity of small target segmentation network by looking at the sensitivity heterogeneity of different layers in the network. The analysis reveals that feature maps in shallow layers and encoder subnetwork are more vulnerable to information loss due to quantization as compared to deep layers and decoder subnetwork. Thus, Li et al.  \cite{DL-599-115} assign a different bit width for each block according to their quantization sensitivity  and propose a Symmetrically Progressive decreasing Mixed-precision Quantization (SPMix-Q\footnote{https://github.com/YeRen123455/SIRST-Quantization-Deployment})  to achieve high-performance segmentation under low-bit quantization (i.e., 2.42 bits for weights and 3.82 bits for activations). Experiments show that SPMix-Q provides similar detection with only 1/13 model size, 1/4.6 memory footprint, and 1/29 computational cost to the full-precision counterparts. \\

\begin{table*}[!ht]
\begin{center}
\scalebox{0.79}{
\begin{tabular}{|l|l|}
\hline
\scriptsize{Deep Neural Networks Methods} &\scriptsize{Acronyms (Authors)}  \\
\hline
\hline
\scriptsize{Detection based Algorithms} &\scriptsize{CNN (YOLOv5) [Detection] (Zhou et al. \cite{DL-6}), TCRNet (YOLOv3) [Detection] (Mc Intosh et al. \cite{DL-7}) } \\
\scriptsize{(19 papers)} &\scriptsize{\textbf{YOLO series:}  IRSTD-YOLO (Tang et al. \cite{DL-541}), YOLOSR-IST (Li and Shen \cite{DL-700}), YOLO-MBL (Shen et al. \cite{DL-701})} \\
\scriptsize{} &\scriptsize{YOLO-TFFM (Xu et al. \cite{DL-702}), SAYOLO (Guo et al. \cite{DL-703}), BOH-YOLO (Zhu and Zhou \cite{DL-704})}\\
\scriptsize{} &\scriptsize{YOLO-LCF (Chen et al. \cite{Miscellaneous-100}), IR-WSANet (Jin et al. \cite{DL-113}), HSFL (Wang et al. \cite{DL-599-203})}\\
\scriptsize{} &\scriptsize{YOLO-MSF (He et al. \cite{DL-599-400})}\\
\scriptsize{} &\scriptsize{\textbf{DETR series:} RT-DETR-CST (Duan et al. \cite{DL-118}), SEF-DETR (Liu et al. \cite{DL-145}), IRST-RTDETR (Chen et al. \cite{DL-160})}\\
\scriptsize{} &\scriptsize{REDETR-ISTD (Li and Wei \cite{DL-540}), MI-DETR (Liu et al. \cite{DL-169}), CSFPR-RTDETR-CR (Wang and Sun \cite{DL-710})}\\
\scriptsize{} &\scriptsize{DCFD-DET (Liu et al. \cite{DL-713}), GHRT-DETR (Sun et al. \cite{DL-714}),  FGOM-RTDETR (Wang et al.\cite{Applications-S-33})}\\
\scriptsize{} &\scriptsize{KE-RT-DETR (Ji et al.\cite{DL-599-260}), SPSM-DETR (Cui et al.\cite{DL-599-434}) }\\
\hline
\scriptsize{Segmentation based Algorithms} &\scriptsize{SSTNet (ConvLSTM) (Chen et al. \cite{DL-38})} \\
\scriptsize{(29 papers)} &\scriptsize{STNet (U2-Net) (Kumar et al. \cite{DL-47}), PFFT (Zhu  et al. \cite{DL-90}) } \\
\scriptsize{} &\scriptsize{ID-UNet (Chen  et al. \cite{DL-98}), IR-SDTNet (Sun et al. \cite{DL-130}) } \\
\scriptsize{} &\scriptsize{KPF-Net (Wu  et al. \cite{DL-121}), SP-KAN (Yuan et al. \cite{DL-164}), GLKA-UNet (Zhang  et al. \cite{DL-599-173})} \\
\scriptsize{} &\scriptsize{ADSUNet (Siamese U-Net) (Zhang et al. \cite{DL-139}), UIU-Net [Single-Frame] (Wu et al. \cite{DL-220}), CN-UNet (Li et al. \cite{DL-293}), CI-UIUNet (Wang et al. \cite{DL-599-362})} \\
\scriptsize{} &\scriptsize{DHiF (Li et al. \cite{DL-301}), ISTDU-Net (Hou et al. \cite{DL-506}), RepISD-Net (Wu et al. \cite{DL-571})} \\
\scriptsize{} &\scriptsize{LASNet (Chen et al. \cite{Dataset-300}), FCNet [Ship] (Guo et al. \cite{Applications-S-45}), CKConv (Lin et al. \cite{DL-572})} \\
\scriptsize{} &\scriptsize{STCEUNet (Expanded U-Net) (Deng et al. \cite{DL-599-76}), SPMix-Q (ResUnet) (Li et al. \cite{DL-599-115})} \\
\scriptsize{} &\scriptsize{DCUNet (Huang et al.  \cite{DL-599-134}), PCLC-Net (Xu et al.  \cite{DL-599-168})} \\
\scriptsize{} &\scriptsize{Shape-biased: CESBR (ResNet) (Lin et al. \cite{DL-373}, SBR (ResNet) (Lin et al. \cite{DL-518})}\\
\scriptsize{} &\scriptsize{SFC-INRNet (Xu et al. \cite{DL-599-5}), ISGLNet (Zhang et al. \cite{DL-599-34})}\\
\scriptsize{} &\scriptsize{Heterogeneous Parallel Networks: HPN-SOE (Liu et al. \cite{DL-599-62})} \\
\scriptsize{} &\scriptsize{Pseudo Time Surface: Network: PTS-Net (Wang et al. \cite{DL-599-394})} \\
\scriptsize{} &\scriptsize{\textbf{SAM series:} IRSAM (Zhang et al. \cite{DL-599-12}), SAM with IR Adapters \cite{DL-510}, SAM-SL (Fu et al. \cite{DL-599-16})} \\
\scriptsize{} &\scriptsize{E2SAM (Li et al. \cite{DL-599-30})} \\
\scriptsize{} &\scriptsize{Scalpel-SAM (Liu et al. \cite{DL-599-31}), SimIR  (Zhang et al. \cite{DL-599-11}), LSP-ST (Zhang et al. \cite{DL-599-33})}\\
\scriptsize{} &\scriptsize{TEP-SAM (Xing et al. \cite{DL-720}), FWSAM-Net (Chen et al. \cite{DL-599-373})}\\
\scriptsize{} &\scriptsize{ASG-SAM (Chen et al. \cite{DL-599-410}, PhyIRSTD (Jiang et al. \cite{DL-721})}\\
\scriptsize{} &\scriptsize{\textbf{Related STMD series (visible spectrum, not designed for IRSTD):} ESTMD (Wiederman et al. \cite{DL-599-169-1}), DSTMD (Wang et al. \cite{DL-599-169-2})}\\
\scriptsize{} &\scriptsize{Feedback STMD (Wang et al. \cite{DL-599-169-3}, $\mathcal{F}$-STMD (Jing et al. \cite{DL-599-169-5}, ST-STMD (Wang et al. \cite{DL-599-169-4})} \\
\scriptsize{} &\scriptsize{Frac-STMD (Xu et al. \cite{DL-599-169-6}), $\mathcal{RT}$-STMD (Ling et al. \cite{DL-599-169}), $\mathcal{T}$-STMD (Jing et al. \cite{DL-599-169-10})} \\
\scriptsize{} &\scriptsize{$\mathcal{RF}$-STMD (Ling et al. \cite{DL-599-169-11}), sSTMD (Chen et al. \cite{DL-599-169-12}) }\\
\hline
\end{tabular}}
\caption{Deep Neural Networks Supervised Detection and Segmentation-based Methods in Infrared Small Target Detection: An Overview. For each category, only the key references are indicated. Please see the corresponding section for more references.}
\label{OverviewDL-1}
\end{center}
\end{table*}

\begin{table*}[!ht]
\begin{center}
\scalebox{0.79}{
\begin{tabular}{|l|l|}
\hline
\scriptsize{Deep Neural Networks Methods} &\scriptsize{Acronyms (Authors)}  \\
\hline
\hline
\scriptsize{Learning Mechanisms} &\scriptsize{Conventional Learning Mechanisms} \\
\scriptsize{}  &\scriptsize{Single point supervised learning \cite{DL-363,DL-364,DL-367,DL-368}, Weakly supervised learning  \cite{DL-365,DL-366,DL-599-132}, Source-free domain adaptation learning \cite{DL-390,DL-393,DL-599-232}} \\ 
\scriptsize{}  &\scriptsize{Contrastive Learning \cite{DL-569,DL-599-319}, Dual-Stream Edge-Target Learning \cite{DL-532}}\\
\scriptsize{}  &\scriptsize{Data Augmentation \cite{DL-67,DL-317,DL-397,DL-235,DL-599-130,DL-599-136,DL-599-137}
}\\
\hline 
\scriptsize{Attention Mechanisms} &\scriptsize{Conventional Attention Mechanisms} \\
\scriptsize{} &\scriptsize{Pyramid Attention Mechanisms, Hierarchical Attention Mechanisms \cite{DL-102}, Spatial Attention Mechanisms \cite{DL-32,DL-535}} \\
\scriptsize{} &\scriptsize{Direction-Induced Attention Mechanisms \cite{DL-33,DL-34}, Contextual Attention Mechanisms \cite{DL-10,DL-13,DL-599-272}} \\
\scriptsize{} &\scriptsize{Cellular Interactive Attention  \cite{DL-517}, Bidirectional Interactive Attention \cite{DL-531}} \\
\hline
\scriptsize{Adaptive and Multi-Cue Mechanisms} &\scriptsize{Adaptive Mechanisms \cite{DL-299}, Recurrent Mechanisms \cite{DL-97}, Diffusion Mechanisms \cite{DL-147,DL-162,DL-234}} \\
\hline
\scriptsize{Guidance Mechanisms} &\scriptsize{Prior Guidance \cite{DL-331}), Features Guidance \cite{DL-335}, Semantic Guidance \cite{DL-96,DL-134}, Motion Guidance \cite{DL-169,DL-599-35,DL-599-36}, Saliency Guidance \cite{DL-370,DL-599-32}, Physics-Informed Guidance \cite{DL-204,DL-599-295}} \\
\scriptsize{} &\scriptsize{Memory Guidance \cite{DL-511}, Feedback Guidance \cite{DL-142,DL-146}, Geometry Guidance \cite{DL-170}, Ordinary Differential Equation Guidance \cite{DL-290} , Fuzzy-Logic Guidance \cite{DL-396}, Reconstruction Guidance \cite{DL-599-179}} \\
\hline 
\scriptsize{Robust and Lightweight Algorithms} &\scriptsize{Robust Algorithms \cite{DL-3,Dataset-700}, Sparsity Algorithms \cite{DL-60,DL-513}, Aliasing Rectification Algorithms \cite{DL-238,DL-369}, Low Illumination Algorithms \cite{DL-16,DL-103,DL-103-1}} \\
\scriptsize{} &\scriptsize{Lightweight Algorithms \cite{DL-1,DL-2,DL-70}}\\
\hline
\scriptsize{Wavelets} &\scriptsize{} \\
\hline
\scriptsize{Architectures} &\scriptsize{Mamba Architecture \cite{DL-275}, Transformers based Neural Networks \cite{DL-18,DL-19,DL-20}} \\
\hline
\hline
\scriptsize{Generative Adversarial Networks} &\scriptsize{GAN (UNet) (Zhao et al. \cite{DL-4}), DAENet (Jin et al. \cite{DL-333}), CDFGAN  (Zhou et al. \cite{DL-514}) } \\
\hline
\scriptsize{Diffusion Models}  &\scriptsize{IRSTD-Diff (Li et al. \cite{DL-599-101}), IRDiff (Zhao et al. \cite{DL-599-172})  } \\
\hline
\scriptsize{Context Adversarial Networks} &\scriptsize{MDvsFA-cGAN [Single-Frame] (Wang et al. \cite{DL-15}), Two-Stage U-Skip Net (Wang et al. \cite{DL-17}), AGPCNet [Single-Frame] (Zhang et al. \cite{DL-21}) } \\
\hline
\hline
\scriptsize{Mixture-of-Experts} &\scriptsize{CoMoE [MIRST] (Duan et al. \cite{DL-196}), FAME-Net (Du et al. \cite{DL-599-70}), HyTBE (Li et al. \cite{DL-599-313})} \\
\scriptsize{} &\scriptsize{TBD-Net (Gong et al. \cite{DL-599-430})} \\
\hline
\hline
\scriptsize{Large Language Models} &\scriptsize{SAIST [Single-Frame] (Zhang et al. \cite{LLM-1}), MIRSAM (Zhang et al. \cite{LLM-2}), Text-IRSTD (Huang et al. \cite{LLM-3}), TIPP [MIRST] (Zhou et al. \cite{LLM-5})} \\
\scriptsize{} &\scriptsize{DGNet (Yu et al. \cite{LLM-6}), ADGNet (Wang et al. \cite{LLM-7})} \\
\hline 
\scriptsize{Vision-Language Models} &\scriptsize{MoPKL [Multi-Frame][MIRST] (Shen et al. \cite{VLM-4}), iMoPKL [Multi-Frame][MIRST] (Shen et al. \cite{VLM-5}), SeViL [Multi-Frame][MIRST] (Duan et al. \cite{VLM-1}), (Yu et al. \cite{VLM-2}), SPIRIT (Xu et al. \cite{VLM-3})} \\
\scriptsize{} &\scriptsize{GIRVFM ( Ma et al. \cite{VLM-7}),FD-LVT (Jiang \cite{VLM-10})} \\
\hline
\hline
\scriptsize{Graph/Hypergraph Learning Models}  &\scriptsize{Graph: CS-ViG-UNet (Lin et al. \cite{DL-291}), IRST-CGSeg  (Jia et al. \cite{PSTD-42}), GCLNet (Shen et al. \cite{GM-1})} \\
\scriptsize{} &\scriptsize{LPA-NBM (Bampton et al. \cite{GM-3}), GEP-UNet (Yang and Cui \cite{GM-4}), IRGraphSeg (Jia et al. \cite{DL-599-95})} \\
\scriptsize{} &\scriptsize{SGINet (Deng and Xiao \cite{GM-6})} \\
\scriptsize{} &\scriptsize{Hypergraph: HyperTea  (Qi et al. \cite{GM-2}), (Song et al. \cite{GM-5})} \\
\hline
\end{tabular}}
\caption{Deep Neural Networks Supervised Segmentation-based Methods in Infrared Small Target Detection: An Overview. For each category, only the key references are indicated. Please see the corresponding section for more references.}
\label{OverviewDL-2}
\end{center}
\end{table*}

\section{Conventional Deep Neural Networks} 
\label{ConventionalDNNs}
Deep Neural Networks-based methods for infrared small target detection can be divided into two categories namely detection based methods and segmentation based methods \cite{DL-12}.

\subsection{Detection based Methods} 
In 2018, Liu et al. \cite{DL-0}  first design a generic target detection
framework for infrared small target detection by using a
a Multi-Layer Perception (MLP) network with $5$ layers for infrared small target detection. In 2020, McIntosh et al. \cite{DL-7} fine-tuned several  generic target detection network and employ  the optimized eigen-vectors as input to achieve improved performance. In general, the detection based methods can be classified into two categories: traditional general detectors (Faster R-CNN, SSD, YOLO series) and Transformer-based detectors (RT-DETR series \cite{MOD-Theory-206}). About the infrared small target-specific detectors, they fall in the same two categories as follows:
\begin{itemize}
\item YOLO series such as IRSTD-YOLO \cite{DL-541}, YOLOSR-IST \cite{DL-700}, YOLO-MBL \cite{DL-701}, YOLO-TFFM \cite{DL-702}, SAYOLO \cite{DL-703}, BOH-YOLO \cite{DL-704}, YOLO-LCF \cite{Miscellaneous-100}  and IR-WSANet \cite{DL-113}.
\item DETR series such as RT-DETR-CST \cite{DL-118}, SEF-DETR \cite{DL-145}), IRST-RTDETR \cite{DL-160}, REDETR-ISTD \cite{DL-540}, MI-DETR \cite{DL-169}, CSFPR-RTDETR-CR \cite{DL-710}, DCFD-DET \cite{DL-713}, GHRT-DETR \cite{DL-714} and FGOM-RTDETR \cite{Applications-S-33} to cite the more representative algorithms.
\end{itemize}

\subsection{Segmentation based Methods} 
Segmentation-based methods have attracted increasing attention because these methods can produce both pixel-level classification and localization outputs. 

\begin{itemize}
 \item The first category concerns dedicated methods with diverse strategies employing various mechanisms such as attention mechanisms (context \cite{DL-10,DL-13}, spatial \cite{DL-32}, direction-induced \cite{DL-33, DL-34, DL-211}), diffusion mechanisms \cite{DL-147,DL-234}, learning mechanisms \cite{DL-115,DL-117}, feedback mechanisms \cite{DL-142,DL-146}, multi-branch mechanisms\cite{DL-92,DL-114,DL-243}, multi-scale mechanisms \cite{DL-41,DL-63,DL-100,DL-577}, motion mechanisms \cite{DL-169,DL-199,DL-271}, semantic mechanisms \cite{DL-96,DL-232,DL-134}, feature guidance \cite{DL-37,DL-44,DL-61} and ODE mechanisms  \cite{DL-272,DL-290} to cite a few. Moreover, different architectures are employed based on Mamba \cite{DL-75, DL-268,DL-269}, on Transformers \cite{DL-18,DL-19,DL-20} and on wavelets \cite{DL-5,DL-205}. 
\item The second category of segmentation based methods are in the Segment Anything Model (SAM) series.(IRSAM \cite{DL-599-12}, SAM with IR Adapters \cite{DL-510}, SAM-SL \cite{DL-599-16}, E2SAM \cite{DL-599-30}, SimIR \cite{DL-599-11}, LSP-ST  \cite{DL-599-33}, SAIST \cite{LLM-1}, MIRSAM \cite{LLM-2}). SAM is a general object segmentation model designed by the Meta AI team. SAM provides strong generalization and versatility using training in large-scale, diverse datasets. Its core architecture comprises an image encoder, a prompt encoder, and a mask decoder. The image encoder employs Vision Transformer (ViT) to extract global features, the prompt encoder processes various forms of user prompts (such as points, boxes, or masks), and the mask decoder provides segmentation results using these prompts. Its architectural design allows real-time responsiveness to dynamically adjust input prompts and generate segmentation masks in real time. Nevertheless, their effectiveness on infrared small target detection remains limited \cite{DL-599-12} because of the significant domain discrepancy between pretraining datasets and infrared imagery, which typically exhibits strong noise, weak texture, low-contrast targets, and a severe imbalance between foreground and background information. IRSAM \cite{DL-599-12} and SAM-SL \cite{DL-599-16} typically employ handcrafted priors during fine-tuning while SAIST \cite{LLM-1} incorporate scene-level priors from language models to improve infrared small target detection. But, handcrafted priors often suffer from poor generalizability and may hinder the model’s end-to-end
learning capacity \cite{DL-518}.  To achieve adaptation for infrared small target
detection without compromising generalization and scalability, LSP-ST \cite{DL-599-33} addresses the limited adaptability of vision foundation models in infrared small target detection using a shape-aware inductive bias that guides adaptation beyond texture.
\end{itemize}

Note that methods that are detection based methods (nineteen in number) are labeled with "[Detection]" after their acronym in the overview tables either they can be considered as segmentation based methods.

\subsection{Dedicated Architectures}
This section reviews the designed architectures employed for infrared small target detection. In 2022, Hou et al. \cite{DL-506} design an infrared small target detection based on the U-Net architecture (ISTDU-Net). ISTDU-Net is a deep learning network based on a U-shaped structure. It converts a single frame infrared image into a target probability likelihood map of image pixels. ISTDU-Net employs feature map groups in network down-sampling and uses a fully connected layer in jump connection to avoid a large number of backgrounds with similar structures from the global receptive fields Experiments show that ISTDU-Net detects small infrared targets in complex backgrounds. ISTDU-Net provides better receiver operating characteristic (ROC) curve with a low false alarm rate and with the area under curve (AUC) value is 0.9977. \\

In 2023, \textbf{in the context of single-frame detection}, Wu et al. \cite{DL-220} design a  “U-Net in U-Net” (UIU-Net\footnote{\url{https://github.com/danfenghong/IEEE_TIP_UIU-Net}}). UIU-Net embeds a tiny U-Net into a larger U-Net backbone allowing the multi-level and multi-scale representation learning of objects. UIU-Net is trained from scratch, and the learned features enhance global and local contrast information. The UIU-Net model consists of two modules.  First, the resolution-maintenance deep supervision (RM-DS) integrates Residual U-blocks into a deep supervision network to provide deep multi-scale resolution-maintenance features while learning global context information. Second, the interactive-cross attention (IC-A) module encodes the local context information between the low-level details and high-level semantic features. Experiments on the SIRST and Synthetic datasets show that UIU-Net offers powerful generalization performance. Wu et al. \cite{DL-571} propose a learning efficient network via structural Re-Parameterization (RepISD-Net\footnote{https://github.com/dalinlin-Wu/RepISD-Net}) which uses different network architectures but equivalent model parameters for training and inference, respectively. In the training phase, a parallel multibranch Edge Compensation Block (ECB) captures the local salient features and finer contour characteristic. In the inference phase, a multi-branch topology structures are merged into a single branch with only cascaded $3 \times 3$ convolutions for fast inference. Experiments show that RepISD-Net offers similar orebetter detection performance with significant acceleration in inference speed as compared with previous SOTA methods. \\ 

In 2024, Lin et al. \cite{DL-518} develop a unified framework based on a ResNet architecture to learn Shape-Biased Representations (SBR) by incorporating shape information into model learning. SBR uses a large-kernel encoder and a shape-guided decoder to learn discriminative shape-biased representations in an end-to-end manner. The large-kernel encoder describes infrared images into shape-preserving representations by using a few convolutions whose kernel size is as large as $9 \times 9$ instead of $3 \times 3$. The shape-guided decoder simultaneously achieves two tasks. First, it decodes the encoder representations through upsampling reconstruction to reconstruct the segmentation. Second, it hierarchically fuses the decoder representations and edge information through cascaded gated ResNet blocks to reconstruct the contour. Experiments show that SBR outperforms 18 previous SOTA methods. In a further approach, Lin et al. \cite{DL-373} improve SBR by designing a learning Contrast-Enhanced Shape-Biased Representations (CESBR) based on a ResNet architecture. CESBR employs a contrast-shape encoder and a shape-reconstructable decoder to learn discriminative representations that allow to identify target objects. The contrast-shape encoder uses a stem of central difference convolutions. Thus, few large-kernel convolutions allow to extract shape-preserving features. Second, a shape-reconstructable decoder captures the edge map  and is learned by simultaneously optimizing two shape-related consistencies: the internal one decodes the encoder representations by upsampling reconstruction and constraints segmentation consistency. The encoder and decoder are learned in an end-to-end manner, and the resulting shape-biased encoder representations are suitable for identifying infrared small targets. Experiments show that CESBR outperforms previous SOTA algorithms. \\

In 2024, \textit{in the context of single-frame detection}, Kumar et al. \cite{DL-47} design STNet (Small Target detection Network) based on the U2-Net (Nested U-shape Network) architecture. STNet comprises two modules. First, the Multi-Layer Feature Fusion (MLFF) improves the model’s ability to capture and exploit features from multiple layers, combining low-level details with high-level semantic information. Second, the Fast Fourier Block (FFB) enhances the model’s performance by enabling feature extraction in the frequency domain. Experiments on the NUDT-SIRST and IRSTD-1K datasets show that STNet outperforms previous SORA methods in cluttered backgrounds. \\

In 2024, \textit{in the context of dense moving target detection}, Chen et al. \cite{Dataset-300} develop a Linking-Aware Sliced Network (LASNet\footnote{ https://github.com/UESTC-nnLab/DMIST}) which consists of visual feature extraction, motion feature extraction and motion-affinity fusion. Experiments show that LASNet outperforms previous SOTA methods in both sparse and dense target scenarios. \\

In 2025, Chen  et al. \cite{DL-98} propose an Infrared Deep dense connection network based on an U-Net architecture (ID-UNet\footnote{https://github.com/AngryWaves/ID-UNet}). First, an Infrared Small Target Feature Extraction (ISTFE) module is embedded within the ID-UNet architecture to allow cross-layer and continuous interaction between deep high-level and shallow low-level features. Consecutive connections within ISTFE’s fusion allow the preservation of semantic information in deep layers, as well as the resolution information in shallow layers. Second, the UNet structure parameters are compressed reducing the parameters by $81\%$ compared to the conventional UNet. Experiments demonstrate that ID-UNet provides better performance than previous methods in segmentation metrics, including Intersection over Union (IoU), normalized IoU (nIoU), and F1 score.  Zhu  et al. \cite{DL-90} propose a Patch-wise Fast Fourier transform (PFFT). First, it divides the input image into patches and applies the Fourier transform to each patch. Second, convolutional neural networks (CNNs) are employed for learnable high- and low-frequency partitioning via a Learnable Frequency Augmentation module (FAM). More specifically, FAM captures the inter-frequency and intra-frequency feature. Third, a Frequency Spatial Fusion Module (FSFM) merges features from frequency and spatial domains. Experiments show that PFFT outperforms state-of-the-art methods.  Sun et al. \cite{DL-130} propose a denoising detection framework named IR-SDTNet which adopts an enhanced feature representation via dual modeling of local and global features. More specifically, a multiscale feature fusion mechanism combined with a jump connection integrates the encoder’s multi-level features with the decoder’s up-sampling features.  Experiments show that IR-SDTNet suppresses most noise and helps restore fine details under high-noise backgrounds. \\

In 2025, \textbf{in the context of Kolmogorov–Arnold Networks (KAN)},  Wu  et al. \cite{DL-121} design a Kolmogorov–Arnold Network (KPF-Net) which is based on a “Perception first, Fusion then” paradigm. First, a perception module uses convolutional Kolmogorov–Arnold Networks with learnable nonlinear activation functions to extract discriminative target–background features. These features dynamically parameterize convolutional kernels to guide detail fusion. Second, Wu  et al. \cite{DL-121} reconstruct input tensors as attention weights for memory tensors. Experiments on the  SIRST-v1 and IRSTD-1k datasets show that KPF-Net provides state-of-the-art probability of detection and normalized intersection over union. In a further work, Yuan et al. \cite{DL-164} propose a Sparse-sine Perception Kolmogorov–Arnold Network (SP-KAN\footnote{https://github.com/xdFai}) which aims to exploit the superior nonlinear ability of the Kolmogorov–Arnold theory. First, a compressed vision transformer encoder is employed to capture long-range spatial dependencies. Second, a Pattern Complementarity Module (PCM) captures their essential nonlinear interactions. More specifically, PCM unifies channel-wise mappings of tokenized representations with local spatial saliency of structured features. Third, a sparse-sine perception Kolmogorov–Arnold layer (SPKAL) captures the original nonlinear space and a sparse grid-based high-dimensional sinusoidal latent space at the pixel level. Experiments demonstrate that SP-KAN outperforms previous state-of-the-art IRSTD methods in accuracy, robustness, and generalization demonstrating its superior capability in sparse nonlinear modeling. \\

In 2025, \textbf{in the context of multi-frame detection}, Zhang et al. \cite{DL-139} design an Accumulation-Difference-based Siamese U-Net (ADSUNet). First, multi-scale spatial salient features in inter-frame images are extracted by Siamese network. Second, a accumulation-difference attention module adaptively fuses the inter-frame feature maps. Third, a probability likelihood diagram  is employed by a decoding network that restores the scale of feature map. Experiments show that ADSUNet outperforms model-based methods (MPCM, HBMLCM, WSLCM, RLCM, TLLCM, NIPPS \cite{SFD-RPCAMatrix-5}, RIPT, WLDM, FKRW, MGRG, STLCF) and data-based methods (ISTDU-Net \cite{DL-506}, RISTD-Net, DNA-Net \cite{DL-12}, DTUM-Net) in terms of target detectability, false alarm suppressibility, real-time performance, with AUC values of 0.993 and 0.998 in two datasets (SCR < 3), 30.97 fps and a parameter quantity of 1.498 M. In 2026, Deng et al. \cite{DL-599-76} develop a Spatio-Temporal Cooperative Expanded U-shape Network (STCEUNet). First, a progressive spatio-temporal hybrid enhancement architecture is designed in the feature extraction backbone. It establishes cross-frame information transmission channels at different levels and scales for deep interactive fusion of complementary spatio-temporal information. Second, a Central Difference Attention Froup (CDAG) and Multi-Frame Feature Fusion Module (MFFM) are employed for intra-frame feature aggregation and inter-frame feature fusion, respectively. CDAG performs adaptive gradient enhancement based on local saliency to provide effective spatial features while MFFM encodes target motion patterns by computing the correlation and difference in spatio-temporal context between adjacent frames. It distinguishes targets from background based on motion consistency. Experiments show that STCEUNet provides better performance that DTUM, Tridos, STMENet and ADSUNet.

In 2026, Li et al. \cite{DL-293} develop a U-shaped segmentation network architecture (CN-UNet). First, CN-UNet uses a ConvNeXt module as the encoder, which implements a self-attention mechanism via convolutional operations as in  Swin Transformer’s functionality while expanding the receptive field interactions of target features. Second, a Multiscale Local Contrast Learning Module (MSLCLM) selectively filters features aligned with human visual priors. Third,  a slicing-aided hyper-segmentation module  dynamically rescales output images via adaptive slicing operations. Experiments show that  CN-UNet outperforms previous SOTA methods with better visual quality and index evaluation metrics. Li et al. \cite{DL-301} propose a Dynamic High-Frequency convolution (DHiF\footnote{https://github.com/TinaLRJ/DHiF}) to translate the discriminative modeling process into the generation of a dynamic local filter bank. More specifically, DHiF is sensitive to high-frequency components (HFCs) owing to the dynamic parameters of its filters symmetrically adjusted within a zero-centered range according to Fourier transformation properties. DHiF can adaptively and dynamically processes different HFC regions and captures their distinctive grayscale variation characteristics for discriminative representation learning. Experiments on the IRSDT-1k and NUAA-SIRST datasets show that DHiF outperforms most of the time conventional deep neural networks (FC3-Net \cite{DL-335}, DNA-Net \cite{DL-12}, ISNet \cite{DL-211}, MSHNet), and CNN-Transformer hybrid (CNN-T) methods (MTU-Net \cite{Applications-S-41}, SCTransNet \cite{DL-36}, APTNet \cite{DL-599-162}).

\subsection{Learning Mechanims}
Learning mechanisms can be categorized into point supervised learning \cite{DL-363,DL-364,DL-367,DL-368}, weakly supervised learning \cite{DL-365,DL-366,DL-599-132} and source-free domain adaptation learning \cite{DL-390,DL-393,DL-599-232}.  Data Augmentation is also a strategy to improve the learning issue \cite{DL-67,DL-317,DL-397,DL-235}.  Table \ref{OverviewLearningM} shows an overview of the learning mechanisms.

\begin{table*}
\begin{center}
\scalebox{0.79}{
\begin{tabular}{|l|l|}
\hline
\scriptsize{Methods} &\scriptsize{Acronyms} \\
\hline
\hline
\scriptsize{Conventional Learning Mechanisms} &\scriptsize{MTMLNet [Single-Frame] (Yang et al. \cite{DL-115}), MTMLNet+ [Single-Frame] (Luo et al. \cite{DL-576}), TDCNet [Multi-Frame][MIRST] (Fang et al. \cite{DL-117})} \\
\scriptsize{}  &\scriptsize{LODDIS [Multi-Frame][MIRST] (Chen et al. \cite{DL-195})} \\
\scriptsize{}  &\scriptsize{GGL-Net (Zhao et al. \cite{DL-302}), S$D^2$Net (Hu et al. \cite{DL-303})} \\
\hline
\scriptsize{Single Point Supervised Learning}  &\scriptsize{PAL (Yu et al. \cite{DL-368}), LELCM (Yang et al. \cite{DL-367})} \\
\scriptsize{[Plug and Play]}  &\scriptsize{LESPS (2023) [Single-Frame] (Ying et al. \cite{DL-364}), LESPS (2024) (Shao et al. \cite{DL-520})} \\
\scriptsize{}  &\scriptsize{GMVG-DIR (Shi et al. \cite{DL-363}), GSACP (Zhou and Zhang \cite{DL-567}), EDGSP (Huan et al. \cite{DL-593}) }\\
\scriptsize{}  &\scriptsize{LSP (Yao et al. \cite{DL-599-3}), DtoD (Liu et al. \cite{DL-599-6}), TGPL (Li and Zhang \cite{DL-599-75}), SPL-HR (Wu et al. \cite{DL-599-120}),  PAL-NA (Cheng et al. \cite{DL-599-178})} \\
\scriptsize{}  &\scriptsize{PGCE (Cui et al. \cite{DL-599-266}), DME (Cao et al. \cite{DL-599-393})} \\
\hline
\scriptsize{Weakly Supervised Learning}  &\scriptsize{WSSISTD (Cui et al. \cite{DL-365}), PMG (He et al. \cite{DL-366}), MCMIN (Chen et al. \cite{DL-599-132})} \\
\scriptsize{}  &\scriptsize{UNSoL [Plug-in] (Chen et al. \cite{DL-599-343}), UDBR-Net (Ma et al. \cite{DL-599-379})} \\
\hline
\scriptsize{Source-Free Domain Adaptation Learning}  &\scriptsize{IRSTD-SFDA (Jin et al. \cite{DL-393}), PDSECL (Wang et al. \cite{DL-390}), SF-IRPure (He et al. \cite{DL-599-232}) } \\
\scriptsize{}  &\scriptsize{S2-SFDA (Li et al. \cite{DL-599-318}) } \\
\scriptsize{Active Domain Adaptation Learning}  &\scriptsize{ADA-IRSTD (Li et al. \cite{DL-599-230}) } \\
\scriptsize{Multi-Domain Learning}  &\scriptsize{DMLNet [MIRST] (Rong et al. \cite{DL-599-265}) } \\
\hline
\scriptsize{Contrastive Learning} &\scriptsize{FGFRNet (Luo et al. \cite{DL-569}), WeCoL (Duan et al. \cite{DL-599-14}), AC-SLSIoU (Zeng et al. \cite{DL-599-267}, TACG-Net (Dai et al. \cite{DL-599-367} } \\
\hline
\scriptsize{Interactive Learning}  &\scriptsize{HIL-Net (Wang et al. \cite{DL-599-106})} \\
\hline
\scriptsize{Progressive Learning}  &\scriptsize{PFLNet (Anonymous \cite{DL-599-440})}\\
\hline
\scriptsize{Reinforcement Learning}  &\scriptsize{FGL-IRSTD (Liu et al. \cite{DL-599-330})} \\
\hline
\scriptsize{Dual-Stream Edge-Target Learning} &\scriptsize{DETL-Net (Yao et al. \cite{DL-532})} \\
\hline
\scriptsize{Data Augmentation} &\scriptsize{Random-Cloning Models: TSMSA [MIRST] (Peng et al. \cite{DL-67})} \\
\scriptsize{} &\scriptsize{Random-Erasing Models: AugTarget (Chen et al. \cite{DL-397})} \\
\scriptsize{} &\scriptsize{Prior-Guided Models: PGDA (Wang et al.  \cite{DL-599-136}), PGDNet  (Pang et al.  \cite{DL-599-305}), BTPGNet (Chen et al. \cite{DL-599-376}} \\
\scriptsize{} &\scriptsize{Thermodynamics-Inspired Models: SemiAugIR (Zhang et al.  \cite{DL-599-140})} \\
\scriptsize{} &\scriptsize{GANs Models: HQDA (Chen et al. \cite{DL-317}), StyleGAN (Wu et al. \cite{DL-599-105}), MGD-GAN (Ding et al.  \cite{DL-599-137})} \\
\scriptsize{} &\scriptsize{Diffusion Models: Diff-Mosaic (Shi et al. \cite{DL-235}, ISTDiffuser (Zang et al.  \cite{DL-599-174}) }\\
\scriptsize{} &\scriptsize{Hybrid Models: HDMGAN (Ding et al.  \cite{DL-599-130}) } \\ 
\hline
\end{tabular}}
\caption{Learning Mechanisms: An Overview (44 papers).}
\label{OverviewLearningM}
\end{center}
\end{table*}

\begin{enumerate}
\item \textbf{Point Supervised Learning:}  Point-supervised learning methods in infrared small target detection have been employed in clustering-based approaches \cite{PSTD-37,PSTD-46,PSTD-47}, deep learning based approaches \cite{DL-363,DL-364,DL-365,DL-366,DL-368} and hybrid approaches (LELCM\footnote{https://github.com/haopzhang/LELCM}\cite{DL-367}). More specifically, the deep neural networks based methods require dense pixel-level annotations, which are costly and often unreliable for tiny and weak targets. In this context, point supervision provides a more practical alternative but the current point-supervised methods usually construct pseudo-labels based on the distance between pixels and annotated points or cluster centers, which generates spatial bias and  miss genuine target pixels away from these reference points.  In 2023, Ying et al. \cite{DL-364} propose the first work with point-level supervision. Based on the observation that CNNs first learn to segment a cluster of pixels near the targets, and then gradually converge to predict ground truth point label during the training phase supervised by point labels, Ying et al. \cite{DL-364} employ a label evolution framework named Label Evolution with Single Point Supervision (LESPS\footnote{https://github.com/XinyiYing/LESPS}) to progressively expand the point label by exploiting the intermediate predictions of CNNs. Thus, the network predictions finally estimate the updated pseudo labels. The pixel-level target mask is obtained to train CNNs in an end-to-end manner. Experiments show that CNNs trained with LESPS recover over $70\%$ and $95\%$ of their fully supervised performance in terms of pixel-level intersection over union and object-level probability of detection, respectively. In 2024, Shao et al. \cite{DL-520} propose  also a Label Evolution with Single Point Supervision (LESPS) framework giving the same acronym than the previous method \cite{DL-364}. To improve the detection accuracy, a combination of Test-Time Augmentation (TTA) and Conditional Random Field (CRF) for post-processing. An Adjustable Sensitivity (AS) strategy for post-processing employs the advantages of multiple detection results and adds some areas with low confidence to the fine detection image in the form of centroid points. Finally, a multi-stage loss is employed for fine-grained detection. Experiments show that LESPS achieves SOTA performance. Note that LESPS was ranked at the third place in the “ICPR 2024 Resource-Limited Infrared Small Target Detection Challenge Track 1: Weakly Supervised Infrared Small Target Detection”. In 2025, Yu et al. \cite{DL-368} design
a Progressive Active Learning (PAL\footnote{https://github.com/YuChuang1205/PAL}) method. More specifically, a model pre-start concept allows to focus on automatically selecting a portion of easy samples and helping the model to have basic task-specific learning abilities. A refined dual-update strategy allows to promote learning of harder samples and continuous refinement of pseudo-labels. Finally, a decay factor is employed to achieve a dynamic balance between the expansion and contraction of target annotations. In 2026, Shi et al. \cite{DL-363} design a point-supervised method  which employs Gradient-Guided Minimum Variance Growth (GMVG) with Deep Iterative Refinement (DIR). First, GMVG allows to obtain target likelihood from gradient-guided aggregation of contour closure and edge responses. Then, GMVG converts it into structurally coherent pseudo-labels via a Minimum Variance Growth filter. Second, DIR improves the pseudo-labels by employing reliable semantic guidance into an iterative refinement process, thereby reducing error propagation. Thus, irregular target shapes are better preserved. Experiments on NUDT-SIRST, IRSTD-1k, and NUAA-SIRST datasets show that GMVG-DIR improves pseudo-label fidelity and offers competitive point-supervised performance. \\

In 2026, Zhou and Zhang \cite{DL-567} propose to generate a point-to-mask supervision online via an in-batch, point-anchored feature-affinity propagation. This framework named GSACP is an end-to-end testbed that supervises the detector using hard-margin feature affinity gated by local image priors avoiding external label-evolution loops.  Zhou and Zhang \cite{DL-567} theoretically formalize this as Self-Referential Propagation Drift which is  a representation-supervision entanglement that can sharpen true boundaries or distort the feature space to satisfy its own targets. A protocolized single-variable ablation procedure is employed for spanning local EMA teacher decoupling, hard-background contrastive separation, and adaptive support geometry. Experiments on the SIRST3 dataset show that  GSACP outperforms PAL \cite{DL-368}.  Ni et al. \cite{DL-231} propose a  Single-Point Supervision guided Infrared Probabilistic Response Encoding method (SPIRE\footnote{https://github.com/NIRIXIANG/SPIRE-IRSTD}). First, a Point-Response Prior Supervision (PRPS) transforms single-point annotations into probabilistic response map consistent with infrared point-target response characteristics. Second, a High-Resolution Probabilistic Encoder (HRPE) enables encoder-only, end-to-end regression without decoder reconstruction.  SPIRE alleviates optimization instability under sparse target distributions. Extensive experiments on the SIRST-UAVB and SIRST4 datasets show that SPIRE provides competitive detection performance with consistently low false alarm rate (Fa) and significantly reduced computational cost compared to ACM \cite{DL-10}, ALCNet \cite{DL-11}, RDIAN \cite{DL-33}, DNA-Net \cite{DL-12}, MSHNet \cite{DL-41} and SCTransNet \cite{DL-36}. Li and Zhang \cite{DL-599-75} design a Teacher-Guided Progressive Learning (TGPL) method for point supervision. TGPL employs the dynamic evolution of point labels with the feature learning capability of existing models. First, initial pseudo-masks are obtained using a Pseudo-Label Generation (PLG) method and high-confidence samples are selected for model initialization. Second, a progressive sample mining strategy gradually employs samples according to the model capacity. Third, a teacher–student collaborative mechanism iteratively refines pseudo-labels. Experiments on the NUDT-SIRST, IRSTD-1-k, and SIRST3 datasets show that TGPL achieves superior training performance under point supervision than PAL \cite{DL-368}  and LESPS \cite{DL-520}. TGPL reaches $99\%$ of the performance  in terms of detection probability (Pd). \\
\item \textbf{Weakly Supervised Learning:} In 2025, He et al. \cite{DL-366} design a weakly-supervised paradigm to obtain high-quality pseudo masks from the point-level annotation by using a learning-free technique with an hybrid of the learning-based method. The learning-free method is based on a sequential process, progressing from a point annotation to the bounding box that encompasses the target, and subsequently to detailed pseudo masks, while the hybrid is obtained via filtering out false alarms and retrieving missed detections in the network's prediction to obtain a supplement for learning-free masks. Experiments show that this learning-free method provides pseudo masks with an average Intersection over Union (IoU) that is $4.3\%$ higher than the second-best learning-free competitor while the hybrid learning-based method further enhances the quality of pseudo masks, achieving an additional average IoU increase of $3.4\%$. In an other work, Cui et al. \cite{DL-365} propose a Weakly Semi-Supervised Infrared Small Target Detection (WSSISTD) method. This method consists of a teacher network that provides coarse pseudo-masks and a student network that focuses on the final detection task. Cui et al. \cite{DL-365} employ a minimal proportion of pixel-level annotated data to train the teacher network, which subsequently provides coarse pseudo-masks for a volume of cost-effective point annotation data. Then, an adaptive Random Walker (RW) strategy allows to exploit point labels for obtaining RW pseudo-masks. To deal with the limitations of the teacher network trained on a limited amount of pixel-level labels, a RefineNet integrate and refine multiple coarse pseudo-masks. Finally, a combination of partially real masks and refined pseudo-masks is used to  train the student network reducing the labeling cost. Experiments show that WSSISTD provides similar performance than fully supervised methods. \\
\item \textbf{Source-Free Domain Adaptation Learning:} Most the deep neural networks based methods assume that the training and testing data come from the same distribution, an assumption that does not hold in many real-world scenarios. In addition, the impossibility to access source domain data complicates the domain adaptation process. To address these problems, Jin et al. \cite{DL-393} propose a Source-Free Domain Adaptation (SFDA) method which consists of two modules. First, the Multi-expert Domain Adaptation (MDA) module allows the source model to provide pseudo masks for the target domain, facilitating the transfer of knowledge from the source to the target domain. MDA refines these pseudo masks through a series of operations, including target localization,  shape adaptation, and multi-expert decision. Second, the Multi-scale Focused Learning (MFL) module is based on a global-local fusion strategy to focus on critical regions increasing the model's ability to detect small infrared targets.  In 2026, Wang et al. \cite{DL-390} design a Prior-Driven Self-Evolving Curriculum Learning (PDSECL) method which consist of three strategies. 
First, the Explicit Prior-driven Initial Sample Pool Construction (EPISPC) strategy employs consistency and local contrast priors to denoise initial pseudo-labels obtained from the source model in the target domain, and selects easy samples to obtain an initial sample pool, laying a reliable foundation for early-stage adaptation. Second, the Implicit Prior-driven Pseudo-label Refinement (IPPR) strategy allows implicit dynamic information during model training to obtain continuous optimization and updating of pseudo-labels. Third, the Consolidated Prior-driven Curriculum Augmentation (CPCA) strategy treats samples as a consolidated prior and synthesizes harder samples by fusing them with extracted background regions using a curriculum learning mechanism to progressively improve the model's generalization ability. Experiments  on the SIRST, NUDT-SIRST, and IRSTD-1k datasets show that PDSECL improves the adaptive performance of source domain methods to new domains. \\
\item \textbf{Active Domain Adaptation Learning:} Even if unsupervised domain adaptation allows to mitigate limited sample sizes and limited scene diversity a notable difference with fully supervised methods remains, limiting practical
applicability. To address this problem, Li et al. \cite{DL-599-230} design an  Active Domain Adaptation framework for IRSTD (ADA-IRSTD) which consists of three modules. strategies. First, a Target-Proximity
Preserving Masking (TPPM) allows to mask background regions while preserving target and neighboring features, thereby enhancing cross-domain robustness. Second, a False Alarm Prediction Module (FAPM) allows to quantify target indistinguishability and assess the risk of ambiguous detections. Third, a Global Distribution Uncertainty Sampling (GDUS) guides active sample selection based on global prediction uncertainty. Experiments on the NUDT-SIRST, IRSTD-1k, SIRST, and SIRST-v2 datasets show that ADA-IRSTD outperforms previous active domain adaptation (EADA \cite{DL-599-230-1}, RIPU \cite{DL-599-230-2}, DUC \cite{DL-599-230-3}) and MMIDNet . \cite{DL-599-141}. \\
\item \textbf{Dual-Stream Edge-Target Learning:} In 2024, Yao et al. \cite{DL-532} propose a Dual-stream Edge-Target Learning Network (DETL-Net\footnote{https://github.com/rayyao/DETL-Net})  which enhances feature cross-fusion by learning edge details and target regions via a dual-stream framework. First, multilevel features of the image based on the encoder-decoder structure of U-Net are extracted and the feature map is obtained Second, a Dual-GUided C-Fusion Module (DGCFM) capture edge details of small targets and global contextual features of the target region. Third, a Multiscale Context Fusion Module (MCFM) within DGCFM employs central difference convolution to enhance local contrast and extract rich contextual details, thereby retaining edge information and enhancing overall target representation. A Cross-dimension Interactive Aggregation Attention Module (CIAAM) dynamically adjusts feature fusion weights across layers to deal with noise and improve the discrimination of small targets. These diverse modules are sequentially interconnected to progressively refine edge details. The target features are subsequently employed for predicting the final target mask through the segmentation head. Experiments on the NUAA-SIRST and IRSTD-1k datasets show that DETL-Net outperforms several previous SOTA methods. \\
\item \textbf{Contrastive Learning:} In 2026, Luo et al. \cite{DL-569} design a Frequency and Gradient Feature Refinement Network (FGFRNet\footnote{https://github.com/wangqian-w/FGFRNet}). FGFRNet comprises three modules. First, a Temperature-Aware Contrastive Learning (TACL) module exploits pixel-wise thermal priors to obtain discriminative positive and negative pairs in feature space. Second,a Frequency Attention Aware Fusion (FAAF) module employs global frequency domain context with spatial-domain details to enhance semantic localization. Third, a Texture Feature Enhancement (TFE) module uses gradient-based supervision to strengthen fine-grained texture cues. Experiments on the NUAA-SIRST, NUDT-SIRST, and IRSTD-1k datasets that FGFRNet outperforms MDvsFA \cite{DL-15}, ACMNet, ALCNet \cite{DL-11}, ISNet \cite{DL-211}, FC3-Net \cite{DL-335}, Dim2Clear \cite{DL-378}, MSHNet \cite{DL-41}, CSRNet, IRPruneDet \cite{DL-35}, SAGCPDet \cite{DL-60}, and MMLNet \cite{DL-114} in cluttered and low-contrast conditions. Duan et al. \cite{DL-599-14} propose a Weakly-supervised Contrastive Learning (WeCoL\footnote{https://github.com/UESTC-nnLab/WeCoL}) framework which requires simple target quantity prompts during model. Based on the pretrained Segment Anything Model (SAM), a target mining strategy integrates target activation maps. Second, a multi-frame energy  contrastive learning enhances the reliability of pseudo-labels based on the similarity between positive and negative samples in feature. Third,  a long-short term motion-aware learning framework  simultaneously model the local motion patterns and global motion trajectory of small targets. Experiments on the DAUB and ITSDT-15K datasets show  that WeCoL outperform several times early fully-supervised methods. \\
\item \textbf{Data Augmentation:} IRSTD faces a massive and diverse demand for samples using diverse augmentation strategies namely Random-Cloning models \cite{DL-67}, Random-Erasing models \cite{DL-397}, prior-guided models \cite{DL-599-136}, thermodynamics-Inspired models \cite{DL-599-140}, GANs \cite{DL-599-105,DL-599-105,DL-599-137}, diffusion models
\cite{DL-235,DL-599-174} and hybrid models \cite{DL-599-130}. However, several research generate infrared data for various scenes from visible images with Generative Adversarial Networks (GANs).  In 2023, Chen et al. \cite{DL-397} propose a data augmentation method named AugTarget\footnote{https://github.com/UESTC-nnLab/AugTarget} which consists of two steps: target augmentation and batch augmentation. First, the target augmentation allows to generate sufficient targets using random target representation. Second, the batch augmentation allows to diversify training samples. The initially-generated image samples of small targets are also enriched by randomly aggregating the feature representation of different images. Experiments show that AugTarget improves the mean intersection over union (mIoU). But, AugTarget is specifically
designed for single-frame images and fails to keep temporal correlations inherent in  image sequences. In 2024, Shi et al. \cite{DL-235} design Diff-Mosaic\footnote{https://github.com/YupeiLin2388/Diff-Mosaic}, a data augmentation method based on the diffusion model. This model attempt to alleviate the challenge of diversity and realism of data augmentation methods via diffusion prior. First an enhancement network called Pixel-Prior generates highly coordinated and realistic Mosaic images by harmonizing pixels. Second an image enhancement strategy named Diff-Prior uses diffusion priors to model images in the real-world scene, further enhancing the diversity and realism of the images.  In 2025, Peng et al. \cite{DL-67} design an augmentation method specifically for sequence-based infrared images. The input image sequences have the
number of targets of diverse sizes and locations increased while maintaining temporal coherence with the use of a target augmentation method and enriching the scene information within a single image through a batch augmentation algorithm. Thus, the augmented video clip is employed as input for a temporal modeling model in order to learn the semantic information. However, these two methods neglect the differences between different modal images, thus losing realism and usability. To address these issues, Chen et al. \cite{DL-317} develop in 2026 a High-Quality Data Augmentation (HQDA) technique. This technique offers better performance by using the intrinsic properties of infrared images. More specifically, a consistency-constrained background generation method allows to obtain high-quality infrared backgrounds from visible images. It uses the nonlocal self-correlation property of infrared images into GAN to improve the realism of generated data. To better preserve the real distribution during target enhancement, a target extraction method based on a robust random walker (RW) is employed.
\end{enumerate}

\subsection{Attention Mechanisms}
Attention mechanisms can be categorized into conventional attention mechanisms, cross attention mechanisms \cite{DL-599-7},pyramid attention mechanisms, hierarchial attention mechanisms, spatial attention mechanisms, direction-induced attention mechanisms and contextual attention Mechanisms. Table \ref{OverviewAttentionM} shows an overview of these diverse attention mechanisms.

\begin{table*}
\centering
\scriptsize
\renewcommand{\arraystretch}{1.25}
\begin{tabular}{|l|p{0.70\textwidth}|}
\hline
\textbf{Methods} & \textbf{Acronyms} \\
\hline
\hline
Conventional Attention Mechanisms & ALCNet [Single-Frame] (Dai et al. \cite{DL-11}), DNA-Net [Single-Frame][Dense] (Li et al. \cite{DL-12}), DAEA (Lin et al. \cite{DL-219}), DCANet (Zang et al. \cite{DL-43}), STPSA-Net (Liu et al. \cite{DL-74}), STAM-UNET (Shahraki et al. \cite{DL-119}), FAA-Net (Zhuang et al. \cite{DL-105}), DENet (Zuo et al. \cite{DL-107}), DPS-Net (Liu and Huang \cite{DL-138}), NCENet (Wu et al. \cite{DL-140}), DCGANet (Chen et al. \cite{DL-163}), CSQNet (Yuan et al. \cite{DL-202}), CARG (Zhang et al. \cite{DL-267}), C2IRSTD (Wang et al. \cite{DL-294}), SSCFNet (Wang et al. \cite{DL-296}), PHFNet (Li et al. \cite{DL-379}), LCAE-Net (Wang et al. \cite{DL-394}), CS-LDRNet (Lei et al. \cite{DL-395}), SWA-HFFM (Miao et al. \cite{DL-533}), CCRANet [Single-Frame] (Wang et al. \cite{DL-599-61}), BDSNet (Wang et al. \cite{DL-599-221}), BCEANet (Li et al. \cite{DL-599-312}) \newline
\textit{Gabor Modulation:} GMN (Bian et al. \cite{DL-599-437}) \quad \textit{Adaptive:} A2-STMC (Wu et al. \cite{DL-599-317}) \quad \textit{Dual-Attention:} MCC-Net (Zhou et al. \cite{DL-599-200}) \newline
\textit{Curvature:} SCMT-Net (Yang et al. \cite{DL-360}), CHFNet (Zhang et al. \cite{DL-599-64}) \quad \textit{Multi-Attention:} MFAM (Li et al. \cite{DL-599-291}) \\
\hline
Top-K Attention Mechanisms & LGTS (Huang et al. \cite{DL-599-370}), TriHANet (Li et al. \cite{DL-599-371}), HPFNet (Zhang et al. \cite{DL-168}), DifRankNet (Fang et al. \cite{DL-599-372}) \\
\hline
Pyramid Attention Mechanisms & FPN (Tong et al. \cite{DL-374}), NS-FPN [Detection] (Yuan et al. \cite{DL-304}), FE-FPN (Wang et al. \cite{DL-599-242}), PAU2-Net (Peng et al. \cite{DL-45}), MCFD (Zong and Shan \cite{DL-599-21}), Nested FPN (Lu et al. \cite{DL-599-142}), ISPANet (Wang et al. \cite{DL-599-214}), MAPFF (Yang et al. \cite{DL-599-241}) \\
\hline
Hierarchical Attention Mechanisms & HAFNet (Zhang et al. \cite{DL-102}) \\
\hline
Spatial Attention Mechanisms & SANet (2023) (Zhu et al. \cite{DL-32}), DSAD (Li et al. \cite{DL-535}), CSAN-UNet (Zhong et al. \cite{DL-599-107}), CenterADNet (Sun et al. \cite{DL-599-164}), DSDF-Net (Dong et al. \cite{DL-599-402}), SaKD (Ma et al. \cite{DL-599-404}) \\
\hline
Direction-Induced Attention Mechanisms & RDIAN [Single-Frame] (Sun et al. \cite{DL-33}), LDCNet (Tang et al. \cite{DL-34}), ISNet (Zhang et al. \cite{DL-211}), P2P-HDNet (Ni et al. \cite{DL-68}), EDDNet (Ji et al. \cite{DL-398}), DTUM (Li et al. \cite{DL-512}), DCONet (Zhang et al. \cite{DL-599-10}) \\
\hline
Asymmetric Attention Mechanisms & AAU-Net (Tong et al. \cite{DL-594}) \\
\hline
Circular Attention Mechanisms & CFANet (Ji and Wang \cite{DL-599-395}) \\
\hline
Contextual Attention Mechanisms & ACM (U-Net) [Single-Frame] (Dai et al. \cite{DL-10}), U2-Net (Qin et al. \cite{DL-13}), DCAPNet (Chen et al. \cite{DL-42}), KCPNet [Ship] (Han et al. \cite{DL-30}), FSCFNet (YOLOv1) [Detection] (Xu et al. \cite{DL-111}), GAC-Net (Zhang et al. \cite{DL-342}), KDDNet [Ship] (Zhang et al. \cite{Applications-S-35}), URAnet (Wang et al. \cite{DL-504}), MDCENet (Ma et al. \cite{DL-590}), LRFNet (Wang et al. \cite{DL-578}), BAUENet (Chen et al. \cite{DL-599-66}), ACANet (Ling et al. \cite{DL-599-114}), SCGNet (Zong and Zhu \cite{DL-599-264}), SCUA-Net (Lin et al. \cite{DL-599-272}), RFONet (Zhang et al. \cite{DL-599-301}), CGPFNet (Yuan et al. \cite{DL-599-405}) \\
\hline
Coordinate Attention Mechanisms & SRCANet [Ship] (Wu et al. \cite{Applications-S-44}), CAFF-Net (Shi et al. \cite{DL-599-78}) \\
\hline
Energy Field Attention Mechanisms & CEFANet (Fang et al. \cite{DL-560}) \\
\hline
Fine-Grained Channel Attention Mechanisms & FCCF-Net (Liu et al. \cite{DL-599-71}) \\
\hline
Interior Attention Mechanisms & IAANet (Wang et al. \cite{DL-336}) \\
\hline
Multi-Path Attention Mechanisms & MAPPR++-Net (Abbas et al. \cite{DL-599-44}) \\
\hline
Multi-Feature Attention Mechanisms & MFAFNet (Zhao et al. \cite{DL-599-163}) \\
\hline
Patch Attention Mechanisms & APAFNet (Wang et al. \cite{DL-599-219}) \\
\hline
Hypergraph Attention Mechanisms & MSCD-HA (Zhang et al. \cite{DL-564}) \\
\hline
\end{tabular}
\caption{Attention Mechanisms: An Overview (81 papers).}
\label{OverviewAttentionM}
\end{table*}

\begin{enumerate}
\item \textbf{Top-K Attention Mechanisms:} Top-k selection preserves the most task-relevant responses suppressing redundant features and improving attention sparsity. For IRSTD, Top-k selection is employed to suppress background interference and enhance target responses. In 2025, Li et al. \cite{DL-599-371} developed a Top-k mechanism in TriHANet to strengthen target-aware self-attention responses. In 2025, Fang et al. \cite{DL-599-372} proposed DifRankNet, which applies Top-k selection in both spatial and channel dimensions to improve feature extraction capability.  In 2026, Zhang et al. \cite{DL-168} designed HPFNet with a dual-k selection and fusion strategy to improve detection performance. In 2026, Huang et al. \cite{DL-599-370} design LGTS which extracts hierarchical multi-scale target representations using complementary local and global sparse attention mechanisms. First, a global attention Top-k selection module (GATS) is employed at the deepest feature level to perform sparse transformer attention modeling. Second, a Local Attention Top-k Selection module (LATS) is used for multi-scale shallow features, where a progressive coarse-to-fine sparse attention refinement strategy allows to strengthen local target saliency and suppress background interference. \\
\item \textbf{Pyramid Attention Mechanisms:} In 2021, Tong et al. \cite{DL-374} propose a feature fusion approach of Feature Pyramid Networks (FPN).  A multi-scale feature fusion module allows to  extract the feature maps of infrared small target in different layers of the network and to fuse them effectively. Experiments show that FPN outperforms previous deep learning-based methods. In 2024, Peng et al. \cite{DL-45} design a Pyramid Attention U-shaped network (PAU2-Net). More specifically, a pyramid attention encoder allows to establish long-range channel dependencies and improve the interaction between local and global information. PAU2-Net can extract small target features and improves high-level understanding of the scene to reduce false alarm rates. About the adaptive multiscale supervision method, Peng et al. \cite{DL-45} generate a group of multiscale labels by pooling to guide the deep outputs of the network. The labels maintain the target position information at different scales and help preserving the target information as the network deepens. In 2025, Yuan et al. \cite{DL-304} improve performance from noise suppression perspective by designing a Noise-Suppression Feature Pyramid Network (NS-FPN). It integrates a Low-Frequency guided Feature Purification (LFP) module and a Spiral-aware Feature Sampling (SFS) module into the original FPN structure. The LFP module allows to tackle the noise features by purifying high-frequency components to obtain feature enhancement without noise interference. The SFS module uses spiral sampling to fuse target-relevant features in feature fusion process. NS-FPN can be integrated into existing IRSTD frameworks. Experiments on the IRSTD-1k and NUAA-SIRST datasets  show that NS-FPN significantly reduces false alarms. \\
\item \textbf{Hierarchical Attention Mechanisms:} In 2025, Zhang et al. \cite{DL-102} propose a Hierarchical Attention Fusion Network (HAFNet) based on the U-Net architecture. More specifically, a Dual-Branch Semantic Perception Module (DSPM) is designed as the feature extraction backbone to increase contextual semantic interactions. This module is based on a dual-branch feature extraction with standard and dilated convolutions while using spatial and Channel Attention Modules (CAMs) to distinguish small targets from background noise. A skip connection is employed by merging a Hierarchical Feature Fusion Encoder (HFFE) and a Hierarchical Feature Fusion Decoder (HFFD). These two modules employ hierarchical attention-guided and encoded feature injection skip connections (ESCs) to obtain  fusion of multiscale and multilevel semantic features between the encoder and decoder. Experiments on the NUAA-SIRST, IRSTD-1K, and NUDT-SIRST datasets show that HAFNet outperforms previous  IRSTD methods. \\
\item \textbf{Spatial Attention Mechanisms:} In 2023, Zhu et al. \cite{DL-32}  propose a Spatial Attention Network (SANet) with global average contrast learning. Instead of common detection strategies by pixel-level segmentation, SANet extends conventional contrast methods to target detection framework of deep learning in order to obtain robust performance. For feature extraction, a group of Cross Stage Partial Networks (CSPNet) allows to capture the local semantic information. A cluster of Spatial Attention Modules with global average contrast (SAG) allows to obtain global spatial semantics. A series of Selective Kernel convolution (SKConv) fuses semantic and spatial features. Finally, a Spatial Pyramid Pooling (SPP) scheme is employed for the detection model. Experiments on two public datasets show that SANet outperforms most of the time state-of-the-art methods. \\
\item \textbf{Direction-Induced Attention Mechanims:} In 2022, Zhang et al. \cite{DL-211} propose an Infrared Shape network (ISNet). More specifically, a Taylor Finite Difference (TFD) based on edge block aggregates and improves the edge information from different levels to increase the contrast between target and background and also lay a foundation for extracting shape information with mathematical interpretation.  A Two-Orientation Attention Aggregation (TOAA) block first computes the low level information with attention mechanism in both row and column directions. Second, TOOA fuses it with the high-level information to take into account the shape characteristic of targets and to suppress noise.  Experiments on the IRSTD-1k show that ISNet outperforms SOTA algorithms. In 2023, Sun et al. \cite{DL-33} design a Receptive-field and Direction-Induced Attention Network (RDIAN). Using the characteristics of target size and grayscale, RDIAN allows to take into account the interclass imbalance between targets and background. Based on convolutional layers with different receptive fields in feature extraction, target features in different local regions are obtained increasing the diversity of target features. Finally, the targets are enhanced in low-level feature maps employing a multidirection guided attention mechanism. In 2024, Tang et al. \cite{DL-34} develop
a Low-rank Differential Cascade Network (LDCNet) to consider the physical properties and deep cascade features of infrared images. First, a cascade feature extraction module is obtained through a multilevel coplanar cascade encoder–decoder structure. This module employs both the deep-level and low-level features of infrared targets and backgrounds. Second, a differential attention mechanism based on the change differential analysis and robust principal component analysis is employed. Finally, the multilevel feature fusion module allows to adaptively employ the spatial and semantic information of different depth feature maps in order to estimate the final detection. \\
\item \textbf{Contextual Attention Mechanisms:}  In 2020, Qin et al. \cite{DL-13} propose a U-Net based methods; The architecture is a two-level nested U-structure which captures more contextual information from different scales and increases the depth of the whole architecture. This architecture can train a deep network from scratch without using backbones.
In 2021, Dai et al.  \cite{DL-10} design an Asymmetric Contextual Modulation (ACM\footnote{https://github.com/YimianDai/open-acm}) module based on an U-Net architecture. In addition of a top-down global contextual feedback, a bottom-up modulation pathway based on point-wise channel attention for exchanging high-level semantics and subtle low-level details is employed.  Experiments show that ACM outperforms local contrast measurement methods (Tophat, LCM \cite{SFD-LCM-4}, ILCM, LSM, FKRW, MPCM), LRSD methods (IPI \cite{SFD-RPCAMatrix-1}, NIPPS \cite{SFD-RPCAMatrix-5}, RIPT, PSTNN \cite{SFD-RPCATensor-5}, NRAM \cite{SFD-RPCAMatrix-10}) and deep neural networks methods (FPN \cite{DL-374}). In 2024, Chen et al. \cite{DL-42} design a Dynamic Context-Aware Pyramid Network (DCAPNet) which consists of a Dynamic Context Modulation (DCM), a Dynamic Pyramid Context (DPC), and a Shuffle Attention Fusion (SAF). First, the DCM module adaptively captures diverse-scale information adapting to various target dimensions, and enhancing the feature representation abilities. Second, the DPC module allows to adaptively captures multiscale features and  employs contextual information by aggregating multiple DCM modules. Third, the SAF module allows to exchange of information within the same layer and establish correlations between different layers. Experiments on the IRSTD-1 k, SIRST-Aug, and NUDT-SIRST  datasets show that DCAPNet outperforms several previous algorithms. \\
\item \textbf{Energy Field Attention Mechanisms:} In 2026, Fang et al. \cite{DL-560} propose a Cross-layer Energy Field Aggregation Network (CEFANet). It addresses  target energy discontinuities embedded in cluttered backgrounds to achieve dynamic discrimination and feature enhancement of small targets with varying states. First, an Energy Field Attention (EFA) module selectively amplifies target energy mutations while suppressing background noise. Second, a Cross-layer State Fusion (CSF) module based on channel-wise state awareness and adaptive weighting mechanisms allow precise modulation of low-level features under the guidance of high-level semantic states. Experiments show that CEFANet improves the detection performance of infrared small targets. \\
\item \textbf{Hypergraph Attention Mechanisms:} In 2026, Zhang et al. \cite{DL-564} design a network that couples a Multiscale Convolutional Decoder (MSCD) with a Hypergraph Attention (HA) mechanism. A Distribution-aware Pixel-wise Wasserstein Loss (DPWL) jointly enforces scale, positional, and distributional consistency. First, MSCD aggregates cross-layer representations via a multi-receptive-field convolutional structures to keep fine-graind target details and amplifies saliency. Second, HA exploits learnable soft hyperedges to capture high-order dependencies across channel and spatial dimensions. Third, DPWL injects a distribution-consistency constraint into the optimization process to improve convergence robustness and localization accuracy. Experiments on the IRSTD-1k, NUDT-SIRST, DenseSIRST, and NUAA-SIRST dataset show that MSC-HA provides an average performance of $75.89\%$ in IoU, $96.70\%$ in Pd, and $8.67 \times 10^{-6}$ in Fa. 
\end{enumerate}

\subsection{Adaptive and Multi-Cue Mechanisms}
This section reviews adaptive mechanisms as well as multi-scales mechanisms.

\begin{table*}
\begin{center}
\scalebox{1.0}{
\begin{tabular}{|l|l|}
\hline
\scriptsize{Methods} &\scriptsize{Acronyms}  \\
\hline
\hline
\scriptsize{Adaptive Mechanims}  &\scriptsize{TAMNet (Wang et al. \cite{DL-299}), MDAFNet [Dense] (Li et al. \cite{DL-516}), AER-net (Zhang et al. \cite{DL-542})} \\
\scriptsize{}  &\scriptsize{CEFANet (Fang et al. \cite{DL-560}), AMID (Yang et al. \cite{Applications-S-50}), ISTD-APSNet (Li et al. \cite{DL-599-20})} \\
\hline
\scriptsize{Recurrent Mechanims}  &\scriptsize{RRCANet (Liu et al. \cite{DL-97})} \\
\hline
\scriptsize{Diffusion Mechanisms} &\scriptsize{EDPrior (DPNet) (Tong et al. \cite{DL-147}), APCD (Rong et al. \cite{DL-162}), DCFR-Net (Fan et al. \cite{DL-234})} \\
\scriptsize{} &\scriptsize{DCDNet (Li et al. \cite{DL-599-40})} \\
\hline
\scriptsize{Multi-Branch Mechanisms}  &\scriptsize{DBCE (U-Net) (Xiao et al. \cite{DL-92}), MMLNet (Li et al. \cite{DL-114}), AFDMNet (Wang et al. \cite{DL-598})  } \\
\scriptsize{}  &\scriptsize{D2FENet (Xiao et al. \cite{DL-243}), DB-DCNet [Dense] (Xu et al. \cite{DL-521}), DFCHNet (Guo et al. \cite{DL-599-42})} \\
\scriptsize{}  &\scriptsize{TMNet (Zhang et al. \cite{DL-599-63}),  DBGNet (Chi et al. \cite{DL-599-65}), TADNet (Li et al. \cite{DL-599-231})} \\
\scriptsize{}  &\scriptsize{AEDNet (Zhang et al. \cite{DL-599-270})} \\
\hline
\scriptsize{Multi-Scales/Multi-Levels Mechanisms}  &\scriptsize{MSHNet (U-Net) (Liu et al. \cite{DL-41}), DBMSTN (ConvLSTM) (Li et al. \cite{DL-63}), HDNet (Xu et al. \cite{DL-100})} \\
\scriptsize{}  &\scriptsize{M4Net (Zhang et al. \cite{DL-64}), WGFFNet (Wang et al. \cite{DL-91}), PAMNet (Xu et al. \cite{DL-599-274})} \\
\scriptsize{}  &\scriptsize{PConv (Yang et al. \cite{DL-104}), Asymmetric PConv (ACEPNet)  (Li et al. \cite{DL-597})} \\
\scriptsize{}  &\scriptsize{AeFusionNet (Yang et al. \cite{DL-106}), SDS-Net (Yue et al. \cite{DL-233})} \\
\scriptsize{}  &\scriptsize{SEF-DETR [Detection] (Liu et al. \cite{DL-145}), IRST-RTDETR [Detection] (Chen et al. \cite{DL-160}), AGMFNet (He et al. \cite{DL-165}) } \\ 
\scriptsize{}  &\scriptsize{HPFNet (Zhang et al. \cite{DL-168}), SAFARI-net (Zhang et al. \cite{DL-191}), MEETNet (Zhao et al. \cite{DL-200}) } \\ 
\scriptsize{}  &\scriptsize{ECMSFNet [Real-time] (Jin et al. \cite{DL-210}), HNU-Net (Zhu et al. \cite{DL-230}), CSAFNet (Kong et al. \cite{DL-240})} \\ 
\scriptsize{}  &\scriptsize{MSCA-Net (Lu et al. \cite{DL-242}), MSMA-Net (Ma et al. \cite{DL-292}), DCGANet (Chen et al. \cite{DL-312}) } \\ 
\scriptsize{}  &\scriptsize{MLIGNet (Zhao et al. \cite{DL-314}), MFAGNet (He et al. \cite{DL-318}), FGOM-RTDETR [Detection](Wang et al. \cite{Applications-S-33})} \\
\scriptsize{}  &\scriptsize{MSSLNet (Guo et al. \cite{DL-330}), MFEFNet (Zhang et al. \cite{DL-341}), MSDA-Net (Zhao et al. \cite{DL-361})} \\
\scriptsize{}  &\scriptsize{MSFE-Net [Detection] (Li et al. \cite{DL-376}), STDMANet [MIRST] (Yan et al. \cite{DL-505})} \\
\scriptsize{}  &\scriptsize{MISTNet [MIRST] (Gao et al. \cite{DL-563}), MAR-UNet (Li et al. \cite{DL-577}), MPCNet (Zhang et al. \cite{DL-599-7}) } \\
\scriptsize{}  &\scriptsize{ISmallNet (Hu et al. \cite{DL-599-68}), MLCL-Net (Chuang et al. \cite{DL-599-77}), PMFFNet  (Fang et al. \cite{DL-599-92})} \\
\scriptsize{}  &\scriptsize{MCGC (Kou et al. \cite{DL-599-116}), MMIDNet (Li et al. \cite{DL-599-141})} \\
\scriptsize{}  &\scriptsize{MEFA-Net (Ma et al. \cite{DL-599-160}), MSAFFNet (Tong et al. \cite{DL-599-161}), IR-ADMDet (Li and Wei \cite{DL-599-165}), MEAC (Hu et al. \cite{DL-599-166})} \\
\scriptsize{}  &\scriptsize{MSFANet (Jing et al. \cite{DL-599-195}), EAMCNet (Ma  and Li \cite{DL-599-197}), BiMA  (Deng et al. \cite{DL-599-205})} \\
\scriptsize{}  &\scriptsize{DMSnet (Hu et al. \cite{DL-599-206}), BPR-Net (Du et aL \cite{DL-599-218}} \\
\scriptsize{}  &\scriptsize{DCFNet (Du et al. \cite{DL-599-237}, FCSF-Net (Liu et al. \cite{DL-599-263}), SMENet (Shang et al. \cite{DL-599-268})} \\
\scriptsize{}  &\scriptsize{CG-IRNet (Zhu and Liao \cite{DL-599-269}), SMENet (Zhao et al. \cite{DL-599-287}), DAMA-Net  (Zhang et al. \cite{DL-599-297})} \\
\scriptsize{}  &\scriptsize{CDMS-Net (Wu et al. \cite{DL-599-336}), RDANet (Liu et al. \cite{DL-599-333})}\\
\scriptsize{}  &\scriptsize{MCATNet (Li et al. \cite{DL-599-441})}\\
\hline
\scriptsize{}  &\scriptsize{MoDe [MIRST] (Luo et al. \cite{DL-599-364}), MCANet (Chen et al. \cite{DL-599-390})} \\
\scriptsize{Microscale Mechanisms}  &\scriptsize{HMNet (Hu et al. \cite{DL-599-131})} \\
\hline
\end{tabular}}
\caption{Adaptive and Multi-Cues Mechanisms: An Overview (79 papers).}
\label{OverviewAMCM}
\end{center}
\end{table*}

\begin{enumerate}
\item \textbf{Adaptive Mechanisms:} In 2024, Li et al.  \cite{DL-516}
design Moderately Dense Adaptive Feature Fusion Network (MDAFNet) which consists of three feature layers as the backbone. This module allows to connect all the internal features with each other and employs a weighted sum of different layers as the output allowing feature reuse and keeping infrared small target features in the deep layers. A Coarse-to-Fine detection Head (CFHead) is employed and an auxiliary loss allows the network to estimate target contour. Experiments show that MDAFNet provides similar performance than previous SOTA algorithms with requirements for real-time detection on RTX3080 GPU. In an other work, Zhang et al. \cite{DL-542} propose an Adaptive feature Enhancement and hierarchical Refinement network (AER-Net\footnote{https://github.com/fuqingzhang/AER-Net}) which extracts discriminative features of small targets while exploiting the semantics of different layers. A U-shaped architecture with a Feature Representation Enhancement Module (FREM) combines the local correlation of (CNNs) and the global correlation of Transformers to improve features while preventing the loss of tiny targets. A Daisy-shaped Correlation-aware Module (DCM)  capture global contexts for mitigating clutter interference in deeper layers. A Gated Refinement Module (GRM) refine low-level features from the encoder using high-level features as guidance for capturing local details and  removing noise in shallower layers. Experiments on three public datasets show that  AER-Net offers significant improvements in the mean intersection over union (IoU) metric. \\

In 2026, Wang et al. \cite{DL-299} design a Triple Adaptive Multiplexing Network (TAMNet\footnote{https://github.com/RuWang-prog/TAMNet}) which used a dedicated adaptive multiplexing strategies for feature extraction, contextual understanding, and prediction generation, respectively. First, a Parallel Attention Residual Block (PARB) mitigates feature distortion via an attention decoupling and keeps feature diversity by adaptively integrating feature streams. Second, a Local Saliency Module (LSM)  implicitly infer stable target-background relationships from multi-scale contextual dependencies. Third, the prediction mask is generated  via a multi-level feature fusion. Experiments show that  TAMNet offers SOTA performance in addition of generalization capability on public IRSTD datasets. \\
\item \textbf{Recurrent Mechanisms:} In 2025, Liu et al. \cite{DL-97} design a Recurrent Reusable-Convolution Attention Network (RRCA-Net) which incorporates Reusable-Convolution Block (RuCB) in a recurrent manner without introducing extra parameters. Thus, the high-level information of small targets in the deep layers is maintained and further refined. A Dual Interactive Attention Aggregation Module (DIAAM) allows to promote the mutual enhancement and fusion of refined information. Thus, RRCA-Net achieves high-level feature refinement and improves the correlation of contextual information between adjacent layers. A target characteristic inspired loss function (DpT-k loss) integrates physical and mathematical constraints. Experiments on NUAA-SIRST, IRSTD-1k and DenseSIRST) datasets show that RRCA-Net outperforms previous SOTA algorithms while keeping a small number of parameters. In addition RRCA-Net can be employ as a plug and play module to introduce consistent performance improvement in existing  IRSTD methods. \\ 
\item \textbf{Diffusion Mechanisms:} In 2024, Fan et al. \cite{DL-234} propose a Diffusion-based Continuous Feature Representation Network (DCFR-Net) which consists of two branches: Diffusion-based Continuous HR Feature Representation (DCHFR) and ISDTD. First, DCHFR employs Implicit Neural Representation (INR) into a conditional denoising diffusion model to capture extremely small target contours. Second, ISDTD exploits the shared encoder from DCHFR to obtain the HR feature representation, which is fed into Multiscale Implicit Feature Alignment (MIFA) and Spatial-Frequency Feature Interaction (SFFI). MIFA aggregates different-layer features in a resolution-free manner. SFFI provides profound SFFI and global-local receptive field mixture to increase the contrast between infrared targets and intricate backgrounds. Experiments on the NUAA-SIRST, IRSTD-1k, and NUDT-SIRST datasets show that  DCFR-Net outperforms previous algorithms. \\

In 2026, Tong et al. \cite{DL-147} propose an Energy Diffusion Prior (EDPrior) method. A feature recovery network retains the attention of small targets in deep networks by employing a simple encoding space and a multi-stage decoding space. This scheme reduces redundant skip connections. The curvature separation module exploits the thermal radiation diffusion properties of small targets as prior knowledge and extracts the target features via a learnable discrete Fourier transform. By reducing redundancy and incorporating prior knowledge, EDPrior learns effective target features for the detection task more efficiently and purposefully, rather than relying on complex networks to learn the variable backgrounds in different scenes. In an other approach, Rong et al. \cite{DL-162} design an end-to-end infrared background prediction method based on a Conditional Diffusion model with an Adaptive blocking (APCD) strategy. The adaptive blocking strategy uses both local and global information  and accelerates the inference speed of the diffusion model. The multi-scale attention segmentation module allows to suppress background clutter and enhance the target. Experiments show that APCD outperforms PSTNN \cite{SFD-RPCATensor-5}, RIPT \cite{SFD-RPCATensor-3}, ALCNet \cite{DL-11} and AGPCNet. \\
\item \textbf{Multi-Branch Mechanisms:}  In 2024, \textit{in the context of dense target detection}, Xu et al. \cite{DL-521}
design a Dual-Branch Dense Collaborative Network (DB-DCNet) which employs two dedicated detection branches. The first branch minimizes missed detections via target-sensitive feature learning. The second branch reduces false alarms through clutter-aware optimization techniques. Both branches adopt the same densely connected network structure but employ different feature extraction modules. The first branch exploits small-sized kernels and hybrid attention mechanisms to segment target edges. The second branch employs convolutions with multiple receptive fields and residual connections to filter out background clutter. The detection results of the two branches are processed by a lightweight feature selection network for feature selection and fusion. Experiments on the  NUDT-SIRST, IRSTD-Air, IRSTD-1k, and WideIRSTD-Full datasets show that  DB-DCNet provide better results than previous SOTA algorithms.  \\

In 2025, Xiao et al. \cite{DL-92} design a  Dual-Branch Contrast-Enhanced U-Net (DBCE U-Net) which employs  a Contrast Conv Module. The Contrast Conv Module improves contrast and feature representation through adaptive feature segmentation and convolution operations. The dual-branch combines ResNeSt modules for deep feature extraction and feature enhancement modules for  visual perception augmentation. DBCE U-Net exploits gradient information to improve small target features and increase the robustness via a dual-branch structure. Experiments on the NUDT-SIRST dataset show that DBCE U-Net offers superior detection performance  with an IoU and Pd of $94.60\%$ and $99.05\%$, respectively. In an other work, Li et al. \cite{DL-114} design a Multi-branch Mutual-guiding Learning Network (MMLNet\footnote{ https://github.com/qianngli/MMLNet}) which consists of three branches: edge, positioning, and detection. Each of branch consists of a module from a unique perspective. In the detection branch, a multidimensional lossless encoder optimized via a downsampling strategy and multilevel feature fusion allow to mitigate feature loss. In the positioning branch, a target positioning strategy identifies candidate targets from the image by means of a learnable multi-kernel pattern. In the edge branch, a conventional architecture help the model to capture the target shape. A mutual-guiding fusion module adjust information within and between branches. Experiments  show that MMLNet offers similar performance than previous SOTA algorithms. Xiao et al. \cite{DL-243} develop a Dual-Domain Feature Enhanced Network (D2FENet) which synergistically integrates wavelet transform with CNN operations. It allows to extract feature in both frequency and spatial domains via parallel processing pathways. The frequency-domain branch exploits wavelet transform to capture high-frequency details. The spatial branch uses pixel reorganization for feature fidelity downsampling. Experiments on the NUDT-SIRST dataset show that D2FENet preserves target integrity, suppresses false alarms, and offers superior performance compared to previous SOTA methods. \\

\item \textbf{Multi-Scales/Multi-Levels Mechanisms:} In 2023,  Ma et al. \cite{DL-292} present a Multiscale Super-Resolution Enhancement and Multilevel Attention Fusion Network (MSMA-Net).  First, MSMA-Net  determines the rough location of targets through a coarse preliminary screening. Second, the network employs multiscale super-resolution to transform low-resolution feature maps into high-resolution representations. Third, the network uses a multilevel feature fusion attention mechanism to facilitate effective information transmission and fusion in multiscale and multilevel feature representations. Experiments show that MSMA-Net suppresses false alarms and provides better performance when the target present a small size and obscure feature information. In the context of MIRST, Yan et al. \cite{DL-505} develop a Spatio-Temporal Differential Multiscale Attention Network (STDMANet). First, STDMANet employs a temporal multiscale feature extractor to obtain spatiotemporal features from multiple time scales and then resorted them to the spatial multiscale feature refiner to improve the semantics of the features. Second, a mask-weighted heatmap loss allows to train the network with only center point annotations and balances missing detection and false alarm. Experiments show that STDMANet  improve the metric F1 score up to 0.9744, surpassing SOTA algorithms by 0.1682. \\

In 2024, Liu et al. \cite{DL-41} design  a Multi-Scale Head with an UNet architecture (MSHNet\footnote{https://github.com/ying-fu/MSHNet}) using a Scale and Location Sensitive (SLS) loss to each scale of the predictions. The SLS loss tackle the limitations of previous losses. First, a weight for the IoU loss based on target scales  help the detector to separate targets with different scales. Second, a penalty term based on the center points of targets help the detector to localize targets more precisely. MSHNet outperforms model-based methods (IPI \cite{SFD-RPCAMatrix-1}, NRAM \cite{SFD-RPCAMatrix-10}, RIPT, PSTNN \cite{SFD-RPCATensor-5}, MSLSTIPT \cite{SFD-ML-2}) and data-based methods (MDvsFA \cite{DL-15}, ALCNet \cite{DL-11}, ISNet \cite{DL-211}, DNA-Net \cite{DL-12}). \\

In 2025, Li et al. \cite{DL-63} propose a Dual Branch Multiscale Spatio-Temporal Network (DBMSTN). First, DBMSTN employs a multiscale spatial feature extraction module that exploits inter-frame difference and saliency feature to highlight small targets at different scales and suppress complex backgrounds. Second, a dual-branch spatio-temporal feature extraction module with improved gating unit in convolutional LSTM (ConvLSTM) improve the extraction of motion features to deal with their uncertainty. Third, a fusing module combines multilevel spatio-temporal features. A weighted mean squared error (MSE) loss function is employed with adjustable weights of positive and negative samples to address the data imbalance problem. Experiments show that DBMSTN achieves a F1 score.
In an other work, Xu et al. \cite{DL-100} develop an Hybrid-Domain Network (HDNet\footnote{https://github.com/xumingzhu989/HDNet-TGRS}) with multiscale high-frequency information enhancement. More specifically, HDNet combines frequency-domain features with spatial-domain CNN features to increase the target-background contrast and suppress background interference. HDNet consists of two branches: the spatial-domain branch and the frequency-domain branch. In the spatial domain,  a Multiscale Atrous Contrast (MAC) convolution module improves the perception of small. In the frequency domain, a Dynamic High-Pass Filter (DHPF) module computes low-frequency signal energy and dynamically removes specific low-frequency information to capture high-frequency image details. Experiments on the IRSTD-1K, NUAA-SIRST, and NUDT-SIRST datasets show that HDNet outperforms SOTA methods. Zhang et al. \cite{DL-64} provide a Multi-level Multi-patch Multi-receptive Multi-dimensional attention Network (M4Net) to capture information interaction among high-level and low-level features. First, a Multi-level feature extraction module (MFEM) with multilayer vision transformer (ViT) allows to fuse multi-scale features. Second, a Multi-Patch Attention Module (MPAM) and a Multi-Receptive Field Module (MRFM) capture and improve the feature information. Third, a Multi-Dimension Interactive Module (MDIM) connect the attention mechanism on multiscale features to improve the learning ability. Experiments show that M4Net provides better results than RDIAN \cite{DL-33}, DNA-Net \cite{DL-12} and UIU-Net \cite{DL-220}. Wang et al. \cite{DL-91}) design a Wide-scale Gated Fully Fusion Network (WGFFNet) which employs a classic encoder–decoder structure. First, a Stepped Fusion Block (SFB) embedded in the feature extraction stage allows to capture finer local context across multiple scales during encoding. Along the decoding path, the multi-level features are progressively integrated by a Fully Gated Interaction (FGI) Module. A boundary difference loss optimizes the edge details of targets. Experiments on the SIRST-V2 \cite{Dataset-200} and IRSTD-1k show that WGFFNet outperforms NRAM \cite{SFD-RPCAMatrix-10}, PSTNN \cite{SFD-RPCATensor-5}, ALCNet \cite{DL-11}, DNA-Net \cite{DL-12}, ISNet \cite{DL-211}, UIU-Net \cite{DL-220} and PBT \cite{DL-536}. \\

In 2025, Yang et al. \cite{DL-104} design a Pinwheel-shaped Convolution (PConv) and scale-based dynamic loss. PConv is employed instead of standard convolutions in the lower layers of the backbone network because it better aligns with the Gaussian-like spatial distribution of infrared small target, enhances feature extraction, expands the receptive field, and employs only a minimal increase in parameters. Even if several loss functions combine scale and location losses, they do not adequately account for the varying sensitivity of these losses across different target scales. To address this problem, a scale-based dynamic (SD) Loss dynamically adjusts the influence of scale and location losses based on target size. Employing PConv and SD Loss into previous SOTA algorithms, experiments on the IRSTD-1K and SIRST-UAVB dataset, show significant performance improvements. Yue et al. \cite{DL-233} propose a Shallow-Deep Synergistic detection Network (SDS-Net\footnote{https://github.com/PhysiLearn/SDS-Net}) which models multilevel feature representations to enhance both the detection accuracy and computational time. First, a dual-branch architecture  separately models the structural characteristics and semantic properties of features keeping shallow spatial details while capturing deep semantic representations. Second, SDSNet employs an adaptive feature fusion module to dynamically model cross-layer feature correlations. Experiments on the NUAA-SIRST, NUDT-SIRST, and IRSTD-1K dataset show that SDS-Net outperforms Top-Hat, WSLCM, IPI \cite{SFD-RPCAMatrix-1}, ACM \cite{DL-10}, ALCNet \cite{DL-11}, RDIAN \cite{DL-33}, ISTDU-Net \cite{DL-506}, IAANet  \cite{DL-336}, DNA-Net \cite{DL-12}, UIU-Net \cite{DL-220}, and SCTransNet \cite{DL-36}. \\

In 2025, Zhu et al. \cite{DL-230} design a lightweight Hierarchical nested U-shaped Network (HNU-Net) which improve the traditional dense nested structure, using multiple independent U-shaped sub-networks to reduce the semantic gap between the encoder and the decoder, and avoid the fine-grained information of small targets from being submerged in complex and densely nested structures. More specifically, HNU-Net consists of three modules: 1) an Hierarchical Nested Interaction Module (HNIM) which transmits  information between the encoder and the decoder, 2) a Multi-scale Residual Block (MSRB) which extracts the features of the input feature map at multiple scales, and 3) a Channel-wise Spatial Attention (CSA) which combines the information of the encoder and the decoder across features and captures fine-grained information through the multi-head mechanism. Experiments on the NUAA-SIRST, NUDT-SIRST, and WideIRSTD datasets show that HNU-Net reduces the parameter quantity by more than $20 \%$ while increasing the average intersection over union (IoU). In an other work, Lu et al. \cite{DL-242} develop a Multi-Scale Context Aggregation Network (MSCA-Net\footnote{https://github.com/YonghengSun1997/MSCA-Net})  which improves feature preservation and global-local fusion. MSCA-Net consists of three modules: the Multi-Scale Enhanced Dilated Attention Module (MSEDA), the Positional Convolutional Block Attention Module (PCBAM), and the Channel Aggregation Module (CAB). 
MSEDA captures contextual information across different receptive fields keeping fine-grained features. PCBAM enhances spatial understanding by modeling position-aware dependencies. CAB adaptively aggregates multi-level features across channels to highlight key information. Experiments show that MSCA-Net performs well for strong small target detection performance in complex backgrounds. \\

In 2025, Chen et al. \cite{DL-312} design a Dynamic Content-Guided Attention Multiscale Feature Aggregation Network (DCGANet). First, a Selective Variable Convolution (SVC) module  integrates the advantages of standard convolution, irregular deformable convolution, and multi-rate dilated convolution. SVC expands the receptive field and improves non-local features. Second, a two-stage content-guided attention module focuses to salient regions within the feature maps and determines whether these regions correspond to targets or background interference. Third, an Adaptive Dynamic Feature Fusion (ADFF) module allows DCGANet to adaptively integrate contextual features. Experiments demonstrate that DCGANet outperforms ALCNet \cite{DL-11}, AGPCNet \cite{DL-21}, SCTransNet \cite{DL-36}, SeRankDet \cite {DL-503} and DCGADet. \\
 
In 2026, \textit{in the category of detection based methods}, Liu et al. \cite{DL-145} propose SEF-DETR which refines query initialization and comprises three components: Frequency-guided Patch Screening (FPS), Dynamic Embedding Enhancement (DEE), and Reliability-Consistency-aware Fusion (RCF). FPS exploits the Fourier spectrum of local patches to obtain a target-relevant density map. DEE enhances multi-scale representations in a target-aware manner. RCF refines object queries by enforcing spatial-frequency consistency and reliability. Experiments demonstrate that SEF-DETR outperforms detection based method (PConv \cite{DL-104}, NS-FPN \cite{DL-304}) and segmentation based methods (ACM \cite{DL-10}, ISNet \cite{DL-211}, ACLNet, DNA-Net). Chen et al. \cite{DL-160} design a method named IRST-RTDETR\footnote{https://github.com/amumu77-lang/IRST-RTDETR}. First, a global Contrast Perception-Guided Edge-Adaptive Sharpening preprocessing technique (GCP-TEAS) integrates global contrast and local edge information to improve the boundaries of low-significance targets. Second, a Cross-Stage Edge Feature Enhancement Module (CSEFEM), a Multi-Scale Multi-Head Self-Attention Fusion module (MSMH-AIFI), and a Gated Edge Feature Fusion module (GEFC3) improve small target detail retention, edge representation, and robustness through edge enhancement, cross-scale modeling, and gated modulation, respectively. Experiments show that IRST-RTDETR achieves F1 score improvements over previous SOTA methods. In an other work, Li et al. \cite{DL-376} design a Multi-Scale Feature Enhancement Network (MSFE-Net). First, a Cascading Feature Fusion Module (CFFM) and a Weighted Dilated Convolutional Pyramid (WDCP) improve shallow texture and deep semantic features, respectively. Second, a Weighted Feature Fusion Enhancement Module (WFFEM) learns differential and combined features across multiple branches. Finally, a Multi-scale Feature Stairstep-upsampling Fusion Module (MFSFM) refines high-resolution texture and semantic features. Experiments on the NWPU VHR-10 dataset show that MSFE-Net provides $92.8\%$ in mAP50 and $62.6\%$ in mAP75 outperforming methods based on YOLOv6 and YOLOv7. \\

In 2026, \textit{in the category of segmentation based methods}, He et al. \cite{DL-165} design an Attention-Guided Multi-scale Feature fusion Network (AGMFNet). First, a Multi-Scale Receptive Field Expansion module (MRFE) structure allows to  integrate non-local context information and broaden the acceptance domain. Second, a dual-branch weighting mechanism increases the performance of the model and reduces the complexity of the model through a dynamic modification of the channel weights. Third, a three-Level Asymmetric Cross layer Fusion (TACF) mechanism allows to achieve the effective integration of the low-level, middle-level and high-level features, and ensure that the model can not only capture the details but also the semantics. The loss function  mitigate the imbalance between the background and the target of an infrared image. Experiments show that AGMFNet outperforms ISTDU-Net \cite{DL-506}, MTU-Net \cite{Applications-S-41}, AGPCNet, DNA-Net \cite{DL-12}, UIU-Net \cite{DL-220}, DMFNet \cite{DL-599-261}, and TODENet \cite{DL-543}. Zhang et al. \cite{DL-168} develop a Hierarchical Perception Fusion Network (HPFNet). First, a Patch-Wise Context Feature Extraction module (PCFE) employs a Patch Nonlocal Block, convolutional blocks and attention mechanism to capture global–local feature extraction and enhancement Second, a Multi-Level Sparse Cross-Fusion module (MSCF)  performs cross-level feature interaction between encoder and decoder representations allowing fusion between low-level spatial details and high-level semantic cues. A dual Top-K sparsification mechanism filters irrelevant background features allowing the attention mechanism to focus more on the target region. Third, an Efficient Upsampling Module (EUM) combines upsampling with multi-branch dilated convolutions improve feature reconstruction and improve localization accuracy. Experiments show that HPFNet provides similar performance than ISNet and UIU-Net suppresing occasional false alarms and missed detections under challenging conditions. Zhang et al. \cite{DL-191} design a scale-adaptive framework named SAFARI-Net which consists of a scale-adaptive processing pipeline dynamically adjusting strategies based on estimated target characteristics, a frequency-aware feature extraction module with learnable parameters improving target saliency across spectral components, and a progressive memory refinement mechanism keeping representations. A triple-path encoder processes spatial, frequency, and contextual information vie dedicated pathways with bidirectional refinement for comprehensive enhancement. An adaptive clutter suppression mechanism modulates response based on scale predictions. Experiments show that SAFARI-Net achieves $95.77\%$ IoU on the NUDT-SIRST dataset and $70.17\%$ on the IRSTD-1 dataset with improvements for micro-scale targets. Zhao et al. \cite{DL-200} propose a Morphology-Edge Enhanced Triple-cascaded Network (MEETNet) which uses a triple-cascaded architecture that keeps high resolution and improves information interaction between different stages. MEETNet employs an Edge-Detail Enhanced Module (EDEM) and a Detail-Aware Multi-Scale Fusion Module (DMSFM). These modules provide edge-detail enhanced features related to contrast and edge information enhancing target saliency and edge representation. DMSFM employs a multireceptive field mechanism to merge target details with deep semantic insights. Experiments on the NUAA-SIRST and NUDT-SIRST datasets show that MEETNet outperforms previous SOTA  methods. \\ 

In 2026, \textit{in the category of real-time algorithm}, Jin et al. \cite{DL-210}
design an Efficient Convolution and Multiscale Feature Fusion Network (ECMSFNet\footnote{https://github.com/liubiaohua/ECMSFNet}) based on convolution and attention-guided Multiscale Feature Weighting and Fusion (MSFWF). First, a Dual-Branch Hybrid Convolution (DBHConv) module extracts infrared small target features. Second a Three-Branch Attention-Guided (TBAG) module improves the input features from both spatial and channel dimensions. Third, a MSFWF module is employed at the added detection head to adaptively weight the features and optimize the feature propagation path. Experimental show that ECMSFNet outperforms previous approaches with a real-time detection speed of 74.63 frames/s. In an other work, Kong et al. \cite{DL-240} propose a Context-Aware and Semantic-Guided Adaptive Filtering Network (CSAFNet\footnote{https://github.com/LingchuanK/CSAFNet}). First, a Parallel Self-Awareness Attention (PSAA) exploits physical priors to capture global context and incorporates wavelet transforms to strengthen local detail. Second, a Cross-Semantic Adaptive Filtering Module (CAFM) in feature fusion deeply explores information from shallow features and improves the relative saliency of target representations.  Third, a Dynamic Multiscale Spatial Pyramid (DMSSP) module enhances edge precision and segmentation accuracy. Experiments on the  NUAA-SIRST and IRSTD-1K datasets show that CSAFNet outperforms previous SOTA methods. In other work, Guo et al. \cite{DL-330} develop a Multiscale Sparse Large-Kernel Network (MSSLNet\footnote{https://github.com/xiaodacheng01/msslnet-main}). A Sparse Large-Kernel Convolution block (SLCB) employs a sparse-dense dual-branch design to boost target detection while keeping critical information flow. A Multiscale Sparse Large-kernel Block (MSSLB) model contextual information for target-background separation improved by a cross-gated module (CGM). A Bidirectional Guided Fusion (BGF) module optimizes cross-layer detail-semantic fusion within skip connections. MSSLNet is deployable with only 0.293 M parameters and 2.99 GFLOPs. Experiments on the NUAA-SIRST and IRSTD-1K datasets show that MSSLNet provides similar performance than SOTA algorithms in both accuracy and efficiency.  \\

In 2026, Zhao et al. \cite{DL-314} design a Multiscale Local Intensity and Gradient Network (MLIGNet) which embeds model-driven priors into a unified convolutional architecture. MLIGNet is based on the Intensity and Gradient Encoding Block which exploits morphological cues by encoding local intensity and gradient responses. A Crossing Fusion Module realizes semantic alignment between heterogeneous features. A Multiscale Fusion Module aggregates cross-scale information to increase robustness against variations in target size and signal-to-noise ratio. A lightweight dilated convolutional enlarge the receptive field without compromising spatial resolution. Experiments on the SIRST and NUDT-SIRST datasets show that MLIGNet offers a suitable balance between detection accuracy and computational efficiency. In an other work, He et al. \cite{DL-318} develop a Multi-scale Fusion and Attention Mechanism Guidance Network (MFAGNet). First, an Hybrid Dual Encoder (HDE) achieves feature-level denoising while extracting features. Second, a Multi-Scale Expansion Attention Module (MEAM) improves feature extraction ability by combining separable convolutions with channel attention mechanisms. Third, a Pixel-level Attention Module (PAM) allows optimization of features encoded by the encoder. Experiments on the ISDD dataset show  that MFAGNet achieves an mIoU of $74.12\%$, a Precision of $99.33\%$, and a reduced Fa of $0.0002477$. In an other work, Zhang et al. \cite{DL-341} propose a Multi-aspect Feature Extraction and Fusion Network (MFEFNet) which consist of three modules: a Multi-Shape Receptive Field Extraction Module (MSRFEM), a Spatial-Channel-Global-Local Joint Re-extraction Module (SCGLJRM), and a Multi-Scale Cross-Attention Mechanism (MSCAM). MSRFEM exploits dilated deformable convolutions and the Res2Net architecture to capture multi-scale and multi-shape target characteristics as well as locally correlated features. SCGLJRM improves global context modeling across spatial and channel dimensions by aggregating complementary local-global information with a cascade of Local spatial-Local channel (LL), local spatial-Global Channel (GC), and global spatial-Local Channel (LC) sub-modules. MSCAM mitigates the feature fusion imbalance in Transformer-based cross-attention by employing correlation matrix transposition and fusion before integrating with feature maps. Experiments on the SIRST, IRSTD-1K, and NUDT-SIRST datasets show that MFEFNet provides superior detection accuracy compared to several SOTA methods. \\

In 2026, \textit{in the context of ship detection}, Wang et al. \cite{Applications-S-33} design  a Far-shore-Guided Object-focusing Multiscale network with real-time detection Transformer (FGOM-RTDETR) which use multiscale local and global features. A Feature Grouping Module (FGOM) improves multiscale representation and comprises three components: a Feature Reparameterization (FRep) module, a Coordinate Attention Golden Feature Pyramid Network (CAGoldFPN), and a Multiscale Stacked Network (MuSSNet). FRep addresses the loss of channel information due to the small size of thermal infrared ship targets and the complexity of background features. CAGoldFPN enhances multiscale feature fusion.  MuSSNet mitigates issues of high target similarity and the susceptibility of small targets to being overlooked. Experiments show that FGOM-RTDETR provides performance gains: precision increases from 0.921 to 0.942, mAP50 increases from 0.938 to 0.958, and recall increases from 0.923 to 0.934. Zhao et al. \cite{DL-361} propose a Multi-Scale Direction-Aware Network (MSDA-Net\footnote{https://github.com/YuChuang1205/MSDA-Net}) which integrates the high-frequency directional features  as domain prior knowledge into neural networks. First, a High-Frequency Direction Injection (HFDI) module without trainable parameters  injects the high-frequency directional information of the original image into the network. Second, a Multi-Scale Direction-Aware (MSDA) module promotes the full extraction of local relations at different scales and the full perception of key features in different directions. Third, a Feature Aggregation (FA) structure handle target disappearance in high-level feature maps. Finally, Feature Calibration Fusion (FCF) module avoids feature bias during cross-layer feature fusion. Experiments show that  MSDA-Net outperforms RPCA-Net, ALCNet \cite{DL-11}, DNA-Net \cite{DL-12}, UIU-Net \cite{DL-220}, ISNet \cite{DL-211}, ILNet, MSHNet \cite{DL-41}, MMLNet and SDSNet. \\

In 2026, Li et al. \cite{DL-577} design a Multi-scale Attention Residual network (MAR-UNet\footnote{https://github.com/DearAida/MARUnet}) which is based on a Hierarchical Contextual Enhancement Aggregation (HCEA). It adaptively enhances target features. First, a Multi-Granularity Feature Extraction (MGFE) module integrates standard convolutions, Centered Differential Convolutions (CDC), and Dilated Convolutions (DC) to capture both local high-frequency gradients and long-range contextual information. Second, a Linear Cross-Channel Attention (LCCA) module models global channel dependencies with linear complexity  and dynamically adjust attention weights. Third, a  convolutional FeedForward Network (FFN) with $3 \times 3$ convolutions refines feature boundaries. Experiments on the NUAA-SIRST, NUDT-SIRST, and IRSTD-1K show that MAR-UNet outperforms 
ACM \cite{DL-10}, ALCNet \cite{DL-11}, RDIAN \cite{DL-33}, FC3-Net \cite{DL-335}, DNA-Net \cite{DL-12}, UIU-Net \cite{DL-220}, IRSAM \cite{DL-599-12}, L2SKNet \cite{DL-370} and SCTransNet \cite{DL-36}.
\end{enumerate}

\subsection{Guidance Mechanisms}
The existing guidance mechanisms concern features guidance, motion guidance, saliency guidance, memory guidance, feedback guidance, geometry guidance, ODE guidance and fuzzy-logic guidance. Table \ref{OverviewGuidanceM} shows an overview of these diverse guidance mechanisms.

\begin{table*}
\begin{center}
\scalebox{0.79}{
\begin{tabular}{|l|l|}
\hline
\scriptsize{Methods} &\scriptsize{Acronyms}  \\
\hline
\hline
\scriptsize{Causal Guidance}  &\scriptsize{CHAL (Duan et al. \cite{DL-599-192})} \\
\hline
\scriptsize{Prior Guidance}  &\scriptsize{PSGF (Zhang et al. \cite{DL-331}), BPGA (Lin et al. \cite{DL-534}), ECFNet (Fang et al. \cite{DL-599-246}), MEP-CSNet (Wang et al. \cite{DL-599-368})} \\
\scriptsize{} &\scriptsize{RHNet (Fang et al. \cite{DL-599-446})} \\
\scriptsize{} &\scriptsize{Dual Prior: DPMSNet (Wang et al. \cite{DL-599-421}), DPC (Yu et al. \cite{DL-599-433})  } \\
\scriptsize{} &\scriptsize{Gaussian Prior: GPConv [Detection] (Liu et al. \cite{DL-346}), PG-C2F (Cui et al. \cite{DL-399}), GA-Conv (Wei et al. \cite{DL-599-407}) } \\
\scriptsize{} &\scriptsize{Segmentation Prior: MGA-Net (Ding and Wu \cite{DL-599-18}), MISA [MIRST] (Wu et al. \cite{DL-599-315})} \\
\scriptsize{} &\scriptsize{Motion-Oriented Prior: MOPNet [MIRST] (Li et al. \cite{DL-599-19})}\\
\scriptsize{} &\scriptsize{RGB Prior Guidance: IRPNet (Yao et al. \cite{DL-599-41})} \\
\hline
\scriptsize{Features Guidance}  &\scriptsize{F3C-Net [Single-Frame] (Zhang et al. \cite{DL-335}), JFG (Xu et al. \cite{DL-37})} \\ 
\scriptsize{}  &\scriptsize{HMAMFU-net (Lee and Park \cite{DL-44}), PGDN-Net [Dense] (Liu et al. \cite{DL-61})} \\
\scriptsize{}  &\scriptsize{MDS-DWGANet (Liu et al. \cite{DL-132})} \\
\scriptsize{}  &\scriptsize{OIPF (Ma et al. \cite{DL-66}), STME [MIRST] (Peng et al. \cite{DL-72}), SGIRNet (Chi et al. \cite{DL-73})} \\
\scriptsize{}  &\scriptsize{AGDFE (Yang et al. \cite{DL-93}), DGGENet (Lu et al. \cite{DL-110}), SFLNet (Li et al. \cite{DL-144})} \\
\scriptsize{}  &\scriptsize{PQGNet (U-Shaped) (Liu et al. \cite{DL-161}), GCFLNet (Wang et al. \cite{DL-166}), PSHNet (Chen et al. \cite{DL-236})} \\
\scriptsize{}  &\scriptsize{DEFANet (Du et al. \cite{DL-194}), LGDNet (Guo et al. \cite{DL-221}) } \\
\scriptsize{}  &\scriptsize{SPIRE (Ni et al. \cite{DL-231}), ESENet (Yang and Cui \cite{DL-245})} \\
\scriptsize{}  &\scriptsize{Na-IRSTD (Xu et al. \cite{DL-561}), FEER-UNet [Ship] (Peng et al. \cite{Applications-S-43})} \\
\scriptsize{}  &\scriptsize{GIDNet (Gao et al. \cite{DL-599-15}), IRSTD-WFETAP (Xu et al. \cite{PSTD-50}), TDFGNet (Li et al. \cite{DL-599-199}) } \\
\scriptsize{}  &\scriptsize{RFAN (Liu et al. \cite{DL-599-204}), CMNet (Li et al. \cite{DL-599-211})} \\
\scriptsize{}  &\scriptsize{AFE-Net (Wang et al. \cite{DL-599-212}), AIFENet (Zheng et al. \cite{DL-599-213})  } \\
\scriptsize{}  &\scriptsize{HFANet (Tan et al. \cite{DL-599-233}), EEA-Net  (Wang et al. \cite{DL-599-262}) } \\
\scriptsize{}  &\scriptsize{DECNet (Chen et al. \cite{DL-599-303}), ABFINet (Wang et al. \cite{DL-599-443}) } \\
\hline
\scriptsize{Frequency Guidance}  &\scriptsize{SFGA-Net (Wang and Sun \cite{DL-599-284}), FPFNet (Zhang et al. \cite{DL-599-286}), FSGANet (Xiao et al. \cite{DL-599-281}) } \\
\scriptsize{}  &\scriptsize{LFENet (Xiao and Hu\cite{DL-599-296}), SFC-INRNet (Xu and Liu \cite{DL-599-314}) } \\
\scriptsize{}  &\scriptsize{SFERA-Net (Wu et al. \cite{DL-599-345}), SFJR-Net (Liu et al. \cite{DL-599-346}) } \\
\scriptsize{}  &\scriptsize{FSDL-Net (Qi et al. \cite{DL-599-377}), HDPNet (Zhong et al. \cite{DL-599-392}, HiSGD (He et al. \cite{DL-599-399})} \\
\scriptsize{}  &\scriptsize{ER-Net (Dai et al. \cite{DL-599-420}), LCB-MSHNet (Pan et al. \cite{DL-599-442})} \\
\hline
\scriptsize{Semantic Guidance}  &\scriptsize{TSI-Net (Zhuang et al. \cite{DL-96}), TSNet  (Ji et al. \cite{DL-134})} \\
\scriptsize{}  &\scriptsize{SemDetNet (Wu et al. \cite{DL-137}), FSGNet  (Zhang et al. \cite{DL-213}), CCDNet (Liao an Yin \cite{DL-214})} \\
\scriptsize{}  &\scriptsize{HLSR-net (Ma et al. \cite{PSTD-34}), AuxDet (Shi et al. \cite{DL-310}), BAFE-Net (Xiao et al. \cite{PSTD-40}) } \\
\scriptsize{}  &\scriptsize{ET-DSC (Sang et al. \cite{DL-377}), SDG-ResNet [Dense] (Zhang et al. \cite{DL-501}), UCDNet (Xu et al. \cite{DL-519})  } \\
\scriptsize{}  &\scriptsize{PASSNet (Pan et al. \cite{DL-599-175}), SDAISTD (Chi et al. \cite{DL-599-198})} \\
\scriptsize{}  &\scriptsize{SUDA (Zhang et al. \cite{DL-599-306}), SGFR-Net (Tu et al. \cite{DL-599-321})} \\
\scriptsize{}  &\scriptsize{DFR-Net (Li et al. \cite{DL-599-332}), RSFNet (Tan et al. \cite{DL-599-320}) } \\
\scriptsize{}  &\scriptsize{PCFNet  (Zhang et al. \cite{DL-599-350}), SGUIU-Net (Luo et al. \cite{DL-599-374}) } \\
\scriptsize{}  &\scriptsize{STDF-Net (Ge et al. \cite{DL-599-397}) } \\
\hline
\scriptsize{Gradient Guidance}  &\scriptsize{STGBD-Net (Hu et al. \cite{DL-599-220})} \\
\scriptsize{Edge Guidance}  &\scriptsize{MLEDNet (Li et al. \cite{DL-599-121}), Weak-Edge-IRSTD (Wu et al. \cite{DL-599-434})} \\
\hline
\scriptsize{Shape Guidance}  &\scriptsize{BTE-ShapeNet (Zhang et al. \cite{DL-599-90}), BCR-Net (Li et al. \cite{DL-599-342})} \\
\hline 
\scriptsize{Target Guidance}  &\scriptsize{STIFNet (Xiang et al. \cite{DL-599-331})} \\
\hline
\scriptsize{Mask Guidance}  &\scriptsize{DICMG-Net (Yao et al. \cite{DL-599-79})} \\
\hline
\scriptsize{Motion Guidance}  &\scriptsize{MI-DETR [Detection][MIRST] (Liu et al. \cite{DL-169}), MIST-Net [MIRST](Zhang et al. \cite{DL-199}), MOCID [MIRST](Zhang et al. \cite{DL-271}) } \\
\scriptsize{}  &\scriptsize{S2MVP [MIRST] (Duan et al. \cite{DL-599-35}), DEMNet [MIRST] (He et al. \cite{DL-599-36}) } \\
\scriptsize{}  &\scriptsize{DMRNet [MIRST] (Zhang et al. \cite{DL-599-234}) }\\
\scriptsize{}  &\scriptsize{TCF-Net (Shi et al. \cite{DL-599-292}), IRD-Net  (Lu et al. \cite{DL-599-293})} \\
\scriptsize{}  &\scriptsize{MOCA [MIRST] (Wang et al. \cite{DL-599-340}), ATMNet [MIRST] (Yuan et al. \cite{DL-599-341})} \\
\scriptsize{}  &\scriptsize{MPR-Net  [MIRST] (Wan et al. \cite{DL-599-347})} \\
\hline
\scriptsize{Saliency Guidance}  &\scriptsize{L2SKNet [Single-Frame] (Wu et al. \cite{DL-370}), CSPENet (Deng et al. \cite{DL-591}), LSL-ISTD (Xi et al. \cite{DL-599-32}) } \\
\scriptsize{}  &\scriptsize{MSER-induced Saliency Measure (Li et al. \cite{DL-599-67}), STIDNet
(Zhang et al. \cite{DL-599-167}), STDRNet  (Pan et al. \cite{DL-599-300}) } \\
\scriptsize{}  &\scriptsize{SFCIMTL-Net (Zhang et al. \cite{DL-599-304}) } \\
\hline
\scriptsize{Physics-Informed Guidance}  &\scriptsize{PI-SAM (Li et al. \cite{DL-204}), NUDTNet  (Wu et al. \cite{DL-599-91}), ST-PINet (Fang et al. \cite{DL-599-295})} \\
\scriptsize{}  &\scriptsize{PICANet (Liu et al. \cite{DL-599-360}), PMFNet (Fang et al. \cite{DL-599-365})} \\
\hline
\scriptsize{Memory Guidance}  &\scriptsize{Tridos [Multi-Frame] [MIRST] (Duan et al. \cite{DL-511}), DCPNet (Zhang et al. \cite{DL-500}), MFCDet (Wang et al. \cite{DL-599-403})} \\
\hline
\scriptsize{Entropy Guidance} &\scriptsize{EACNet (Yin et al. \cite{DL-599-351})} \\
\hline
\scriptsize{Feedback Guidance} &\scriptsize{DFINet (Wu et al. \cite{DL-142}), FeedbackSTS-Det (Huang et al. \cite{DL-146}), ADGFNet (Han et al. \cite{DL-599-398})} \\
\hline
\scriptsize{Geometry Guidance}  &\scriptsize{GeoRankNet (Cai et al. \cite{DL-170}), RPR-Net (Gao et al. \cite{DL-197}), LoHGNet (Ma et al. \cite{DL-599-1}) } \\
\hline
\scriptsize{Ordinary Differential Equation Guidance}  &\scriptsize{OFSPNet (Xue et al. \cite{DL-272}), FreqODEs (Chen and Ye \cite{DL-290}), D3-Net (Sun et al. \cite{DL-599-366})} \\
\hline
\scriptsize{Fuzzy-Logic Guidance} &\scriptsize{FLINet (Zhang et al. \cite{DL-396})}   \\
\hline
\scriptsize{Multi-Domain Guidance}  &\scriptsize{MFMF-NET [MIRST] (Xie et al. \cite{DL-241}) } \\
\hline
\scriptsize{Perception Guidance}   &\scriptsize{S2CPNet (Fu et al. \cite{DL-298}), FAPNet (Wang et al. \cite{DL-599-273})} \\
\hline 
\scriptsize{Position Guidance}   &\scriptsize{GPNet (Zhang et al. \cite{DL-599-110})} \\
\hline 
\scriptsize{High Resolution Guidance}   &\scriptsize{H2D-Net (Zhang et al. \cite{DL-599-207})} \\
\hline
\end{tabular}}
\caption{Guidance Mechanisms: An Overview (119 papers).}
\label{OverviewGuidanceM}
\end{center}
\end{table*}

\begin{enumerate}
\item \textbf{Causal Guidance:} Most of the existing methods are target-centered by directly learning the target features from backgrounds. They are facing difficulties in effectively capturing stable features due to weak target signals. To address this problem, Duan et al.  \cite{DL-599-192} design a Causal-guided Hierarchical Anomaly-aware Learning (CHAL) framework. CHAL focuses on background learning and the targets are considered as the anomalies in backgrounds. First, a spatio-temporal neural field  models the background evolution patterns from generative perspective. Second, a hierarchical anomaly-aware learning decomposes anomaly discovery. Third, a causal-guiding mechanism allows to block the spurious correlations often caused by background confounders, and enhance true target causality. Experiments show that CHAL outperforms single-frame data-based models and multi-frame data-based models. \\
\item \textbf{Prior Guidance:} Small target is prone to feature attenuation during downsampling and false-alarm amplification during multilevel feature fusion. To address these issue, Zhang et al. \cite{DL-331} design in 2026 a dual-prior guided network with cross-scale calibration and Pixel-Selective Gated Dusion (PSGF). First, a depth-aware prior module integrates a Radial Flux Consistency Prior (RFCP) to suppress structural clutter. Second, a Local Directional Saliency Prior (LDSP) information enhance the isotropic small targets. Third, a multiscale feature interaction module aligns cross-scale representations through channel calibration and contextual aggregation to avoid cross-scale bias. The PSGF module adaptively balances shallow spatial details and deep semantic features to avoid background interference while keeping target integrity. Experiments on the SIRST, NUDT-SIRST, and IRSTD-1K datasets show that PSGF outperforms several previous methods. \\
\item \textbf{Features Guidance:} To address the low-contrast between small targets and noisy background in infrared images, Zhang et al. \cite{DL-335} design a FC3-Net\footnote{https://github.com/IPIC-Lab/SIRST-Detection-FC3-Net} by exploring feature compensation and cross-level correlation. More specifically, FC3-Net consists of a Fine-detail guided Multi-level Feature Compensation (F-MFC) module, and a Cross-level Feature Correlation (CFC) module. First, the F-MFC module compensates the information loss of details due to the downsampling layers in CNN by aggregating features from multiple adjacent levels. To address the background noise, the CFC module employs an energy filtering kernel based on the higher-level features with less background noise to filter out the noise in the middle-level features. Finally, it fuses them with the low-level ones to learn a strong target representation. Using them into the encoder-decoder structure, the FC3-Net produces a more accurate target mask with a better shape and details. \\ 

In 2024, Lee and Park \cite{DL-44} design a Hybrid Network based on Hierarchical Multipatch Feature Encoder (HMAMFU-Net\footnote{https://github.com/skylih87/HMAMFU-net}). The encoder  divides the image into multiple small pieces and stacks them hierarchically to extract features. HMAMFU-Net offers a hybrid encoder structure that combines a convolution-based multiscale encoder to extract local information and a hierarchical-based multipatch encoder to extract global information by running in parallel. Thee loss function fuses the hierarchical information obtained from each layer. Experiments on the NUDT-SIRST and NUAA-SIRST datasets show that HMAMFU-net outperforms several SOTA algorithms. Guo et al. \cite{DL-221} develop a Location-Guided Dense nested attention Network (LGDNet). LGDNet comprises a Pixel Attention Guided Feature Extraction Module (PAG-FEM), a Channel Attention Guided Feature Fusion Module (CAG-FFM), and a detection module.  PAG-FEM employs the DNIM dense nested blocks from the DNA-Net as the backbone, integrating both channel and pixel attention mechanisms. Second, CAG-FFM uses upsampling and convolution operations to align the feature sizes, while using the channel attention mechanism to obtain channel information. Third, these features are fused via stacking, addition, and averaging operations to obtain discriminative features. The detection module empoys eight-connected neighborhood clustering method to obtain the centroid coordinates of the targets for subsequent detection evaluation. Experiments show that LGDNet provides better results than previous SOTA methods. \\

In 2025, Chen et al. \cite{DL-236} develop a hybrid deep-learning framework that combines dense spatial heatmap supervision with geometry-aware regression (PSHNet). It generates position–scale heatmaps to guide coarse localization, which are refined through sub-pixel offset and size regression. A Complete IoU (CIoU) loss is employed as a geometric regularization term to enhance alignment between predicted and ground-truth bounding boxes. An Enhanced Low-level Feature Module (ELFM) is incorporated using multi-scale dilated convolutions and channel attention to preserve fine spatial details. Experiments on the NUDT-SIRST and IRSTD-1k datasets show that PSHNet outperforms DNA-Net \cite{DL-12} and ALCNet  \cite{DL-11} in IoU, detection probability, and false alarm rate. In an other work, Xu et al. \cite{PSTD-50} design an infrared small target detection method based on Weak Feature Enhancement and Target Adaptive Proliferation (IRSTD-WFETAP). First, a sparse sampling mechanism and hybrid filtering method capture the complex shapes and edge information of small targets while reducing the loss and shift of scarce features. Second, a multiscale feature enhancement module employs a vertical-horizontal bidirectional attention and a multiscale feature encoding to obtain stable feature interaction channels between the encoder and decoder. Third, a Target Data Self-Adaptive proliferation Strategy (DSAS) allows to address the imbalance of positive and negative samples for generalization and expression capability. A target-background joint loss  alleviate the imbalance issue and help the network converge smoothly. Experiments on the NUAA-SIRST and  IRSTD-1k datasets show that IRSTD-WFETAP provides better performance in nIoU, Pd, Fa, and F1-measure compared to previous methods. \\

In 2025, Liu et al. \cite{DL-61} propose a Prior Guided Network (PGDN-Net) which combines traditional manual features with a deep learning network model. First, three prior features are extracted namely the high-order Riesz transform feature, the compactness and heterogeneity feature (CH), and the corner feature of the structure tensor (ST). Second, these features are input into a dense nested network for guidance with a two-orientation attention aggregation module and a channel and spatial attention module. Each feature offers their respective guiding roles in different depths. Experiments show that PGDN-Net provides better performance in background suppression, target enhancement, probability of detection, and false alarm rate than previous algorithms. Liu et al. \cite{DL-132} develop a lightweight detection network named MDS-DWGANet which comprises a Multi-Directional Sparse Convolution module (MDSC) and a Dynamic Weight Guided Attention module (DWGA). First, MDSC constructs sparse convolution branches in four directions (horizontal, vertical, and two diagonals) to simulate directional gradient responses, and uses a sparse gating mechanism to focus on high-response edge regions, maintaining structural awareness while reducing parameter overhead. Second, a Dynamic Weight Guided Attention module (DWGA) mitigates feature dilution and attention drift by integrating four attention pathways with a dynamic channel allocation mechanism. Practically, DWGA can be considered as a deep generalization of regional contrast enhancement. Experiments on  the NUDT-SIRST and IRSTD-1 K datasets show that MDS-DWGANet provides $98.30\%$ in detection probability (Pd) and $93.04\%$ in mean Intersection over Union (mIoU) with $1.4$ M parameters. \\

In 2025, Ma et al. \cite{DL-66} design an Orthogonal Inputs Perception Fusion (OIPF) network to capture the learnable target features in deeper layers without increasing network depth. The orthogonal data pairs through intensity inversion and enhancement are then fed into a dual-input framework to provide incremental information flow. Second, a Relationally Aware Module (RAM) generates spatial weight maps by exploiting the relationships between data pairs across layers. Extensive experiments show that OIPF significantly improves ISTD performance. Yang et  al.\cite{DL-93} develop an Attention-Guided Dual Feature Extraction (AGDFE\footnote{ https://github.com/wwwyangxiaoyucom/new-network}) network which comprises two levels of feature extraction: a multi-scale feature extraction mechanism and a contextual feature extraction mechanism. First, the multi-scale feature extraction mechanism implements the feature extraction operation of the first layer. Thus, the information interaction between neighboring layers is done by fusing and splicing the features of different layers. Second, the target attention mechanism enhances the model’s attention to small targets and offers effective attention allocation for the detection of small infrared targets. Third, the multi-resolution feature interaction mechanism preserves the detail information and improves the ability to perceive small targets. The contextual feature extraction mechanism implements the second-level feature extraction operation background environments. Experiments show that the AGDFE provides $99.81\%$ vs. $87.58\%$ Precision, $98.96\%$ vs. $88.12\%$ Recall, $99.38\%$ vs. $87.85\%$ F-measure and $98.77\%$ vs. $78.33\%$ mIoU on NUDT-SIRST and SIRST Aug datasets, respectively. \\

In 2025, \textit{in the context of generalization in the presence of unseen scenes}, Chi et al. \cite{DL-73} develop a contrast-invariant feature extraction framework for Single-domain Generalization (SGIRNet). First, a Contrast Modulation module (CM) addresses the feature domain gap by interleaving enhancement and suppression operations. Second, a Position and Channel Invariant Feature extraction Module (PCIFM) extracts target features that are not affected by contrast changes. SGIRNet is integrated into several existing models to evaluate the performance. Experiments show that SGIRNet improves the detection performance of existing models in the unseen domain providing superior generalization ability. \\

In 2025, \textit{in the context of MIRST}, Peng et al. \cite{DL-72} develop a Spatio-Temporal feature fusion networks (STME\footnote{https://github.com/UESTC-nnLab/STME}) which comprises two parallel feature extraction branches. The fist one extracts temporal features across entire video clip. The second one isolates spatial features from a selected key frame. A self-refining module allows to adaptively adjust the extracted temporal features and fusing spatio-temporal features at different scales Extensive experiments show that STME reduces floating point operations per second (FLOPs). \\

In 2025, \textit{in the context fo plug and play component}, Lu et al. \cite{DL-110} design a Dual Geometry-Guided Experts Network (DGGENet) based on the Mixture-of-Experts (MoE) paradigm with a Target-Background Decoupling Module (TBDM). Specifically, TBDM uses a geometry-guided dynamic routing mechanism functioning as an adaptive gating network. Thus, it continuously analyzes input features to dynamically establish two specialized expert groups. The first one  is dedicated to target channels and the seoond one is dedicated to background channels. Each group functions as a sub-network specialized for its assigned feature subspace. Using the RPCA principles, the inter-group expert iteration promotes low-rank background and sparse target representations during collaborative feature refinement. Second; a cross fusion module acts as state feedback allowing semantically consistent interaction between the optimized representations. This closed-loop interaction explicitly models target-background correlations. Experiments show that DGGENet) provides superior performance on the NUDT-SIRST dataset and the IRSTD-1K dataset. TBDM is thus a versatile plug-and-play component which improved diverse ISTD frameworks. \\

In 2026, Liu et al. \cite{DL-161} design a Perceptual Query Guided Network (PQGNet\footnote{\url{https://github.com/PepperCS/PQGNet}}). First, a Perceptual Query Supervision Mechanism (PQSM) employs a perceptual loss to constrain spatial feature learning of each encoder layer. Second, a Perceptual Feature Construction Module (PFCM) captures perceptual features to preserve target localization information. Third, a Perceptual Query Quidance Module (PQGM) employs a cross-attention to guide global and regional feature queries through skip connections, optimizing target feature extraction. A max pooling-wavelet hybrid layer (MWHL) and high-frequency enhanced wavelet layer (HEWL) exploit discrete wavelet transform properties to improve deep semantic representations using shallow high-frequency details. Experiments on the NUDT-SIRST, NUAA-SIRST, and IRSTD-1K datasets demonstrate that PQGNet outperforms SOTA methods with a competitive balance between computational complexity and accuracy. 

In 2026, \textit{in the context of gradient features}, Wang et al. \cite{DL-166} propose a Gradient-Compensation-based Feature Learning Network (GCFLNet) based on a multi-module collaborative design to improve feature representation and fusion. First, an Edge Enhancement Module (EEM) captures fine-grained edge information while suppressing background noise  via smoothing operations. Thus, the extracted edge features are embedded into a Global–Local Feature Interaction (GLFI) module. A Multi-Scale Information Compensation (MSIC) module exploits the complementary characteristics of multi-scale features across spatial and channel dimensions, guiding efficient fusion of high-level and low-level information. Experiments on the NUDT and IRSTD-1K datasets show that GCFLNet outperforms filter-based approaches (Top-Hat, Max–Median), local contrast-based methods (TLLCM, WSLCM), model-based methods (RIPT, IPI), and data-based approaches (UIU-Net \cite{DL-220}, DNA-Net \cite{DL-12}, RDIAN \cite{DL-33}) in heterogeneous backgrounds. Du et al. \cite{DL-194} design a Dual-Path Edge-Guided Frequency-Aware Network (DEFANet) which allow edge-target collaborative modeling for enhanced feature representation. DEFANet features comprises a main branch for holistic target modeling and an edge branch for boundary transition perception. A two core modules: Frequency-Aware Dual Enhancement Module (FADE) and Edge-Guided Integration Module (EGI) allow interaction and representation in both branches. First, FADE uses a Frequency-Decoupled Attention Enhancement Mechanism to enhance both branches in the frequency domain, strengthening holistic modeling in the main branch and boundary representation in the edge branch. Second, EGI exploits a Dual-Path Group-Wise Guidance Mechanism to integrate edge features into the main branch. Experiments show that DEFANet provides similar performance than MDvsFA \cite{DL-15}, LAFT, SIRST, and SIATD. Yang and Cui \cite{DL-245} develop an Edge-Shape Enhanced Network (ESE-Net) which enhance edge feature representations to improve target discrimination.  First, a Multiscale Spatial Edge Attention (MSEA) module enhances target edges by perceiving directional gradient changes. Second, an Edge Guidance Module (EGM) extracts edge features in the frequency domain through a wavelet transform and performs reversible down sampling, discarding low-frequency components before fusing with spatial features. Third, a Multiscale Group Convolution Module (MGCM) is integrated in deep layers to preserve target details and mitigate the risk of small target loss. Experiments on the NUAA-SIRST and IRSTD-1K datasets show that ESE-Net outperforms UIU-Net \cite{DL-220}, DNA-Net \cite{DL-12}, and RDIAN \cite{DL-33}. \\

In 2026, Xu et al. \cite{DL-561} design a a Native-resolution feature extraction and fusion framework (Na-IRSTD) which employs native-resolution features to keep subtle target cues. An token reduction and selection strategy selects target patches with high accuracy and confidence, boosting the low-level details of the feature while effectively reducing native-resolution patch tokens compared to dense processing. Experiments show that Na-IRSTD outperforms models-based methods (NRAM \cite{SFD-RPCAMatrix-10}, PSTNN \cite{SFD-RPCATensor-5}) and data based methods (ACM \cite{DL-10}, DNA-Net \cite{DL-12}, ISNet \cite{DL-211}, UIU-Net \cite{DL-220}, RPCANet \cite{SFD-RPCAMatrix-46}, IRMamba, SAIST \cite{LLM-1}). Li et al. \cite{DL-144} propose a Spatial-Frequency Feature Learning Network (SFLNet) which attempt to solve the challenges in small target detection through collaborative modeling of the spatial and frequency domains. First, SFLNet adopts an edge-guided decoder structure that preserves the structural integrity and shape information of small targets during reconstruction. Second, this module employs a spatial–frequency joint learning mechanism that allows complementary and enhanced information exchange between the spatial and frequency domains. Experiments show that SFLNet outperforms previous SOTA algorithms in precise localization of targets and preservation of target shapes. \\
\item \textbf{Reconstruction Guidance:} In 2026, Yan et al. \cite{DL-599-179} design a target-aware invertible encoder (InvDet\footnote{https://github.com/SchulerYan/InvDet/tree/main}) that combines information preservation and target-aware enhancement within a reconstruction-guided detection framework. First, an invertible pathway builds the input from feature latents considering information loss as an optimizable quantity.  Second, a Target-Aware Reconstruction Modulation (TARM) module operates only in the inverse path.  Third, a Geometry–Content Tolerance Metric (GCTM)  focuses on truly informative regions and yields a pixel-wise weight map that  regularizes the reconstruction branch. 
\item \textbf{Semantic Guidance:} Most of deep neural networks methods suffers from semantic degradation when transferring high-level features from deep to shallow layers, limiting the precise localization of small targets. To address this problem, several researches propose to employ semantic guidance. In 2024, Ma et al. \cite{PSTD-34} develop a small target detection network based on High- and Low-Frequency Semantic Reconstruction (HLSR-net)  to suppress background clutter interference and to extract sparse features of targets. First, high-frequency target and low-frequency background features are extracted separately via fast Fourier frequency transformation. Second, target enhancement and background suppression are  carried out through the designed high- and low-frequency reconstruction networks, respectively. Third, a fusion output strategy provides the detection of small targets. Experiment show HLSR-net achieves better performance with an accuracy improvement of $6\%$. Xiao et al. \cite{PSTD-40}) design a  Background-Aware Feature Exchange Network (BAFE-Net\footnote{https://github.com/GrokCV/BAFE-Net}) which is a multi-task architecture that jointly addresses target detection and background semantic segmentation. BAFE-Net employs a dynamic cross-task feature hard-exchange mechanism allowing effective exchange of target and background semantics between the two tasks. Experiments on the  DenseSIRST dataset show that BAFE-Net outperforms ACM \cite{DL-10}, UIU-Net \cite{DL-220}, MTU-Net \cite{Applications-S-41}, DNA-Net \cite{DL-12}, RDIAN \cite{DL-33}, URAnet \cite{DL-504}, ABCNet \cite{DL-502} and SeRankDet \cite {DL-503}. Hu et al. \cite{Applications-S-42} design a scene semantic prior-assisted infrared ship detection using multitask perception network (SMPISD-MTPNet\footnote{https://github.com/greekinRoma/SMPISD-MTPNet}). This network is based on multitask perception: one task  predict targets, and the other focuses on scene perception to suppress false alarms caused by background interference. A Scene Semantic Extractor (SSE) guide the network according to the features extracted based on expert knowledge and the gradient-based module. A data augmentation is employed and a training trick named soft fine-tuning enhance the network’s generalization and suppress the distortion due to the augmentation process. Experiments on the IRSDSS datasets show that SMPISD-MTPNet outperforms previous SOTA methods. \\

In 2025, Zhuang et al. \cite{DL-96} develop a Temporal-Semantic Interaction Network (TSI-Net) for multi-frame infrared small target detection. TSI-Net employs a Temporal Feature Extraction Module (TFEM) which allows to capture the motion characteristics of targets in consecutive frames. To tackle insufficient feature interaction, a Feature Interaction-Aware Module (FIAM) based on a multi-scale semantic interaction Transformer allows robust cross-level feature fusion. Finally, a multi-task weighted hybrid loss function allows to balance classification accuracy, region alignment, and pixel-level boundary delineation. Experiments on the NUDT-MIRSTD and SIATD datasets show that TSI-Net outperforms ALCNet \cite{DL-11}, DNA-Net \cite{DL-12}, UIU-Net \cite{DL-220}, MSLSTIPT \cite{SFD-ML-2}, ASTTV-NTLA \cite{SFD-RPCATensor-32}, IMNN-LWEC \cite{SFD-RPCATensor-34} and STDMANet \cite{DL-505}. Shi et al. \cite{DL-310} propose an Auxiliary Metadata Driven Infrared Small Target Detector (AuxDet\footnote{https://github.com/GrokCV/AuxDet}) to incorporate metadata into the IRSTD paradigm for scene-aware optimization. Using a high-dimensional fusion module, AuxDet dynamically employs metadata semantics with visual features, guiding adaptive representation learning for each individual sample. A lightweight prior-initialized enhancement module using 1D convolutional blocks refine fused features and recover fine-grained target cues. Experiments on the WideIRSTD-Full dataset demonstrate that AuxDet outperforms ALCNet \cite{DL-11}. \\

In 2026, Ji et al. \cite{DL-134} propose a Three-Stage Networks (TSNet) using an encoder-decoder architecture with an upsampling-free subnetwork to fuse spatial and semantic information. The first two stages extracts semantic features viz a Patch-based Multi-scale Feature Enhancement Module (PMFEM). The final stage uses a sub-network without upsampling to avoid spatial information loss that occurs in traditional downsampling processes. This sub-network consists of a Cross-feature Fusion Block (CFB) combined with 2D Selective Scan (SS2D) to capture spatial information while employing high-level semantic features. Experiments on the SIRST, NUDT-SIRST, and IRSTD-1k datasets show that the TSNet outperforms ACM \cite{DL-10}, ALCNet \cite{DL-11}, DNA-Net \cite{DL-12}, UIU-Net \cite{DL-220}, TMNet, RDIAN \cite{DL-33} and AGPCNet \cite{DL-21}. In an other work, Wu et al. \cite{DL-137} design a Semantic-Detail Collaborative Network (SemDetNet). First, a background-aware global semantic module allows to enhance the discriminative representation of non-target regions through reverse compensation in multi-level downsampling. Second, a detail compensation module exploits shallow multi-granularity details. This module further exploits coarse-grained prediction masks to guide deep features, enabling semantic reconstruction and edge refinement. Experiments show that SemDetNet provides better results than HCFNet, MSHNet \cite{DL-41}, MTU-Net \cite{Applications-S-41} and DNA-Net \cite{DL-12}. In an other work, Zhang et al. \cite{DL-213}
propose a lightweight detection framework incorporating Frequency-aware and Semantic Guidance networks (FSGNet\footnote{https://github.com/Wangtao-Bao/FSGNet}). First, a multi-directional interactive attention module captures fine-grained and directional features. Second, a multi-scale frequency-aware module exploits Fast Fourier transform to preserve salient target structures. Third, a global pooling module captures high-level semantic information.Then, it is upsampled and propagated to each decoder stage via a global semantic guidance flows allowing semantic consistency and precise localization across scales. Experiments on the NUAASIRST, IRSTD-1K, NUDT-SIRST, and SIRSTAUG datasets show that outperforms ACM \cite{DL-10}, RDIAN \cite{DL-33}, AGPCNet \cite{DL-21}, DNA-Net \cite{DL-12}, UIU-Net \cite{DL-220}, RPCA-Net \cite{SFD-RPCAMatrix-64}, MSHNet \cite{DL-41} and L2SKNet \cite{DL-370}. Liao an Yin \cite{DL-214} design  Camouflage-aware Counter-Distraction Network (CCDNet). First, a backbone with Weighted Multi-branch Perceptrons (WMPs), aggregates self-conditioned multi-level features to represent the target and background. Second, a Aggregation-and-Refinement Fusion Neck (ARFN) to refine structures/semantics from shallow/deep features maps, and bidirectionally obtain the relations between the targets and the backgrounds. Third, a Contrastive-aided Distractor Discriminator (CaDD), enforces adaptive similarity computation both locally and globally between the real targets and the backgrounds to discriminate distractors, so as to reduce the false alarm rate. Sang et al. \cite{DL-377} propose an Edge-Target Deep Semantic Consistency (ET-DSC\footnote{ https://github.com/Yuweiw-1024/ET-DSC}) semantic adaptive balancing framework. In encoder, shallow-layer modeling and gated fusion enhance target boundaries. In decoder, semantic consistency constraints keep real boundaries and suppress false edges. A semantic allocation mechanism  between shallow and deep layers allows cooperative optimization between edge compensation and semantic preservation. Experiments on the NUDT-SIRST and NUAA-SIRST datasets show that ET-DSC outperforms ACM \cite{DL-10}, UNet++ \cite{MOD-Theory-202-1}, HCFNet, DNA-Net \cite{DL-12}, MTU-Net \cite{Applications-S-41} and MSHNet \cite{DL-41}. Zhang et al. \cite{DL-501} propose a Semantic Density-Guided ResNet (SDG-ResNet). It is a plug-in backbone that attaches a lightweight semantic density head to the deepest stage and injects the predicted density map into intermediate layers through Semantic Density-Guided Refine (SDGR) blocks with residual spatial
gating. \\
\item \textbf{Motion Guidance:} Multi-frame approaches learn motion implicitly through deep neural networks but often require additional motion supervision or explicit alignment modules  In 2026, Liu et al. \cite{DL-169} propose Motion Integration DETR (MI-DETR\footnote{https://github.com/nliu-25/MI-DETR}) bio-inspired with a dual-pathway detector which processes one infrared frame per time step while modeling motion. First, a Retina-inspired Cellular Automaton (RCA) converts raw frame sequences into a motion map defined on the same pixel grid as the appearance image. RPCA allows parvocellular-like appearance and magnocellular-like motion pathways to be supervised by a single set of bounding boxes without extra motion labels or alignment operations. Second, a Parvocellular-Magnocellular Interconnection (PMI) Block allows a bidirectional feature interaction between the two pathways giving a biologically motivated intermediate interconnection mechanism. Third, a RT-DETR decoder employs features from the two pathways to provide detection results. Experiments on the ITSDT-15K, IRDST-H, and DAUB-R datasets show that MI-DETR outperforms single frame detection methods (SANet \cite{DL-32}, AGPCNet \cite{DL-21}, RDIAN \cite{DL-33}, DNA-Net \cite{DL-12}, RPCA-Net \cite{SFD-RPCAMatrix-64}, L2SKNet \cite{DL-370}) and multi-frame detection methods (SSTNet \cite{DL-38}). Zhang et al. \cite{DL-199} design a Motion-guided Spatial–Temporal integrated Network (MIST-Net). First, the input strategy encodes the dominant motion characteristics across multiple consecutive frames using series expansion and temporal motion intensity analysis. This strategy allows to build a motion frame that enforces motion information from several frames into one. The motion frame and target frame are then fed into the integrated network to extract motion and spatial features, respectively. Second, a Mutual Calibration Module (MCM) refine the motion maps using spatial cues and improves the spatial maps with motion information. Third, a semantic gap between motion and spatial features is obtained by using a Motion-Spatial-Coupled selective state space Module (MSCM). MSCM dynamically adjusts the fusion of motion and spatial information. Experiments on SIRST and ITSDT-15K datasets show that MIST-Net outperforms previous SOTA algorithms. Zhang et al. \cite{DL-271} propose MOCID\footnote{https://github.com/TanzanOY/MOCID} which is a spatio-temporal backbone with Fourier-inspired Spatio-temporal Attention (FISTA) layers. Each FISTA layer sequentially processes the features from spatial and temporal views to capture clip-level temporal motion context. This context is then embedded into dynamic convolutional kernels for subsequent spatial feature extraction allowing motion difference guidance and generating features. For frame-level feature fusion, a Displacement-aware Mamba Module (DAM)  capture detailed frame-to-frame displacement information. DAM employs a Temporal Interpolation and Displacement-aware Scan method to perform spatio-temporal difference-aware displacement modeling. Thus, MOCID capture motion and displacement contexts, Experiments on the IRDST and DAUB datasets how that MOCID provides a superior balance between throughput and performance compared to other methods. \\
\item \textbf{Saliency Guidance:} Most of the conventional deep neural networks struggle to perceive the sensitivity of small targets and adopt heavy constructions to preserve potential target features ignoring domain-specific insights and suffering from poor explainability. In 2025, Wu et al. \cite{DL-370} attempt to address this problem by revisiting the saliency principles and develop a Learnable Local Saliency Kernel Network (L2SKNet\footnote{https://github.com/fengyiwu98/L2SKNet}). A learnable Local Saliency Kernel Module (LLSKM) employs the concept of “Center subtracts Neighbors,” guiding the network to capture the saliency features. In addition LLSKM uses a strategic dilation and is refined into a more compact form by factorizing it into two orthogonal 1-D kernels making it lightweight. Experiments show that L2SKNet outperforms previous methods. \\
\item \textbf{Physics-Informed Guidance:} Conventional SAM-based detectors are based on external prompts and encoders pretrained with natural images. This training limits their practicality when prior knowledge is scarce and targets are extremely small \cite{DL-530}. To address this problem, Li et al. \cite{DL-204} design in 2026 a Physic-Informed SAM (PI-SAM) method which embeds infrared physical priors into the encoder. PI-SAM consists of three components. First, a Physics-aware Encoder Adapter (PEA)  estimates local luminance contrast and a SNR via center–surround mechanism  modulates multilevel features across spatial and channel dimensions. Second, a heatmap-prompt mechanism converts physical cues into non interactive prompt heatmaps and refines predictions through conditionally parameterized convolutions. Third, a Physics-Constrained Loss (PCL)  imposes both contrast- and SNR-aware regularization with sparsity and total-variation constraints. Experiments show that PI-SAM substantially improves the SAM baseline, confirming that incorporating physical principles into foundation models improves it suitability in terms of robustness and accuracy of IRSTD. \\
\item \textbf{Memory Guidance:} In 2024, Duan et al. \cite{DL-511} design a
Triple-domain Strategy (Tridos\footnote{https://github.com/UESTC-nnLab/Tridos}) with the frequency-aware memory enhancement on spatio-temporal domain. Tridos extracts and enhances frequency features by a local-global frequency-aware module with Fourier transform. The memory enhancement captures the spatial relations of infrared targets among video frames and encodes temporal dynamics motion features via differential learning and residual enhancing. A residual compensation reconcile possible cross-domain feature mismatches. Experiments on DAUB, ITSDT-15K and IRDST datasets show that Tridos outperforms ISNet \cite{DL-211}, UIU-Net \cite{DL-220}, RPCA-Net \cite{SFD-RPCAMatrix-64}, ST-Trans and SSTNet \cite{DL-38}. In 2026, Zhang et al. \cite{DL-500} design  a unified memory-enhanced spatio-temporal fusion framework DCPNet for MISTD. First, a Deformable Spatio-Temporal Module (DSTM) captures inter-frame consistency. Second, a Content-Guided Adaptive Upsampling (CGAU) module preserves weak target structures. Third, a Prototype-aware Memory Reconstruction (PMR) module improves target-background separability. Experiments demonstrate that DCPNet outperforms 23 previous SOTA algorithms methods on pixel- and target-level metrics offering a $5\%$ IoU gain in complex scenes. \\
\item \textbf{Feedback Guidance:} While multi-frame detection methods exploits temporal cues, most of the methods fails in long-range dependency modeling and robustness. To address these challenges, feedback guidance has been proposed by several researchers. In 2026, Wu et al. \cite{DL-142} design a Dynamic Feedback Iterative Network (DFINet) which guides the model training and prediction based on the History Prediction Mask (HPMK) of previous rounds. In the training phase, DFINet extract the key features of real targets by training in multiple iterations with limited data. In the prediction phase, DFINet  correct the wrong results via feedback iterative to increase the robustness. First, a Dynamic Feedback Feature Fusion Module (DFFFM) dynamically interacts HPMK with feature maps via an attention mechanism to guide feature mining and error correction. Second, a Cascaded Hybrid Pyramid Pooling Module (CHPP) allows to capture both global and local information. Third, a Dynamic Semantic Fusion Module (DSFM) employs feedback information to guide the fusion of high-level and low-level features for better feature representation in diverse scenarios.  Experiments on the NUDT-SIRST, IRSTD-1k, and SIRST datasets show that DFINet outperforms several SOTA methods. In an other work, Huang et al. \cite{DL-146} propose a sparse frames-based spatio-temporal semantic feedback network called FeedbackSTS-Det \footnote{https://github.com/IDIP-Lab/FeedbackSTS-Det}. This feedback strategy with a closed-loop semantic association mechanism comprises paired forward and backward refinement modules that work cooperatively across the encoder and decoder. Both modules incorporate an embedded Sparse Semantic Module (SSM). SSM performs structured sparse temporal modeling to capture long-range dependencies with low computational cost. The overall procedure keeps a consistent training-inference pipeline allowing reliable performance transfer and increasing robustness. Experiments show that FeedbackSTS-Det outperforms model-based methods (TopHat, NRAM \cite{SFD-RPCAMatrix-10}, PSTNN \cite{SFD-RPCATensor-5})  and data-based methods (LCNet, DNA-Net  \cite{DL-12}, RDIAN \cite{DL-33}, ISTDU-Net \cite{DL-506}, RPCANet \cite{SFD-RPCAMatrix-46}, L2SKNet \cite{DL-370}) \\
\item \textbf{Noise-Suppression Guidance:} In 2025, Xu et al. \cite{DL-530} propose a Noise-Suppression-Driven approach that adapts SAM (NSDSAM) at both internal and external levels. First, a hybrid adapter for suppressing the noise of feature maps consists of  an Multilayer Perceptron (MLP) adapter and a self-attention adapter. The self-attention adapter allows entropy-aware reconstruction of features from noisy inputs and employs a gating mechanism for soft-attention fusion mitigates SAM’s sensitivity to noise. Second, a Spatial–Frequency Hybrid Module (SFHM) jointly processes spatial and frequency domains to address the self-attention model’s bias toward low-frequency components. Experiments show that NSDSAM achieves SOTA performance. \\
\item \textbf{Geometry Guidance:} In 2025, Cai et al. \cite{DL-170}
develop a Geometry-Aware Feature Ranking Network (GeoRankNet) which consists of three novel modules. First, a Geometry-Aware Deformable Difference Convolution (GAD2C) learns adaptive sampling offsets to align convolutional kernels with irregular target contours. Second, a Dynamic Feedback-Driven Top-K Attention which is a training-aware feature ranking mechanism adjusts the pooling cardinality in real time based on IoU and loss feedback. Third, a Gated Multi-Attention Fusion (GMAF) jointly exploits channel, spatial, and global context attention and uses a lightweight gating network on globally pooled features to adaptively fuse multi-scale encoder and decoder representations. Experiments on diverse SIRST based datasets show that GeoRankNet outperforms previous methods in both detection accuracy and robustness under complex backgrounds.  In 2026,
Gao et al. \cite{DL-197} design a method named RPR-Net\footnote{https://github.com/GaoScience/point-to-mask} that bridges low-cost point supervision and mask-level detection via two components. First, a Physics-driven Adaptive Mask Generation (PAMG) module converts point annotations into compact target masks and geometric cues. Second, a lightweight Radius-aware Point Regression Network (RPR-Net) formulates IRSTD as target center localization and effective radius regression using spatiotemporal motion cues. PAMG provides pseudo masks and geometric supervision during training, while the geometric predictions of RPR-Net are fed back to PAMG for pixel-level mask recovery during inference. Experiments show that RPR-Net+PAMG outperforms ACM \cite{DL-10}, AGPCNet \cite{DL-21}, DNA-Net \cite{DL-12}, ISNet \cite{DL-211}, RDIAN \cite{DL-33}, SCTransNet \cite{DL-36}, DTUM \cite{DL-512} and SSTNet \cite{DL-38}. \\
\item \textbf{Ordinary Differential Equation Guidance:} Most of the fusion methods do nor effectively use global information to enforce the local features of the target. To address this problem, Xue et al. \cite{DL-272} design the OFSPNet\footnote{https://github.com/1yanchen3/12.git} model in order to improve detection accuracy and detail-capturing ability via diversified feature representation and adaptive fusion. First, OFSPNet uses a parallel architecture integrating CNN, transformer, and Mamba—CNN. Mamba relies on its sole selection mechanism to filter out irrelevant data and keep the important features of key visual cues. Second, an ODE-inspired adaptive feature fusion module is developed as an information bottleneck to tackle high-frequency noise, while simultaneously strengthening target features via back-propagated gradients. Third, a modular design is carried out for the transformer and Mamba structures. With optimization strategies, the inference speed has been increased by $15.28\%$ providing the dual advantages of efficient inference and high detection accuracy. In 2024,
Chen and Ye \cite{DL-290} propose a self-attention operation to interact with frequency information and design Frequency Interaction Attention (FIA). To tackle irrelevant disturbances, a family of neural Ordinary Differential Equation (ODE) modules are stacked behind FIA based on high-accuracy ODE solvers. A coefficient learning is employed to block design to lower the risk of gradient vanishing that most neural ODE networks have. A family of frequency neural ODE networks named FreqODEs is then obtained. Experiments on the NUAA-SIRST and IRSTD-1k datasets show the superiority of FreqODEs. \\
\item \textbf{Fuzzy-Logic Guidance:} In 2026, Zhang et al. \cite{DL-396} design a Fuzzy-Logic-guided Network (FLINet) which employs fuzzy sets for uncertainty-aware feature fusion. FLINet is based on a U-shaped encoder-decoder architecture and consists of two modules. First, the Global-Local Fuzzy computation module (GLF) module consists of a dual-branch fuzzy reasoning at both local-window and global-image levels to model feature uncertainty during cross-level fusion.  More specifically, the local branch increases the contrast-sensitive target cues employing an adaptive membership-based amplification. Thee global branch tackles clutter-dominant responses through global statistical normalization, jointly improving target-background separation and morphology preservation. Second, the Difference-Aware Feature fusion module DAF module increased the cross-level integration by using inter-level discrepancy maps to guide a self-attention-based fusion. The DAF module prioritizes complementary fine-grained details from shallow layers and semantic cues from deep layers to reduce small-target information loss. Experiments on the NUDT-SIRST, NUAA-SIRST, and IRSTD-1K datasets show that FLINet offers superior performance across multiple evaluation metrics.
\item \textbf{High Resolution Guidance:} In 2025, Zhang et al. \cite{DL-599-207} propose a High-resolution Guided Hierarchical Discriminative Network (H2D-Net) which consists of a High Resolution Guidance (HRG) module and a Holistic Distractor Filter (HDF). More specifically,
an extra hierarchical network with fixed scale is allows to rectify the representation bias of the U-shape network through a group of the HRG modules. Thus, it facilitate bidirectional interaction between the fine-grained spatial details and multi-scale representations. Secod, the refining HDF module is embedded into the bottleneck between the encoder and decoder for the purpose of interrupting feedforward propagation of the false feature response.
\end{enumerate}

\subsection{Robust Algorithms}
Table \ref{OverviewRLA} shows an overview of the robust algorithms.

\begin{table*}
\begin{center}
\scalebox{0.79}{
\begin{tabular}{|l|l|}
\hline
\scriptsize{Methods} &\scriptsize{Acronyms}  \\
\hline
\hline
\scriptsize{Robust Algorithms} &\scriptsize{RISTDnet (Hou et al. \cite{DL-3}), MISTNet , SANet (2026) (Zhang et al. \cite{DL-566})} \\
\hline
\scriptsize{Refining False Detections Algorithms}  &\scriptsize{RefineFD (Du et al. \cite{DL-599-45}), RefineDet (Xu et al. \cite{DL-599-47}) } \\
\hline
\scriptsize{Sparsity Algorithms} &\scriptsize{SAGCP (SAGCPDet) (Wu et al. \cite{DL-60}), ACSNet (Wu et al. \cite{DL-513})} \\
\hline
\scriptsize{Aliasing Rectification Algorithms}  &\scriptsize{FGARNet (Peng et al. \cite{DL-238}, EAA-Net (Zhang et al. \cite{DL-369})} \\
\hline
\scriptsize{Low-illumination Algorithms}  &\scriptsize{Nv-Net (Chen et al. \cite{DL-16}), CFD-Net (Zhang et al. \cite{DL-103}, FDI-Net (Chen et al. \cite{DL-103-1})} \\
\hline
\scriptsize{Lightweight Algorithms} &\scriptsize{Faster R-CNN [Detection] (Bai et al. \cite{DL-1}), TBC-Net [Single-Frame] (Zhao et al. \cite{DL-2}), TY-RIST [Detection] (Zhao et al. \cite{RT-10}), IADF-YOLO (Sun et al. \cite{RT-11})} \\ 
\scriptsize{} &\scriptsize{Lightweight FPGA-CNN (Karamov and Zhegulin \cite{RT-20})} \\
\scriptsize{} &\scriptsize{LR-Net (Yu et al. \cite{DL-40}), ILNet [Single-Frame] (Li et al. \cite{DL-70})}\\
\scriptsize{} &\scriptsize{IR-WSANet [Detection](Jin et al. \cite{DL-113})} \\ 
\scriptsize{} &\scriptsize{RT-DETR-CST [Detection](Duan et al. \cite{DL-118}), LLCEFA (Wang et al. \cite{DL-190})} \\
\scriptsize{} &\scriptsize{TFSP-IR (Zhang et al. \cite{DL-207})} \\ 
\scriptsize{} &\scriptsize{STMG-PCNet (Zhao et al. \cite{DL-295}), UAMF (Cui et al. \cite{DL-297})} \\ 
\scriptsize{} &\scriptsize{DeepPro (Li et al. \cite{DL-306}),  WSNet (Lu et al. \cite{DL-316})} \\
\scriptsize{} &\scriptsize{BiisNet (Xin et al. \cite{DL-362})} \\ 
\scriptsize{} &\scriptsize{MDPNet-SD (Chen et al. \cite{DL-371}), LMDNet (Zang et al. \cite{DL-372}), LE-IRSTD [Detection] (Jing et al. \cite{DL-550})} \\
\scriptsize{} &\scriptsize{LCICNet (Ji et al. \cite{DL-596}), NHPKD  (Cai et al. \cite{DL-599-43}), AMFUNet (Chung et al. \cite{DL-599-60})} \\
\scriptsize{}  &\scriptsize{LWMCNet (Zhang et al. \cite{DL-599-444}), HALO-Net (Wei et al. \cite{DL-599-447}) } \\
\hline
\end{tabular}}
\caption{Robust and Lightweight Algorithms: An Overview (31 papers).}
\label{OverviewRLA}
\end{center}
\end{table*}

\begin{enumerate}
\item \textbf{Robust Algorithms:} In 2021, Hou et al. \cite{DL-3} propose a Robust Infrared Small Target Detection Network (RISTDnet). First, a feature extraction framework combines handcrafted feature methods and convolutional neural networks. Second, a mapping network between feature maps and the likelihood of small targets in the image is provided, and a threshold is applied on the likelihood map to obtain the targets. Experiments show that RISTDnet detect small targets with different sizes and low SNRs in complex backgrounds with better robustness than previous algorithms. In 2026, Zhang et al. \cite{DL-566} design  a Selective Attention-based Network (SANet\footnote{https://gitcode.com/m0-61988291/SANet}) based on the U-Net framework with two modules. First, a Dual-path Semantic-aware Module (DSM) employs standard convolutions for local spatial detail preservation with pinwheel-shaped convolutions for expanded, direction-sensitive receptive fields, followed by a Convolutional Block Attention Module (CBAM) for fine-grained spatial–channel feature recalibration. Second, a Selective Attention Fusion Module (SAFM) replaces conventional static skip connections with a spatially adaptive, learnable weighting mechanism to perform context-aware, cross-scale feature fusion. Experiments the NUAA-SIRST, IRSTD-1K, and NUDT-SIRST datasets show that SANet consistently outperforms DNA-Net \cite{DL-12}, UIU-Net \cite{DL-220}, SCTransNet \cite{DL-36}, and L2SKNet \cite{DL-370}. \\

In the context of \textit{multi-frame infrared small target detection}, Gao et al. \cite{DL-563} design a robust baseline networks named (MISTNet\footnote{https://github.com/GR-ray/MIST}). based on the Information Bottleneck theory. First, a shifted neighborhood compensation block models multi-scale correspondences for implicit motion compensation to handle irregular and fast motion. Second, a progressive distillation decoder  hierarchically filter out redundancy while keeping target-relevant information. Extensive experiments on the MIST dataset show that MISTNet outperforms 31 previous methods with an $6\%$ gain in the IoU metric. \\ 
\item \textbf{Sparsity Algorithms:} Most of the pruning methods mainly focus on the optimization of classification networks, with an emphasis on semantic information. But, the spatial details are excessively pruned in the shallow layers causing a significant degradation of detection performance. Based on the sparse distribution of small targets in infrared images,  Wu et al. \cite{DL-60} design in 2025 a Sparsity-Aware Global Channel Pruning (SAGCP) method. More specifically, sparse modeling allows to code the target region for the first time, and sparse priors can be induced into the feature map for the identification of redundant channels. This pruning method exploits the inherent properties of infrared small targets to extract more robust features and obtain more compact models. Applying SAGCP to previous infrared small-target detection methods, the pruned version provides superiority in model efficiency and detection performance. When applying SAGCP to DNA-NeSAGCPt, the pruned model provides a $72.34\%$ reduction in parameters, and a $57.49\%$ gain in floating point operations with a $2.02\%$ increase in intersection over union (IoU). To address the sparse data, the standard “dense” implementation of traditional convolutions is inefficient. To address this limitation, Wu et al. \cite{DL-513}  design an Adaptive Sparse Convolutional Network (ACSNet) with differentiable binarization. First, a differentiable binarization module adaptively learns target and background region features. Second, the features are divided into significant and redundant parts. Third, sparse convolution are employed for lightweight feature extraction. \\
\item \textbf{Aliasing Rectification Algorithms:} Most of the deep neural networks based approaches focus on extracting spatial patterns of small targets. Nevertheless, these approaches often suffer from computational redundancy and overfitting. In addition, aliasing occurs when the sampling frequency doesn’t meet the Nyquist criterion and significantly reduces the accuracy of detection. To address these challenges,  Zhang et al. \cite{DL-369} design in 2025 an anti-aliasing network (EAA-Net) which consists of an Anti-Aliasing Contextual Aggregation (AACA) module and a Partial Differential Channel Pruning (PDCP) module. The AACA module allows to tackle aliasing effects by filtering out channels severely affected by aliasing. The PDCP module allows tackle redundant parameters and low model efficiency. PDCP represents the weight matrix with a Partial Differential Equation (PDE). These modules improve the capacity of anti-aliasing and reduce its size. Experiments on the  IRSTD-1k and SIRST datasets show that EAA-Net outperforms SOTA methods. In 2026, Peng et al. \cite{DL-238} analyze the frequency-domain characteristics of small targets to improve the model’s ability to identify their critical features. As the downsampling process may introduce aliasing that distorts the target signals and results in inaccurate segmentation masks, Peng et al. \cite{DL-238} develop a Frequency Guidance and Aliasing Rectification Network (FGARNet). It employs a Bidirectional Frequency Filter (BFF) to isolate specific frequency bands to improve the saliency of small targets in the frequency domain. A Wavelet Rectifying Downsampling (WRD) module allows to correct aliasing using wavelet-derived high-frequency features. Experiments on the IRSTD-1k dataset show that FGARNet outperforms the second-best by $2.74\%$. \\
\item \textbf{Low Illumination Algorithms:} IRSD suffers from the weak illumination and low-resolution problems. In 2020, Chen et al. \cite{DL-16} design a CNN model to tackle weak illumination in natural environments. First, the Nv-Net network allows to segment the orignal images with an enforcement layer in the front end of the framework. Second, a weighted-sigmoid-cross-entropy loss function computes the error between the prediction of the network and the ground-truth. Third, a mean-variance normalization preprocessing is employed to reduce computation time. Experiments on the LII dataset \cite{DL-16} show that Nv-Net method provides the best detection performance in the low illumination environment. In 2023, Chen et al. \cite{DL-103-1} provide a Fluid Dynamics-Inspired Network (FDI-Net). Based on the Taylor Central Difference (TCD) method, a TCD feature extraction block is employed in which convolution and Transformer structures are fused for local and global information. The pixel motion equation is derived from the Navier–Stokes (N-S) equation obtaining a N-S Refinement Module which refines extracted features with edge details. Thus, the TCD feature extraction block allows to determine the primary movement direction of pixels during detection. The NS Refinement Module corrects some skewed directions of the pixel stream to supplement the edge details. Experiments on IRSTD-1k and SIRST datasets show that FDI-Net outperform previous SOTA methods in the presence of low illumination conditions.In 2025, Zhang et al. \cite{DL-103} develop a Computational Fluid Dynamic Network (CFD-Net) derived from computational fluid dynamics. Based on the unilateral difference equation with third-order,  a unilateral differential residual  structure is employed as the backbone. Thus, the pixel stream only flows in the forward direction. Then, a Switch-Controlled Multidirectional Treatment Tank (SMTT) allows to dynamically guide the pixel stream to the appropriate path for different targets with varying shapes and orientation. Experiments on the IRSTD-1k and SIRST datasets show that  CFD-Net  outperform previous SOTA methods in the low illumination environments.
\end{enumerate}

\subsection{Lightweight Algorithms}
Different strategies have been developed to obtain lightweight networks. In 2019, Zhao et al. \cite{DL-2} propose a lightweight CNN named TBC-Net.  TBCNet comprises a Target Extraction Module (TEM) and a Semantic Constraint Module (SCM). A joint loss function and a training method are also developed. The SCM module provides a semantic constraint on TEM by combining the high-level classification task and addresses the difficulty to learn features due to the class imbalance problem. During the training phase, the targets are extracted and are further classified by the SCM module. During the inference phase, the TEM module is solely employed to detect the small targets. Experiments show that TBC-Net reduces the false alarm due to complex backgrounds. TBC-Net can reach real-time detection on the NVIDIA Jetson AGX  development board making it suitable for deployment in real conditions. In 2024, Yu et al. \cite{DL-40} design a Lightweight and Robust network (LR-Net) without complex structure. LR-Net offers an effective balance between detection accuracy and resource requirements. More specifically, a Lightweight Feature Extraction Attention (LFEA) module extracts target features and strengthens information interaction across channels. A Refined Feature Transfer (RFT) module improves the network's feature refinement extraction ability with little resource requirements. A Low-level Feature Distribution (LFD) strategy allows to  address the problem of small target loss in high-level feature maps.
An Simplified Bilinear interpolation Attention Module (SBAM) promotes the guidance constraints of low-level features on high-level features and the fusion of the two. The traditional resizing method is replaced by a  training and inference cropping strategy. Experimental results show that LR-Net achieves SOTA performance and is ranked at the 3rd place in the "ICPR 2024 Resource-Limited Infrared Small Target Detection Challenge Track 2: Lightweight Infrared Small Target Detection". \\

In 2023, Chen et al. \cite{DL-371} develop a lightweight network with high-speed inference and a knowledge distillation method for learning higher-order semantic information from the transformer network. First, Chen et al. \cite{DL-371} design each stage in the detection network, named Multi-scale Dilated Pyramid Network (MDPNet). This design enlarges the receptive field and improves the ability to extract contextual features. Second, a knowledge distillation allows to bridge the detection performance gap with the transformer. Based on the self-attention mechanism, a semantic distillation sub-network is constructed between the teacher transformer and student convolution network. Experiments on the  SIRST and NUDT-SIRST datasets show that MDPNet outperforms previous networks requiring a number of parameters of only 0.23 M. MDPNet achieves an inference speed of 137 FPS for $320 \times 320$ infrared images. \\

In 2025, Li et al. \cite{DL-70} develop an Infrared Low-level Network (ILNet\footnote{https://github.com/Li-Haoqing/ILNet}) that considers infrared small targets as salient regions characterized by semantic information. More specifically, ILNet employs low-level information instead of treating it uniformly with high-level information. A lightweight feature fusion module, named the Interactive Polarized Orthogonal Fusion module (IPOF), allows to integrate important low-level features from the shallow layers into the deep layers. A dynamic 1-D aggregation layers are inserted into the IPOF allowing to dynamically adjust the aggregation of low-dimensional information according to the number of input channels. In an other work, Jin et al. \cite{DL-113} design a lightweight network based on an improved YOLOv10 (IR-WSANet)  increasing the detection performance via a frequency-spatial joint optimization strategy. First, a discrete wavelet transform convolution (DWaveletConv) is employed into the backbone network. The fusion of high-frequency details and low-frequency semantics is improved by multi-band feature decomposition to suppress noise interference. Second, a cooperative module (POS-SHSA) improves the positioning accuracy of small targets. Experiments on the SIDD and HIT-UAV datasets show that IR-WSANet improves the detection performance in complex scenes through the combination of frequency domain filtering enhancement and space-channel dual attention mechanism. Duan et al. \cite{DL-118} design a lightweight architecture based on RT-DETR named RT-DETR-CST. First, a Cross-Channel Feature Attention Network (CFAN) allows to achieve channel-weighted feature fusion via residual connections to suppress invalid background channels. Second, a Scale-Wise Feature Network (SWN) employs  depthwise separable convolutions and stochastic depth for multi-scale feature extraction. Third, a Texture/Detail Capture Network (TCN) achieves edge/detail capture via linear decomposition. Experiments on the ISDD datasets show that RT-DETR-CST offers real-time performance in infrared ship remote sensing small-target detection.  In an other work, Li et al. \cite{DL-306} investigate the most essential information from a more crucial domain. A theoretical analysis demonstrate that the global temporal saliency and correlation information in the temporal profile offer significant superiority in separating target signals from other signals. First,  Li et al. \cite{DL-306} construct the first prediction attribution tool  and verify the importance of the temporal profile information. S, the IRST detection task is formulated as a one-dimensional signal anomaly detection task. Thus, Li et al. \cite{DL-306} propose an efficient Deep temporal Probe network (DeepPro\footnote{https://github.com/TinaLRJ/DeepPro}) that executes only calculations in the time dimension. Experiments show that DeepPro outperforms existing state-of-the-art IRST detection methods with reasonable computation time. Zang et al. \cite{DL-372} propose a  lightweight model—one named  LMDNet with few parameters.  LMDNet reduces computational load—while surpassing the average detection accuracy and IoU values of previous SOTA models. First, a spatial-channel difference convolution module allows to extract spatial and channel features while maintaining low computational requirement. Second, a global feature enhancement Mamba layer accelerates target localization and offers precise detection in complex backgrounds. Third, a self-learning joint loss function dynamically adjusts weights during training to optimize the contributions of different loss components. Practically, the FPS is several times higher than that of larger model-based methods, with Params and FLOPs of only 0.024M and 0.168G, respectively. Its overall performance outperforms previous lightweight methods. \\

In 2026, Wang et al. \cite{DL-190} provide a Lightweight Local Contrast and Edge Fusion Approach (LLCEFA). LLCEFA consists of a dual-branch architecture with a main branch and an edge branch designed to take into account the synergistic advantages of prior knowledge guidance and data-driven learning. About the main branch, a Local Contrast Attention Fusion (LCAF) module employs the local contrast prior to obtain the input image into more discriminative feature representations. An edge branch employs an Adaptive Edge Weight Extraction (AEWE) module to independently derive edge weights for enhancing intermediate-layer features. This design decouples prior knowledge processing from the backbone's architecture making LLCEFA suitable to be a plug-and-play enhancer compatible with various backbone methods, achieving a performance boost with generating  a minimal additional burden. In an other work, Zhang et al. \cite{DL-207} propose a Tropical-Fish-Swarm Pruning for IR (TFSP-IR). It is a compact  architecture that employs feature restoration, adaptive filtering, and bioinspired pruning. First, a Fine-detail-guided Multilevel Feature Compensation (F-MFC) mechanism aggregates information from adjacent layers. Thus, small-target details persist into the network’s deeper stages. Second, a Cross-level Feature Correlation (CFC) module employs an energy-based filter from high-level, noise-reduced representations. It allows to suppress clutter and enhancing target signal. Third, an Efficient Fast Channel Pruning (EFCP) strategy adjusts pruning depth and breadth according to the evolving spatial relationships within feature maps. Experiments on three datasets show that TFSP-IR offers state-of-the-art performance in both accuracy and efficiency. Zhao et al. \cite{DL-295} propose a lightweight Shape-Texture Mutual-Guided Predictor-Corrector Network (STMG-PCNet). STMG-PCNet offers a balance between accuracy and efficiency by providing a collaborative design of a detection branch and a shape-aware branch. First, a progressive lightweight dilated dense module in the detection branch employs dense connections and cascaded multiscale dilated convolutions to improve local texture retention and precise localization. Second, a shape-aware vision Transformer module in the shape-aware branch allows to model target edges and shape features via global context interaction. Third, a numerical predictor-corrector fusion module achieves lightweight multistage feature fusion. Experiments show that STMG-PCNet outperforms previous algorithms in detection accuracy and, computational efficiency. In an other work, Cui et al. \cite{DL-297} build upon the traditional U-Net architecture a lightweight Adaptive Memory Fusion (AMF) module at the deepest layer of the network. This module allows to take into account the strength of attention mechanisms in capturing long-range contextual information and adaptively fuses multi-frame inputs to improve feature representations. Experiments show that UAMF\footnote{https://github.com/keacifer/UAMF} allows to reduce computational complexity  by two orders of magnitude compared with previous SOTA models. In 2026, Lu et al. \cite{DL-316} develop an architecture named WSNet which contains only 0.054 M parameters and requires 1.050 G FLOPs. WSNet was the most lightweight model in the field achieving the fastest inference speed to date. \\

Binarized Neural Networks (BNNs) offers efficiency in model compression.  Thus, Xin et al. \cite{DL-362} design a Binarized Infrared Small-Target Detection Network (BiisNet\footnote{https://github.com/joeyxin-del/BiisNet}). It keeps the core operations of binarized convolutions while integrating full-precision features into the network's information flow. More specifically, a Dot Binary Convolution retains fine-grained semantic information in feature maps while still exploiting the binarized convolution operations. A smooth and adaptive Dynamic Softsign function offers a more comprehensive and progressively finer gradient during backpropagation improving model stability and promoting an optimal weight distribution. Experiments show that BiisNet outperforms other binary architectures and has strong competitiveness among state-of-the-art full-precision models.

In summary, model compression is the main way for obtaining lightweight networks and structured pruning is a representative lightweight method \cite{MOD-Theory-300}. Due to the extreme characteristics of infrared small targets (e.g., weak texture, shape, and color distribution), directly applying existing pruning methods to infrared small target detection appears challenging because they have been designed for other tasks \cite{MOD-Theory-300} reaching to the following weaknesses as developed in Li et al. \cite{DL-599-118}:
\begin{itemize}
\item \textbf{Error Pruning of Essentialrr Modules:} Existing pruning methods  consider the specially designed modules (e.g., CBAM in DNA-Net \cite{DL-12} , BLAM in ALCNet \cite{DL-11}) and conventional components as equally important. Several essential modules are thus removed generating decrease in performance .
\item \textbf{Structural Coupled Dependencies:} Advanced infrared small target detection networks use multi-scale feature fusion or residual cross-layer connections to obtain strong inter-channel dependencies. But, existing pruning methods are insensitive to these coupled relationships and rely on expert-driven grouping strategies, which struggle to adapt to parameter-sharing mechanisms in cross-layer connections and generate performance degradation. 
\item  \textbf{Weight Correlation Constraint:} Existing pruning methods focus on the identification redundant structures while ignoring the substructures correlation. This fact generates feature distribution perturbation and weaken the recovery and generalization ability
\end{itemize}
To address these three challenges, Li et al. \cite{DL-599-118} design a Structure Grouping collaborative Decorrelated L2-norm Pruning (SGDP\footnote{https://github.com/lyh-nudt/SGDP}) which is a task-aware structured pruning method specifically designed to infrared small target detection. SGDP method is not intended to reduce Params and FLOPs, but to achieve a balance between model compression and small target detection performance. First, a Shield Layer (SL) strategy  protects important layers from pruning. Second, a Hierarchical Coupling Structure Grouping (HCSG) strategy exploits hierarchical coupling relationships and input–output dependencies to partition non-shielded structures into grouped substructures. HCSG avoids the disruption of critical information flow paths of small targets and ensures networks regularity. Third, a Decorrelated L2-norm Regularization (DL2R) strategy  enforces substructure regularization and transform correlation matrices into non-diagonally dominant forms. The redundant weights are suppressed and these features are redistributed to the retained filters. Thus, the layer-wise optimization with regularized weight-bias updating improves pruned model performance.

\section{Wavelets based Methods}
\label{Wavelets}
Table \ref{OverviewWM} shows an overview of the wavelets based methods.

\begin{table}
\scalebox{0.79}{
\begin{tabular}{|l|l|}
\hline
\scriptsize{Methods} &\scriptsize{Acronyms} 
\\
\hline
\hline
\scriptsize{Wavelets} &\scriptsize{HaarConv (YOLOv5-s) [Detection] (Song et al. \cite{DL-5}), HaarTransNet (Fan et al. \cite{DL-205})} \\
\scriptsize{} &\scriptsize{IRPruneDet (Zhang et al. \cite{DL-35})} \\
\scriptsize{} &\scriptsize{WMRNet (Zhang et al. \cite{DL-120}), FSCNet (Li et al. \cite{DL-99}), WCDMF-Net (Li et al. \cite{DL-201}) } \\
\scriptsize{} &\scriptsize{DFAwareNet (Cheng et al. \cite{DL-135}), SDANet (UNet) (Zhang et al. \cite{DL-198})} \\
\scriptsize{} &\scriptsize{WaveTD (Ma et al. \cite{DL-215}), WTAPNet (He et al. \cite{DL-216}), SWAN (Jing et al. \cite{DL-217})} \\
\scriptsize{} &\scriptsize{WRSSNet (Jiang et al. \cite{DL-218}), Ivan-ISTD (Li et al. \cite{DL-246}), IRPruneDeXt (Zhang et al. \cite{DL-247}) } \\
\scriptsize{} &\scriptsize{ARFC-WAHNet (Cui et al. \cite{DL-311}), PMNet  (Lian et al. \cite{DL-343}), IRKAN (Ding et al. \cite{DL-345})} \\
\scriptsize{} &\scriptsize{WT-HMFF (Li et al. \cite{DL-537}), FM-Net (Liu et al. \cite{DL-579}), DWTFreqNet (Li et al. \cite{DL-599-2})} \\
\scriptsize{} &\scriptsize{IR-WASNet (Jin and Chen \cite{DL-599-72}), DSGWT-IRFR (Wu et al. \cite{DL-599-334})} \\
\scriptsize{} &\scriptsize{WTPNet [MIRST] (Gao et al. \cite{DL-599-335}), WAMENet (Xiao et al. \cite{DL-599-344})} \\
\scriptsize{} &\scriptsize{CWLE-Net (Deng et al. \cite{DL-599-378})} \\
\hline
\end{tabular}}
\caption{Wavelets Methods: An Overview (24 papers).}
\label{OverviewWM}
\end{table}

In 2021, Song et al. \cite{DL-5} propose a semi-supervised network based on Haar Wavelet (HaarConv) using YOLOv5-s. The HaarConv module based on the high-frequency features is obtained by Haar wavelet decomposition. A HaarUp-HaarDown module is obtained by using Haar forward and inverse transform,replacing the up-sampling layer of the feature pyramid network and the down-sampling layer of backbone network to preserve smaller target features. A pseudo-label-based module conducts semi-supervised learning and reduces the labeling cost. Experiments show that HaarConv  improve the detection accuracy of dim and small infrared ship targets by more than $2\%$, and the number of parameters increases by only about 0.02M. \\

In 2024, He et al. \cite{DL-216} design a Wavelet Transform-based Augmented Perception Network (WTAPNet\footnote{ https://github.com/MinjieWan/WTAPNet}). First, an Enhancement and Enlarge (EE) module enhances the perceptual capability  by magnifying image resolution and augmenting target features at the same time. Second, a Discrete Wavelet Transform-based Downsampling (DWTD) module and an inverse Wavelet Transform-based Fusion (IWTF) module collaboratively work to extract and combine multiscale features. Third, a bottom-up path fusion strategy highlight and preserve small-target features. Experiments on the NUDT-SIRST and the SIRST datasets show that WTAPNet obtains similar performance than previous SOTA algorithms. Zhang et al. \cite{DL-35} propose a Wavelet Structure-Regularized Soft Channel Pruning Network (IRPruneDet\footnote{https://github.com/hd0013/IRPruneDet}). IRPruneDet represents the weight matrix in the wavelet domain and formulate a wavelet channel pruning strategy. A wavelet regularization induces structural sparsity without incurring extra memory usage. A soft channel reconstruction method keeps key target information against premature pruning. Extensive experiments demonstrate that IRPruneDet provides a $64.13\%$ reduction in parameters and a $51.19\%$ decrease in FLOPS with an improvement of IoU from $73.31\%$ to $75.12\%$ and nIoU from $70.92\%$ to $74.30\%$.\\

In 2025, Zhang et al. \cite{DL-120} propose a Wavelet Mamba with Reversible Structure Network (WMRNet) which comprises a Discrete Wavelet Mamba (DW-Mamba) module and a Third-order Difference Equation guided Reversible (TDE-Rev) structure. First, DW-Mamba uses the Discrete Wavelet Transform to decompose images into multiple subbands, integrating this information into the state equations of a state space model. DW- Mamba minimizes frequency interference keeping a global perspective avoiding background aliasing. Second, TDE-Rev suppresses edge aliasing effects by refining the target edges and promotes feature interactions via a reversible structure. Experiments on the IRSTD-1k and SIRST datasets show  that  WMRNet outperforms previous SOTA algorithms. Li et al. \cite{DL-99} propose a Feature Synthesis with wavelet Coefficients Networks (FSCNet) addressing the challenge of information loss prevalent deep learning models during the down-sampling process. First, FSCNet employs a wavelet coefficient-based feature extraction mechanism to separate small targets from the background. Based on a Feature Attention Gate Module (FAGM), the network enhances high-frequency details. Second, a Group Spatial-Channel Transformer Block (G-SCTB) allows the interaction and enhancement of multi-scale features. Experiments on the NUAA-SIRST, NUDT-SIRST and IRSTD-1K datasets show that FSCNet provides competitive results compared to ACM \cite{DL-10}, ALCNet \cite{DL-11}, ISNet \cite{DL-211}, ISTDU-Net, RDIAN \cite{DL-33}, DNA-Net \cite{DL-12}, UIU-Net \cite{DL-220} and SCTransNet \cite{DL-36}. Ma et al. \cite{DL-215} propose a Wavelet Steered network (WaveTD\footnote{https://github.com/Fortuneteller6/WaveTD}) by combining Discrete Wavelet Transform (DWT) into CNN. First, Haar DWT and inverse DWT (IDWT) are embedded for lossless downsampling and upsampling for more effective feature extraction within the CNN. Second, a hybrid attention mechanism named Wavelet Steered Transformer (WST) enhances the DWT features both spatially and across channels. WST consists of two key components: 1) a Channel-wise Transformer which adapts a channel-wise Transformer to improve semantic information and suppress background clutter and noise, and 2) a Dilated-Gate Convolutional Module which enhances spatial location accuracy. It uses a combination of different kernel sizes and dilation rates to improve spatial accuracy. Experiments show that WaveTD outperforms
MSHNet \cite{DL-41}, WTAPNet \cite{DL-216}, ABCNet, AGPCNet \cite{DL-21}, DNA-Net \cite{DL-12}, LIG \cite{SFD-LCM-43-2}, PSTNN \cite{SFD-RPCATensor-5}, TLLCM and TopHat.\\

In 2025, Jing et al. \cite{DL-217} propose a Synergistic Wavelet-Attention Network (SWAN) to perceive targets from both spatial and frequency domains. First, SWAN exploits a Haar Wavelet Convolution (HWConv) for a deep, cross-domain fusion of the frequency energy and spatial details of small target. Second, a Shifted Spatial Attention (SSA) mechanism models long-range spatial dependencies with linear computational complexity for contextual awareness. Third, a Residual Dual-Channel Attention (RDCA) module adaptively calibrates channel-wise feature responses to avoird background interference while amplifying target-pertinent signals. Experiments show that SWAN outperforms ACM \cite{DL-10}, ALCNet \cite{DL-11}, DNA-Net \cite{DL-12}, RDIAN \cite{DL-33}, UIU-Net \cite{DL-220}, SCTransNet \cite{DL-36} and DATransNet. Jiang et al. \cite{DL-218} propose a Wavelet-Recalibrated Semi-Supervised Network (WRSSNet) that employs synthetic data augmentation, feature reconstruction, and semi-supervised learning to exploit the potential of unlabeled infrared images under limited supervision. First, a dataset containing 843 visible-light small target images employs an improved CycleGAN model to convert them into high-quality pseudo-infrared images. Second, a lightweight Wavelet-Enhanced Channel Recalibration and Fusion (WECRF) module employs wavelet decomposition with both channel and spatial attention mechanisms. WECRF allows adaptive reweighting and  fusion of multi-scale features, highlighting high-frequency details and weak target responses. Experiments on the NUAA-SIRST and IRSTD-1K datasets show that WRSSNet provides better performance than DNA-Net \cite{DL-12} and UIU-Net \cite{DL-220}. Li et al. \cite{DL-246} develop a doubly wavelet-guided Invariance learning framework (Ivan-ISTD\footnote{https://github.com/nanjin1/Ivan-ISTD}). First, training samples aligned with the target domain are generated using Wavelet-guided Cross-domain Synthesis. This wavelet-guided alignment module separates the target background via multi-frequency wavelet filtering. Second, a Real-domain Noise Invariance Learning (RNIL) module extracts real noise characteristics from the target domain to obtain a dynamic noise library. Thus, RNIL learns noise invariance via self-supervised loss. Experiments show that Ivan-ISTD outperforms most of the time ACM-Net, ALCNet \cite{DL-11},DNA-Net \cite{DL-12}, RDIAN \cite{DL-33}, ISTDU-Net,UIU-Net \cite{DL-220}, HCF-Net and SCTransNet  \cite{DL-36} in presence of diverse noise distributions. \\

In 2025, Cui et al. \cite{DL-311} design an Adaptive Receptive Field Convolution and Wavelet-Attentive Hierarchical Network (ARFC-WAHNet\footnote{https://github.com/Leaf2001/ARFC-WAHNet}). It comprises a Multi-Receptive Field Feature Interaction Convolution (MRFFIConv) module to adaptively extract discriminative features by integrating multiple convolutional branches with a gated unit. Second, a Wavelet Frequency Enhancement Downsampling (WFED) module exploits Haar wavelet transform and frequency-domain reconstruction to improve target features and suppress background noise. Third, a High-Low Feature Fusion (HLFF) module integrates low-level details with high-level semantics. Finally, a Global Median Enhancement Attention (GMEA) enhances feature diversity and expressiveness via global attention. Experiments on the SIRST, NUDT-SIRST, and IRSTD-1k datasts show  that ARFC-WAHNet outperforms most of the time
model-driven models (IPI, PSTNN) and data-driven models (ACM \cite{DL-10}, ALCNet \cite{DL-11}, DNA-Net \cite{DL-12}, RDIAN \cite{DL-33}, UIU-Net \cite{DL-220}, AGPCNet \cite{DL-21}, MSHNet \cite{DL-41}, MDIGCNet \cite{DL-599-93}). \\

In 2025, Zhang et al. \cite{DL-247} propose a wavelet structure-regularized multidimensional musical scale Soft Channel Pruning (SCP) method named IRPruneDeXt\footnote{https://github.com/hd0013/IRPruneDet}. It represents the weight matrix in the wavelet domain and formulates a Wavelet Channel Pruning (WCP) strategy. A wavelet regularization induces structural sparsity without incurring extra memory usage. A Multidimensional Musical scale Soft Channel Reconstruction (MMSCR) module adapts the strategy across temporal and spatial dimensions to keep key target information and prevent premature pruning. MMSCR balances pruning and reconstruction via a musical scale feedback effect. Extensive experiments demonstrate that IRPruneDeXt outperforms previous SOTA methods with a $65.68\%$ reduction in parameters,  a $51.77\%$ decrease in floating-point operations (FLOPs), an augmentation of over union (IoU) from $73.31\%$ to $76.17\%$ and of normalized IoU (nIoU) from $70.92\%$ to $75.08\%$. Li et al. \cite{DL-537} develop a Wavelet Transform Convolution and Hierarchical Multi-Scale Feature Fusion Network (WT-HMFF) which consists of a Wavelet Transform Convolution (WTConv) module and a Hierarchical Multi-Scale Feature Fusion (HMFF) module. First, WTConv expands the receptive field through wavelet convolution to capture global contextual information while keeping target shape characteristics. Second HMFF allows the fusion of shallow and deep features  keeping the high resolution of deep feature maps and avoiding the disappearance of small target features. Experiments on the SIRST and IRSTD-1k datasets demonstrate that WT-HMFF provides most of the time better results than MDvsFA \cite{DL-15}, ACM \cite{DL-10}, DNA-Net \cite{DL-12}, Pconv \cite{DL-104}, ALCNet \cite{DL-11}, IRPruneDet \cite{DL-35}, UIU-Net \cite{DL-220}, and SCTransNet \cite{DL-36}. \\

In 2026, Fan et al. \cite{DL-205} propose a HaarTransNet\footnote{https://github.com/Big-Orange-Tiger/HaarTransNet} method which exploits frequency-domain decoupling and target–background relative saliency modeling. A Haar Decoupled Downsampling Block (HDDB) allows to decouple input features along the channel dimension via the Haar Wavelet Transform (WT). HDDB  avoids feature confusion and alleviates small-target attenuation during downsampling. A Haar Channel Cross Transformer (HCCT) provides a channel cross-attention mechanism to model cross-level global correlations. The low-frequency background cue guides attention to dynamically model the relative salience of targets in complex backgrounds. Experiments on the NUDT-SIRST and IRSTD-1k datasets show that HaarTransNet provides robustness in challenging scenes. \\

In 2026, Li et al. \cite{DL-201} design a Wavelet-based Cross-Domain Multistage Feature Fusion Network (WCDMF-Net) that synergizes frequency-domain analysis with spatial-contextual learning. First, a parallel spatial–frequency dual-stream architecture employs Haar Wavelet Transform (HWT) and inverse Discrete Wavelet Transform (iDWT) for multiscale frequency feature extraction and reconstruction. Second, an adaptive spectral–spatial fusion block  enhances features within the spatial and spectral domains and fuses these refined representations to exploit the complementary strengths of both domains. Third, a Multiscale Feature Aggregation Module (MFAM) aggregates multiscale output feature maps in the decoding phase. A multilevel cross-entropy loss function allows to achieve target localization and mask prediction. Experiments on the NUDT-SIRST, NUAA-SIRST, and IRSTD-1k datasets show that WCDMF-Net provides similar performances than SOTA methods. Cheng et al. \cite{DL-135} propose a state space model with Dynamic Frequency cue Aware Network (DFAwareNet\footnote{https://github.com/Fortuneteller6/DFAwareNet}) which exploits spatial and wavelet-domain representations. A Wavelet Guided Module (WGM) refines wavelet components to improve target-related features while suppressing background interference. A Vision State Space Model (VSSM) captures global context with linear complexity. An Hybrid Repair Gate Network (HRGN) preserves weak or subtle targets via a multi-dilated aggregation. A  Mixed Feature Fusion Module (MFFM) enhances target localization through spatial–channel integration. Experiments on four benchmark datasets demonstrate that DFAwareNet achieves competitive or superior detection performance compared to TopHat, model-driven approaches (LIG \cite{SFD-LCM-43-2}, PSTNN \cite{SFD-RPCATensor-5}, TLLCM \cite{SFD-LCM-19}, HBMLCM) and  data-driven approaches (WTAPNet \cite{DL-216}, MSHNet \cite{DL-41}, MTU-Net \cite{Applications-S-41}, AGPCNet \cite{DL-21}, DNA-Net \cite{DL-12}). Zhang et al. \cite{DL-198} develop a Stride-free Dual-domain Adaptive Network (SDANet) using a U-shaped architecture. First, a High-frequency Modulated Downsampling (HMD) performs stride-free downsampling by fusing the low-frequency band and the modulated high-frequency subbands generated from the discrete wavelet transform. HMD provides richer target details during downsampling while suppressing background clutter interference. Second, a Stride-free Wavelet Enhancement Module (SWEM) employs a Multi-scale Wavelet Convolution (MWC) to extract spatial and overlooked frequency-domain representations. In the decoder, an Attention-Routed Scale Adaptive (ARSA) module dynamically combies local and non-local features via input-dependent attention routing to enhance scale robustness. Experiments show that SDANet achieves superior detection accuracy and robustness compared with previous SOTA methods. \\

In 2026, Lian et al. \cite{DL-343} design a Parametric Manifold Network (PMNet\footnote{https://github.com/Lianruoqi/PMNet}) which comprises  frequency-domain analysis with manifold learning. First, PMNet separates high-frequency features from the spatial domain to encode high-frequency target details across channels; accordingly. PMNet preserves the geometric structure of the target while suppressing noise. Second, a low-frequency feature block refines structural features improving discrimination between targets and backgrounds. Extensive experiments show the PMNet outperforms previous SOTA methods in both accuracy and robustness demonstrating the effectiveness of integrating wavelet decomposition with manifold learning. Ding et al. \cite{DL-345} develop a Kolmogorov-Arnold Network (IRKAN) with fixed, node-based activations with learnable functions on the network edges. IRKAN learns optimal, data-driven transformations for complex infrared signals. Wavelet basis functions are used to parameterize these activations. It introduces a  signal-processing prior into the network exploiting the time-frequency localization properties of wavelets to function as a learnable matched filter. Experiments on the WideIRSTD-Full dataset show that IRKAN achieves a mean normalized Intersection over Union of $68.89\%$ and a low False Alarm Rate of $8.33 \times 10^{6}$, with computational efficiency requiring 3.96M parameters.  Zero-shot evaluations on the unseen NUDT-SIRSTD and IRSTD-1K  datasets demonstrate that IRKAN offers cross-resolution generalization capabilities.

\section{Mamba-based State Space Models}
\label{Mamba} 
Mamba is a State Space Model (SSM)-based neural network architecture designed for linear-time sequence modeling.  However, the direct application of Mamba alone is not effective, as it cannot fully exploit local features \cite{DL-94}. Thus, hybrid approaches have been developed. Table \ref{OverviewArchitectureM} shows an overview of the existing methods based on a Mamba architecture.

\begin{table*}
\begin{center}
\scalebox{0.79}{
\begin{tabular}{|l|l|}
\hline
\scriptsize{Methods} &\scriptsize{Acronyms}  \\
\hline
\hline
\hline
\scriptsize{Mamba State Space Model (SSM)-based Neural Networks}  &\scriptsize{
HMCNet (Li et al. \cite{DL-75}), LVMG (Lu and Zhang \cite{DL-94}), IRMamba (Zhang et al. \cite{DL-95}) } \\
\scriptsize{}  &\scriptsize{EAMNet (Jiang et al. \cite{DL-112}), Mamba-IRSTD (Zhang et al. \cite{DL-116}), MRNet (Ye et al. \cite{DL-131})} \\
\scriptsize{}  &\scriptsize{CPMamba (Wu et al. \cite{DL-212}), BDSFusion (Li et al. \cite{DL-250}), MiM-ISTD [Single-Frame] (Chen et al. \cite{DL-260})} \\
\scriptsize{}  &\scriptsize{STLMamba [MIRST] (Qiu et al. \cite{DL-261}), TAPM-Net [MIRST] (Xie et al. \cite{DL-262}), MOU-Mamba (Liu et al. \cite{DL-263})} \\
\scriptsize{}  &\scriptsize{SMILE (Li et al. \cite{DL-264}), MRL-Mamba (Li et al. \cite{DL-265}), SBMambaNet (Yu et al. \cite{DL-266}) } \\
\scriptsize{}  &\scriptsize{SAMamba (Xu et al. \cite{DL-268}), FANM [Dense] (Xu et al. \cite{DL-269})} \\
\scriptsize{}  &\scriptsize{Spectrum-Assisted Mamba (SAMNet) (Li and Wang \cite{DL-273}), SAMamba (Sui et al. \cite{DL-320}), TAMamba  (Tang et al. \cite{DL-321})} \\
\scriptsize{}  &\scriptsize{SCOFM-net (Qiu et al. \cite{DL-344}), MGMamba-YOLOv8 [Detection] (Xiao et al. \cite{DL-515})} \\
\scriptsize{}  &\scriptsize{HMLVNet [MIRST] (Liu et al. \cite{DL-275}), Mamba-ISTD (Ran et al. \cite{DL-565}) } \\
\scriptsize{}  &\scriptsize{LCMamNet (Fan et al. \cite{DL-599-302}), MDONet (Xiao and Hu \cite{DL-599-243})} \\
\scriptsize{}  &\scriptsize{MambaLite-DNANet (Yin et al. \cite{DL-599-391}), FD-MCNet (Zhang et al. \cite{DL-599-396}) } \\
\scriptsize{}  &\scriptsize{NFM-MambaNet (Guo et al. \cite{DL-599-411}) } \\
\hline
\scriptsize{Transformers based Neural Networks} &\scriptsize{IRSTFormer [Single-Frame] (Chen et al. \cite{DL-18}), IR-TransDet (Lin et al. \cite{DL-19}),  (Liu et al. \cite{DL-20})} \\
\scriptsize{} &\scriptsize{SCTransNet [Single-Frame] (Yuan et al. \cite{DL-36}), LMAFormer [Multi-Frame](Huang et al. \cite{DL-46}), DATransNet (Hu et al. \cite{DL-69})} \\
\scriptsize{} &\scriptsize{STC (Zhu et al. \cite{DL-71}), STPSA-Net (Liu et al. \cite{DL-74}), HSHNet (Lin et al. \cite{DL-193}), HSTNet  (Li et al. \cite{DL-101})} \\
\scriptsize{} &\scriptsize{CRPE-Net (Jiang et al. \cite{DL-143}), STUTD (Luo et al. \cite{DL-62}),  CourtNet (Peng et al. \cite{DL-277}) } \\
\scriptsize{} &\scriptsize{IrisNet (Qian et al. \cite{DL-133}), LCR-RTDETR (Chen et al. \cite{DL-136}), EUSS-DETR [Detection] (Cui et al. \cite{DL-192})} \\
\scriptsize{} &\scriptsize{GMTNet (Feng et al. \cite{DL-237}), BSTNA-Net [Swin Transformer) (Li et al. \cite{DL-239}), DTFE-Net  (Hu et al. \cite{DL-244})} \\
\scriptsize{} &\scriptsize{IST-TransNet (Li et al. \cite{DL-276}), FATCNet (Yang et al. \cite{DL-278}), STASPPNet [Swin Transformer] (Wu et al. \cite{DL-279})} \\
\scriptsize{} &\scriptsize{MTU-Net [Single-Frame] (Wu et al. \cite{DL-340}), KECS-Net (Li et al. \cite{DL-350}), RKFormer (Zhang et al. \cite{DL-392}), LVNet (Shen et al. \cite{DL-507}) } \\
\scriptsize{} &\scriptsize{IRTransUNet (Pi et al. \cite{DL-592}), BMPTransNet (MIRSTD) (Jiang et al. \cite{DL-595}), LW-DSTransNet (Huang et al. \cite{DL-599-4}) } \\
\scriptsize{} &\scriptsize{RSLCANet [Swin Transformer] (Zhao et al. \cite{DL-599-50}), LPASU-SwinDet [Swin Transformer] (Ma et al. \cite{DL-599-361})} \\
\scriptsize{} &\scriptsize{Qformer (Yu et al. \cite{DL-599-282}), DifTransNet (Hu et al. \cite{DL-599-290}), CL-STFormer (Ma et al. \cite{DL-599-299})} \\
\scriptsize{} &\scriptsize{Nanetformer  (Xi et al. \cite{DL-599-216}),  CCI-TransNet (Cao and Shu \cite{DL-599-245}), MDNANet (Yin et al. \cite{DL-599-247})} \\
\hline
\end{tabular}}
\caption{Mamba State Space Model (SSM)-based Neural Networks and Transformers-based Neural Networks: An Overview (67 papers).}
\label{OverviewArchitectureM}
\end{center}
\end{table*}

In 2024, Chen et al. \cite{DL-260} develop a nested structure Mamba-in-Mamba (MiM-ISTD) which comprises Outer and Inner Mamba blocks to capture both global and local features. Local patches are considered as “visual sentences” and the Outer Mamba explores the global information. Each visual sentence is then decomposed into sub-patches as “visual words”. The Inner Mamba explores the local information among words in the visual sentence with low computational costs. By using both the visual word and visual sentence features, MiM-ISTD explores both global and local information. Experiments on the NUAA-SIRST and IRSTD-1k datasets show that  MiM-ISTD is $7 times$ faster than several SOTA methods and reduces GPU memory usage by $62.2\%$ on $2048 \times 2048$ images. \\

In 2024, \textit{in the context of infrared dense small target detection}, Xu et al. \cite{DL-269} develop a Fuzzy-Aware Nested Mamba (FANM) network. It leverages shallow, middle, and deep Mamba networks to fully exploit and integrate local and global multi-scale features. The extracted features are then passed to a Feature Pyramid Network (FPN) to balance the spatial resolution and semantic information of each feature map layer. After, the output features at each level are fed into the detection head. It comprises two branches: one for regressing the positions of arbitrarily distributed bounding boxes, and the other one for predicting the quality of the bounding boxes. The proposed fuzzy perceptual loss function allows to optimize these branches. Experiments on the DMIST dataset show that FANM outperforms several previous algorithms in the case of dense scenarios. \\

In 2025, Li et al. \cite{DL-75} propose a Hybrid Mamba–CNN UNet (HMCNet). Its hybrid architecture extracts the local features and model the global context. Experiments on the IRSTD-1k and MISTD \cite{DL-75} datasets show that HMCNet outperforms RLCM, MPCM, PSTNN \cite{SFD-RPCATensor-5} and RDIAN \cite{DL-33} with high-speed inference capabilities. Lu and Zhang \cite{DL-94} integrate a visual Mamba module into a lightweight CNN-based network which combines the ability of CNNs to model local features and a visual Mamba module captures long-distance global features. Therefore, this model extracts both local and long-distance global feature information. Experiments on the NUAA-SIRST and NUAA-SIRSTv2 datasets show that this model provides better accuracy and lower computational complexity. \\

In 2025, Yu et al. \cite{DL-266} propose a Spatial-BiDirectional Mamba Network (SBMambaNet) which exploits Spatial-BiDirectional Mamba Blocks (SBMBs) working in parallel with Skip Connections. SBMambaNe comprises two components. First, aSpatial-BiDirectional Cross Mamba (SCM) captures local spatial information in sequences and enhances the exchange between local spatial features and global channel data. It models bidirectional sequences for global context, and fuses features via gated mechanisms to strengthen long-range dependency and information flow. Second, a Mixed-Scale Spatial-Gate Feed-forward Network (MSGFN) improves feature discriminability via a multi-scale strategy and cross-space-channel information interaction. Second, a CNN-Mamba Channel-wise Fusion module (CMCFusion) guide the fusion of multi-scale channel information and pixel-level spatial relationships at the decoder stage. Experiments show that SBMambaNet outperforms most of the time model-based methods (PSTNN \cite{SFD-RPCATensor-5}, TLLCM \cite{SFD-LCM-19}, NRAM \cite{SFD-RPCAMatrix-10}, MSLSTIPT \cite{SFD-ML-2}, IPI, MLCL-Net \cite{DL-599-77}, data-based methods (MDvsFA \cite{DL-15}, LSPM, ACM \cite{DL-10}, ISNet \cite{DL-211}, GGL-Net, DNA-Net \cite{DL-12}, AGPCNet \cite{DL-21}), three hybrid algorithms combining CNN and ViT (IAANet \cite{DL-336}, Rkformer \cite{DL-392}, FATCNet) and MiM-ISTD. \\

In 2025, Zhang et al. \cite{DL-95} propose IRMamba which is based on an encoder-decoder architecture featuring Pixel Difference Mamba (PDMamba) and a Layer Restoration Module (LRM). First, PDMamba employs the intensity and directional information of pixel differences between scanning positions and their central neighborhoods into the state equation of the state space model (SSM). PDMamba improves target detail representation and avoids background interference by capturing local 2D dependencies from a global perspective. Second, ILRM employs the double-depth image prior into the iterative convergence algorithm, and uses the inter-layer interrelationships to reverse the separation of the target layer. Experiments conducted on the NUAA-SIRST, NUDT-SIRST, and IRSTD-1K datasets show the advantages of IRMamba compared to data-driven models (DNANet \cite{DL-12}, Dim2Clear \cite{DL-378}, ALCNet \cite{DL-11}, ACMNet), hybrid methods (RKformer \cite{DL-392}, SCTransNet \cite{DL-36}, IAANet), and traditional methods (Top-Hat). Jiang et al. \cite{DL-112} design an Efficient Adaptive Mamba Network (EAMNet\footnote{https://github.com/jiangjin1246/EAMNet}). First, a Mamba-based UNet architecture processes separated features in parallel with a low parameter count and computational cost. An Adaptive Filter Module (AFM) applies adaptive filtering by estimating filter parameters via an additional designed sub-network. A Shared Adaptive Spatial Attention Module (SASAM) allows a compact feature representation in generating spatial attention maps. Extensive experiments show that EAMNet  improves accuracy and efficiency compared to previous SOTA methods. \\

In 2025, Ye et al. \cite{DL-131} develop the Mamba Residual Network (MRNet). MRNet comprises two modules: the Mamba Residual Interaction Module (MRIM) and the Attention Enhanced Module (AEM). MRIM improves feature extraction progressively. AEM reduces missed detections and false alarms in complex scenes. A Center-Focused Strategy with an incentive and penalty mechanism allows to focus the network on the target’s center to improve detection accuracy. Experiments on the NUAA-SIRST, IRSTD-1 K, and NUDT-SIRST datasets show that MRNet outperforms existing IRSTD methods outperforms  single-frame methods (ACM, RDIAN \cite{DL-33}, ISTDU-Net \cite{DL-506}, UIU-Net \cite{DL-220}, DNA-Net \cite{DL-12}, SCTransNet \cite{DL-36}, MSHNet \cite{DL-41}, MTU-Net \cite{Applications-S-41}, MiM \cite{DL-260}) and multi-frame methods (SSTNet, Tridos, TSI-Net \cite{DL-96}). Liu et al. \cite{DL-263} design a Multi-Order U-shape Mamba (MOU-Mamba) to explore the suitability of 2D-Selective-Scan (SS2D) technique. First, a Multi-Order 2D-Selective-Scan (MO-SS2D) module gradually minimizes redundant information produced by SS2D operations via multi-order interactions. Second, a Local-Guided 2D-Selective-Scan (LG-SS2D) module improves target discernibility by constructing complementary local and global features at each interaction level. Experiments show that MOU-Mamba outperforms SOTA methods with an effective balance between accuracy and computational complexity. \\

In 2025, \textit{in the context of heterogeneous backgrounds}, Li et al. \cite{DL-264} propose a Mamba-driven approach named Spatial–spectral Mamba Interactive Learning (SMILE) network to deal with heterogeneous backgrounds. The reconstructed data couples Mamba’s flattened multidirectional scanning mechanism. Spatial Mamba and spectral Mamba collaboratively enhance the semantic features of small targets. A dual-path aggregation (DPA) network integrates the saliency and high-frequency attributes of small targets on order to balance detailed and contextual features. A Hybrid Representation Learning Module (HRLM) refines the local features to preserve the edge structure. Experiments show that  SMILE outperforms 14 previous algorithms. Li et al. \cite{DL-265} design a Multi-granularity representation learning with vision Mamba (MRL-Mamba) which is a nested structure with cross-fertilization of global and local information. Each layer of the top-level pyramid network embeds a tiny well-configured contextual pyramid block to extract fine-grained features of small targets. Then, a Mamba module restructures the feature maps to derive coarse-grained features of “visual sentences”. The fusion of contextual information and local feature provides the localization of small targets. An Asymmetric Convolution (AConv) is employed as a subtitute of the Depthwise Convolution (DWConv) in the Visual State Space (VSS) module. A regular convolution in each lateral connection of the nested pyramid network allows to alleviate the parameters and computation. Experiments show that  MRL-Mamba outperforms 12 previous methods. Xu et al. \cite{DL-268} design  a framework based on SAM2's hierarchical feature learning and a Mamba's selective sequence modeling. SAMamba\footnote{https://github.com/zhengshuchen/SAMamba} comprises three components. First, a Feature Selection Adapter (FS-Adapter) is employed for natural-to-infrared domain adaptation via dual-stage selection (token-level with a learnable task embedding and channel-wise adaptive transformations). Second, a Cross-Channel State-Space Interaction (CSI) module is used for global context modeling with linear complexity using selective state space modeling. Third, a Detail-Preserving Contextual Fusion (DPCF) module adaptively combines multi-scale features with a gating mechanism to balance high-resolution and low-resolution feature contributions. Experiments on the NUAA-SIRST, IRSTD-1k, and NUDT-SIRST datasets show SAMamba outperforms ACM \cite{DL-10}, U-Net, ISNet \cite{DL-211}, FC3-Net \cite{DL-335}, UIU-Net \cite{DL-220} and HCFNet in the case of heterogeneous backgrounds and varying target scales.  Li and Wang  \cite{DL-273} propose a Mamba-driven approach dubbed Spectrum-Assisted Mamba Network (SAMNet). First, a reconstructed data couples Mamba’s flattened multidirectional scanning mechanism. Second, a spatial Mamba and a spectral Mamba collaboratively enrich the semantic features of small targets. Third, a Dual-Path Aggregation (DPA) network employs the saliency and high-frequency attributes of small targets. An Hybrid Representation Learning Module (HRLM) refines the local features to inscribe the intact edge structure. Experiments show that SAMNet outperforms 14 previous algorithms. \\

In 2026, \textit{in the context of heterogeneous backgrounds}, Sui et al. \cite{DL-320} design a  Stream Alignment Mamba for Motion In-frared Small Target Detection (SAMamba). First, a Stream Mamba Block based on a dynamic state space model employs bidirectional spatiotemporal scanning to explore inter-frame relationships in sequential images preserving spatial correlations and exploitin the causality of temporal cues for global background representation. Second, a Cross-Frame Feature Enhancement Block aligns feature streams by computing local responses between neighboring and reference frames. Third, a H-erarchical Feature Alignment Block with adaptive adjustment strategies and position-aware mechanisms handle spatial misalignment during upsampling. Experiments on the IRDST, IRSatVideo-LEO, and TSIRMT datasets show that SAMamba provides significant advantages compared to previous SOTA methods. Tang et al. \cite{DL-321} propose an orientation-guided Mamba framework (TAMamba) with a Texture-Aware State-Space Module (TSSM) and a Deep Focus Algorithm (DFA). First, TSSM employs explicit physical orientation priors to provide dynamic texture mask. Second, DFA  enhances target representations by performing selective normalization that enhances sparse small-target signals while avoiding background interference. Experiments on the NUAA-SIRST, IRSTD-1 k, and NUDT-SIRST datasets show that TAMamba provides SOTA  performance with only 2.3 M parameters, reducing parameters by $51\%$ while maintaining superior accuracy and robustness in complex scenarios. \\

In 2026, Zhang et al. \cite{DL-116} design Mamba-IRSTD\footnote{https://github.com/Fidods/Mamba-IRSTD} based on an encoder–decoder architecture. In the encoder, a dual-residual-path Mamba block combines the complementary advantages of CNNS and Mamba-based state space models for global context modeling. A multi-scale feature adaptive aggregation block enhances the representation of features across multiple scales. In the decoder, an edge-aware attention module is coupled with a multi-scale shape-adaptive fusion head to focus explicitly on edge features Experiments show that Mamba-IRSTD outperforms AGPC, FDPF,	ViAMLN,	HFMNet and WaveTD in terms of FLOPs and fps. Wu et al. \cite{DL-212} develop a method guided by Causal Predictive state space and reconstruction refinement named CPMamba. During feature extraction, inherent discriminative power is done by separating predictable low-rank background components from anomalous small targets. The Causal Predictive State–Space Model (CP-SSM) exploits ”discriminability” contextual information to guide attention. A Feature Reconstruction Refinement Block (FRRB) use a detail-oriented pathway for structural recovery and a semantic-aware pathway for noise suppression. Extensive experiments on the NUDT-SIRST, IRSTD-1K, and SIRST-UVAB datasets show that CPMamba provides most of the time beter results than PConv \cite{DL-104}, ACMNet, UIU-Net\cite{DL-220}, DenseNet and SCTransNet. Li et al. \cite{DL-250} propose a Bidirectionally Driven Saliency Fusion network (BDSFusion\footnote{https://github.com/kyrietop11/BDSFusion}). First, a feature extraction module named CSA-CM based on a Conv-Mamba (CM) module and a Channel Self-attention (CSA) module allows to reduce the channel redundancy of the Mamba module. Second, a spatial-frequency cross-domain feature integration module (SFCIM) exploits the complementary characteristics of spatial and frequency-domain information from source images. A task-driven joint loss function combines intensity loss, Haar wavelet mask loss, and background residual loss to guide the fusion network. Experiments on  MWIR and LWIR images show that BDSFusion outperforms ACM-Net, DNA-Net \cite{DL-12}, IR-TransDet \cite{DL-19}, and ILNet \cite{DL-70}.  \\

In 2026, \textit{in the context of multi-frame detection}, Qiu et al. \cite{DL-261} propose a Spatial–Temporal (STLMamba) framework based on long-range dependency modeling with local feature sensitivity and guided by infrared video and frequency priors. First, an Adaptive Multidirectional Frequency Enhancement (AMFE) module leverages spatial frequency cues to enhance targets. Second, a Efficient Local-motion Mamba (ELM) with a learnable local dynamic Gaussian motion (LDGM) mechanism captures motion patterns and suppresses background noise. Experiments show that STLMamba provides 43.14 fps  and 49.22 fps for the NUDT-MIRSDT and TSIRMT, respectively. Xie et al. \cite{DL-262} develop a Trajectory-Aware Mamba Propagation Network (TAPM-Net) which  models the spatial diffusion behavior of target-induced feature disturbances. TAPM-Net comprises two modules: a Perturbation-guided Path Module (PGM) and a Trajectory-Aware State Block (TASB). First, PGM constructs perturbation energy fields from multi-level features and extracts gradient-following feature trajectories that reflect the directionality of local responses. The resulting feature trajectories are fed into a Mamba-based state-space unit that models dynamic propagation along each trajectory while incorporating velocity-constrained diffusion and semantically aligned feature fusion from word-level and sentence-level embeddings. TAPM-Net tackles anisotropic, context-sensitive state transitions along spatial trajectories while keeping global coherence at low computational cost. Experiments on the NUAA-SIRST and IRSTD-1K datasets show that TAPM-Net achieves SOTA performance. \\

\section{Transformers-based Neural Networks}
\label{Transformers} 
Table \ref{OverviewArchitectureM} shows an overview of the existing methods based on a Transformers-based Neural Networks. \\

In 2022, Chen et al. \cite{DL-18} design a Hierarchical Overlapped Small Patch Transformer (HOSPT) to encode multi-scale features from the single-frame image. A Top-down Feature Aggregation Module (TFAM) fuses features from adjacent scales. Experiments show that IRSTFormer outperforms most of the time DNA-Net \cite{DL-12}, AGPCNet \cite{DL-21}, LSPM and MDvsFA. Zhang et al. \cite{DL-392} provides a Runge-Kutta Transformer (RKformer) with random-connection attention. The Parallel Encoder Block (PEB) of the transformer and convolution take their advantages in long-range dependency modeling and locality modeling. Two PEBs are stacked with several residual connections under the guidance of RK methods for solving ODE as the basic encoder block. the Random-Connection Attention (RCA) block learns sparse attention through random connections during training. Experiments show that RKformer provides SOTA performance. \\

In 2023, Lin et al. \cite{DL-19} propose IR-TransDet which combines advantages of the CNN and the Transformer. First, an Efficient Feature Extraction Module (EFEM) employs depthwise convolution and Point-Wise Convolution (PW Conv) to capture the features of the target. Second, a Residual Sim Atrous Spatial Pyramid Pooling (ASPP) module enhances the edge information of the target. Third, an IR-Transformer module employs a self-attention mechanism to exploit the relationship between the global image, the target, and neighboring pixels. Experiments show that IRTransDet provides SOTA performance. Liu et al. \cite{DL-20} develop a method with the transformer. First, the self-attention mechanism of the transformer learns the correlation of image features in a larger range. Second, a feature enhancement module learns discriminative features. Third, a decoder with the U-Net-like skip connection operation allows to capture more information. Experiments show the proposed method offers SOTA performance with a stronger generalization ability and better robustness to noise.  Li et al. \cite{DL-276} propose an
Infrared Small Target Transformer Network (IST-TransNet). An Anti-aliasing Contextual Feature Fusion (ACFF) module improves the translation invariance during contextual feature fusion for locating infrared small targets. A Spatial and Channel Attention (SCA) module allows to To suppress complex background interference and highlight the features of infrared small target. A vision-transformer branch (VTB) suppresses the interference of local flash elements by introducing non-local correlation features. Experiments show that IST-TransNet outperforms the model-driven methods (IPI, NRAM, PSTNN, RIPT) and data-driven methods (DNANet \cite{DL-12}, AGPCNet). \\

In 2023, \textit{in the context of tiny ship detection}, Wu et al. \cite{DL-340} design a Multilevel TransUNet (MTU-Net\footnote{https://github.com/TianhaoWu16/Multi-level-TransUNet-for-Space-based-Infrared-Tiny-ship-Detection}). It consists of a vision Transformer (ViT) CNN hybrid encoder to extract multilevel features. First, local feature maps are extracted by several convolution layers and then fed into the multilevel feature extraction module to capture long-distance dependency. A Copy–Rotate–Resize–Paste (CRRP) data augmentation approach accelerates the training phase. A FocalIoU loss achieves both target localization and shape description. Experiments on the NUDT-SIRST-Sea dataset show that MTU-Net outperforms model-driven and previous data-driven model in SIRST methods in terms of probability of detection, false alarm rate, and intersection over union. \\

In 2023, Peng et al. \cite{DL-277} design CourtNet\footnote{https://github.com/PengJingchao/CourtNet} to dynamically balance the precision and recall rates. CourtNet  consists of three sub-networks. First, a  prosecution network improves the recall rate. Second, a defendant network increases the precision rate. Third, a jury network weights their results to adaptively balance the precision and recall rates. CourtNet employs the structure of Transformers, whose feature resolution remains unchanged. The prosecution network employs a densely connected structure, which can prevent small targets from disappearing in the forward propagation. A fine-grained attention module performs attention inside patches to locate the small targets. Experiments on the ISTD, MFIRST and SIRST datasets show that  CourtNet outperforms traditional and data-based methods with a detection speed of 60.61 fps and the best score of 0.73 in SIRST and 0.62 in MFIRST  \\

In 2024, Yuan et al. \cite{DL-36} design a Spatial-channel Cross Transformer Network (SCTransNet\footnote{https://github.com/xdFai/SCTransNet}) which exploits Spatial-Channel Cross Transformer Blocks (SCTBs) on top of long-range Skip Connections (SKs). In SCTBs, the outputs of all encoders are interacted with cross transformer to provide mixed features, which are redistributed to all decoders to reinforce semantic differences between the target and clutter at full levels. SCTB comprises two modules. First, a spatial-embedded single-head channel cross-attention (SSCA) allows to exchange local spatial features and full-level global channel information to suppress ambiguity among the encoders and facilitate high-level semantic associations of the images. Second, a Complementary Feed-forward Network (CFN) enhances the feature discriminability through a multiscale strategy and cross-spatial-channel information interaction to promote transfer. SCTransNetencodes the semantic differences between targets and backgrounds to boost its internal representation. Experiments on the  NUDT-SIRST, NUAA-SIRST, and IRSTD-1K, datasets show  that SCTransNet outperforms several previous SOTA methods. Huang et al. \cite{DL-46} develop a Local Motion Aware Transformer Network (LMAFormer\footnote{https://github.com/lifier/LMAFormer}). First, a local motion-aware spatiotemporal attention mechanism aligns and enhances multi-frame features to extract local spatiotemporal salient features of targets while suppressing interference from moving backgrounds. Second, a multiscale fusion transformer encoder computes self-attention weights across and within scales during encoding to obtain multiscale correlations among different regions of temporal images, enabling motion background modeling. Third, a multi-frame joint query decoder is employed. The shallowest feature map is mapped to initial query weights. Experiments on the NUDT-MIRSDT, IRDST, and TSIRMT datasets show that LMAFormer provides similar results than previous SOTA methods. \\

In 2024, \textit{in the context of hybrid models}, Yang et al. \cite{DL-278} design a Feature Adaptive Transformer and CNN network (FATCNet\footnote{https://github.com/miliner/FATCNet}). It is a dual branch decoder structure network that employs CNN and Transformer to obtain local and global features. An attention guided adaptive fusion block integrates global and local feature information, suppress semantic gaps, and dynamically fuse features based on network learning. The network encoder incorporates the hierarchical attention module. Experiments on the BSIRST-v2 dataset show that FATCNet outperforms previous SOTA algorithms in terms of probability of detection, F1 score, and intersection over union. 

In 2024, \textit{in the context of heteregeneous backgrounds},Wu et al. \cite{DL-279} propose a a Swin Transformer and multiscale atrous spatial pyramid pooling Network (STASPPNet\footnote{https://github.com/hwuscut/STASPPNet}). First, A Swin-convolution (Swin-Conv) backbone enhances feature extraction Second, a Multiscale Atrous Spatial Pyramid Pooling module (MASPPM) exploits atrous convolutions and max pooling layers to enhance feature representation. Third, a Multiscale Feature Detection Fead (MFDH) allows the detection of targets  with as few as a dozen pixels and enhances the performance for weak infrared targets. Experiments show that STASPPNet offers an average accuracy of $96.9\%$ for mAP50 and $49.7\%$ for mAP50-95 on the NUAA-SIRST dataset. \\

In 2025, Hu et al. \cite{DL-69} provide a Dynamic Attention Transformer Network (DATransNet\footnote{https://github.com/greekinRoma/DATransNet}). DATransNet employs the Dynamic Attention Transformer (DATrans), simulating central difference convolutions (CDC) to extract gradient features. A Global Feature Extraction Module (GFEM) provides a comprehensive perspective to prevent the network from focusing solely on details while neglecting the global information. Experiments show that DATransNet outperforms most of the time ACM \cite{DL-10}, UIU-Net \cite{DL-220}, ALCNet \cite{DL-11}, AGPCNet \cite{DL-21}, RDIAN \cite{DL-33}, DNA-Net  \cite{DL-12} and SCTransNet. Zhu et al. \cite{DL-71} develop a Spatial-Temporal-Channel Collaborative feature learning network (STC\footnote{https://github.com/UESTC-nnLab/STC}). STC consists of three distinct transformer encoders which extract spatial, temporal and channel information, respectively. Furthermore, a transformer decoder fuses the three attention features. A Semantic-Aware positional encoding method for video clips incorporates temporal information into positional encoding and is scale-invariant. Experiments show that STC outperforms SCTransNet. Liu et al. \cite{DL-74} propose a Semantic Token transformer Patch Spatial Attention Network (STPSA-Net). Images are represented as compact semantic tokens by a Semantic Token Transformer (STT) module. STT models spatiotemporal context to refine the original features and enhance the representation capability of small target features. The Patch Spatial Attention Module (PSAM) divides extracted features into patches and integrates spatial and semantic information to restore spatial information. Experiments on the SIRST, MFIRST, and NUDT-SIRST datasets show the STPSA-Net provides superior performance compared with previous SOTA approaches. \\

In 2025, \textit{in the context of hybrid models}, Li et al. \cite{DL-101} design an Hybrid Spatial–channel Sparse Transformer Network (HSTNet\footnote{https://github.com/juranccc/HSTNet}). First, a hybrid Spatial–Channel Sparse Transformer (SCST) modulesparsely models the relationship between targets and background. Second,  a Multiscale Detail Enhancement (MSDE) module preserves small target details during the feature compression process. Third, a Scale-Location aware Joint (SLJ) loss improves target perception at various scales and locations. Experiments on the NUAA-SIRST, IRSTD-1k, and IRSTD-Large datasets show that HSTNet offers SOTA performance. Luo et al. \cite{DL-62}) propose a 3D Spatial-Temporal knowledge aware-based Unsupervised Network (STUTD). First, a Spatial-Temporal Swin Transformer Block (ST-STB) employs a Multiscale Feature Extraction and Aggregation (MFEA) module for feature extraction. Second, a Variational Autoencoder (VAE)-style background reconstruction module with a multi-head gating mechanism allows background reconstruction. Third, a sparse cardinality selection of residuals performs element-wise filtering on the residuals between the original tensors and the reconstructed background to provide a target tensor. Experiments show that STUTD offers satisfactory overall performance and real-time performance. \\

In 2025, \textit{in the context of diverse scenarios (day/night variations, sky/maritime/ground environments)}, Qian et al. \cite{DL-133} design a meta-learned framework named IrisNet which dynamically adapts detection strategies to the input infrared image status. A dynamic mapping is employed between infrared image features and entire decoder parameters through an image-to-decoder transformer. Second a parameterized decoder is represented as a structured 2D tensor preserving hierarchical layer correlations and allows the transformer to model inter-layer dependencies via self-attention while providing adaptive decoding patterns through cross-attention. Third,a high-frequency components supplement target-position and scene-edge information. Experiments on the NUDT-SIRST, NUAA-SIRST, and IRSTD-1K datasets show that IrisNet  outperforms SOTA algorithms (ACM \cite{DL-10}, ALCNet \cite{DL-11}, ISNet \cite{DL-211}, RDIAN \cite{DL-33}, DNA-net \cite{DL-12}, UIU-Net \cite{DL-220}, ISTDU, Dim2Clear, L2SKNet \cite{DL-370}) and transformers based methods (RKformer \cite{DL-392}, TCI-Former, SCTransNet). Li et al. \cite{DL-239} propose an attention-based nested network with a bottleneck Swin Transformer (BSTNA-Net). A residual-attention structure incorporating serial channel and spatial attention is combined with a nested U-Net architecture. A Wavelet Down-sampling Block supersedes traditional pooling layers. A Bottleneck Swin Transformer Block processes deep feature maps, and a shifted window self-attention mechanism enhances the global feature extraction ability of the deep neural network. Experiments on the IRSTD-1k and NUDT-SIRST datasets show that BSTNA-Net outperforms 
ACM \cite{DL-10}, ALCNet \cite{DL-11}, ISNet \cite{DL-211}, DNA-Net \cite{DL-12}, ISTDU-Net, UIU-Net \cite{DL-220}, SCTransNet \cite{DL-36} and HDNet. Shen et al. \cite{DL-507}) propose LVNet that redefines low-level feature learning in hybrid frameworks for multi-frame IRSTD. Because a standard linear patch embeddings in Vision Transformers are insufficient to capture the scale-sensitive local features critical, a a multi-scale CNN front-end t models local features by exploiting the local spatial bias of convolution. Second, a U-shaped video Transformer for multi-frame spatiotemporal context modeling captures the motion characteristics of targets. Experiments on the IRDST and NUDT-MIRSDT datasets show that LVNet outperforms LMAFormer. \\

In 2025, Li et al. \cite{DL-350} design a U-shaped segmentation network named Knowledge-Embedded CSwin-UNet (KECS-Net\footnote{https://github.com/Lilingxiao-image/KECS-Net}). First, KECS-Net incorporates a CSwin transformer module into the encoder of the UNet backbone allowing the extraction of multiscale features Second, a Multiscale Local Contrast Enhancement Module (MLCEM) employs hand-crafted dilated convolution operators to enhance locally salient target responses and suppress background noise. Third, a Slicing-Aided Hyper-Segmentation (SAHS) method resizes and rescales the output image. Experiments show that the proposed KECS-Net outperforms previous SOTA methods in both quantitative metrics and visual quality. \\

In 2026, \textit{in the context of hybrid models}, Feng et al. \cite{DL-237} design a Global Feature Integration Mamba–Transformer Hybrid Network hybrid (GMTNet) which fuses global semantics and local details. GMTNet consists of three modules. First, a Directional State-space Transformer (DiST) module combines direction-aware long-range modeling with fine local cues. Second, a Wavelet-based Multidimensional Lossless Downsampling (WMLD) module preserves high-frequency information during resolution reduction. Thids, a DualAug module  combines geometric and photometric perturbations for robustness. Experiments on the IRSTD-1K, NUAA-SIRST, and NUDT-SIRST show that GMTNet outperforms previous SOTA methods.Lin et al. \cite{DL-193} propose a Hybrid Network with Geometric Enhancement and Sparse Extraction (HSHNet). Jiang et al. \cite{DL-143} design a Cross-layer Relative Position Embedding transformer detection Network (CRPE-Net) which integrates location awareness from multiple perspectives. First, a Spatial Fine-grained Interaction Pyramid (SFIP) at the token level completes multi-scale interaction by dividing feature token groups to capture fine-grained small target information along the spatial dimension. Second, the absolute-position embeddings is replaced by a Cross Relative-Position Embeddings (CRPE) to enhance the cross-layer feature information. Third, a Multi-Head Intensive Prediction learning strategy (MHIP) enhances the detection accuracy by using multi-head detection. Experiments show that CRPE-Net outperforms ALCNet \cite{DL-11}, ISTDU, RDI-Net, and UIU-Net. \\

In 2026, \textit{in the category of detection methods}, Chen et al. \cite{DL-136} design a method based on Local adaptive perception-guided sharpening and improved RT-DETR (LCR-RTDETR) that combines adaptive sharpening preprocessing with an improved real-time detection transformer (RT-DETR). First, a Local Adaptive Perception-Guided Sharpening data augmentation strategy (LAPGS) improves feature expression quality by enhancing edge details and local contrast. Second, a Contrast-aware and value-Range-guided Self-Attention mechanism (CRSA) integrates with an Adaptive Image Feature fusion module (AIFI) to obtain the CRSA-AIFI. Third, a Relative Position-based Attention enhancement module (RPAC3) and a Context-Cross feature Fusion Module (RPAC3-CCFM) incorporating RPAC3 enhance spatial modeling and contextual expression abilities. Experiments show that LCR-RTDETR provides mAP@50 of $86.7\%$, $88.1\%$, and $99.5\%$ on the SIRST, IRSID-1 K, and NUDT-SIRST datasets, respectively. The parameter count and computational load are reduced by $14.4\%$ and $10.9\%$, respectively. The inference speed on the Jetson Orin NX embedded platform is improved by 5.8 FPS, 3.4 FPS, and 5.7 FPS. Cui et al. \cite{DL-192} develop a cross-scale structural reconstruction detection network (EUSS-DETR) based on Real-Time Detection Transformer (RT-DETR). First, a hierarchical multi-representation adaptive module (ESM-Block) enhances structural consistency for weakly textured targets by coupling local details with deep-level semantics. Second, a Spatial Shift Modulation Unit (UCSM) establishes cross-channel directional spatial interactions to enhance weak target responses. A cross-scale Structural alignment module (Saf-Cat) mitigates structural degradation due to scale variations and edge blurring. A Dynamic Scale Aggregation Module (SC3D) integrates targets at different resolutions adaptively through scale sequence modeling. These four modules allow to model chain, spanning from structural enhancement to scale coordination. Experiments show that EUSS-DETR achieves mAP@0.5 of $90.7\%$, $98.0\%$, and $98.4\%$ on three public datasets. \\

In 2026, Hu et al. \cite{DL-244} design a Dual-Transformer Feature Enhancement Network (DTFE-Net) which integrates spatial and frequency domain representations. It consists of three  components: a Dual-Transformer Frequency Selector (DTFS), a Spatial Feature Selector (SFS), and a Spatial-Augmented Frequency Representation (SAFR). First, DTFS uses a dual-transformer mechanism exploiting both wavelet transform and fast Fourier transform. Second, SFS extracts effective spatial features from both local and global perspectives. Third, SAFR refines the spatial–frequency representation via a fine-grained fusion to improve target discernibility. Experiments on the NUAA-SIRST, IRSTD-1K, and NUDT-SIRST datasets show that DTFE-Net provides better performance than previous algorithms. \\

In 2026, \textit{to avoid the dilution of the representation of subtle local motion patterns}, Zhang et al. \cite{DL-599-74} propose a Spatial-Semantic Guided Motion Perception (SGMP). First, SGMP exploits spatial features to obtain salient semantic query vectors and value maps for extracting potential semantic priors of targets.   Second, SGMP employs a Transformer encoder to model long-range motion features, aligning global motion context with local semantic cues across modalities. Thus, 
a lightweight and efficient detection framework (SGMP-UNet)  exploits motion aware
representations for end-to-end small target detection. Experiments on the NUDT-MIRSDT, IRSDT and TSIRMT datasets show that SGMP-UNet consistently outperforms LMAFormer.

In the following sections, we review adversarial networks, language models and graph/hypergraph models The reader can refer to Table \ref{OverviewDL-2} for an overview.

\section{Adversarial Networks}
\label{AdversarialNetworks}
\begin{enumerate}
\item \textbf{Context Adversarial Networks:}
In 2019, Wang et al. \cite{DL-15} propose a generative adversarial learning network in which the generator is divided into two subtasks: miss detection and false alarm.  This method is named MDvsFA-cGAN. 
\item \textbf{Generative Adversarial Networks:} In 2020, Zhao et al. \cite{DL-4} design a GAN model to learn the features of targets and directly predict the intensity of targets. More specifically, the target is recognized and reconstructed by the generator with an U-Net architecture. A five-layer discriminator allows to enhance the data-fitting capability of generator. A $l_2$ loss is employed into adversarial loss for the localization. Thus, the detection problem is formulated as an image-to-image translation problem implemented by GAN. Experiments show that this method on various backgrounds and targets achieves better intersection over union (IoU) values than state-of-the-art methods.
To tackle the inadequate modeling of the dynamic degradation process, Jin et al. \cite{DL-333} develop a Dynamic Adversarial Enhancement Network (DAENet). First, a degradation-enhancement dynamic adversarial pretraining strategy is employed to improve the model’s adaptive capacity to nonstationary variations in thermal radiation. More specifically, this method simulates the physical degradation process of infrared imaging to inject degradation priors, and incorporates a dynamic adversarial mechanism. Second, a frequency-gradient collaborative enhancement module allows to extract frequency-domain directional texture features via the coupling of wavelet decomposition and Gabor filtering. This mechanism allows to introduce an adaptive gradient enhancement branch to reinforce spatial directional gradients and  boosts the saliency of the discriminative features of weak targets. Experiments show that DAENet outperforms state-of-the-art algorithms in complex environments.
\end{enumerate}

\section{Mixture-of-Experts (MoE)}
\label{MoE}
To break through single-domain learning by implementing only one universal detector simultaneously on multiple datasets, Duan et al. \cite{DL-196} design a cross-domain joint learning task framework with prototype-guided Mixture-of-Experts (CoMoE). More specifically, a hyperspherical prototype learning adaptively maintain both domain-specific prototypes and global prototypes improving the cross-domain feature representation. A domain-aware Mixture-of-Experts with Top-$K$ routing strategy allows to select the optimal domain experts. Finally, an adaptive cross-domain feature modulation with noise-guided contrastive learning is employed to enhance the cross-domain feature alignment. \\

In 2026, Du et al. \cite{DL-599-70} propose a Frequency-Adaptive Mixture-of-Experts Network (FAME-Net) which models the frequency heterogeneity of infrared scenes. First, a Cross-Frequency Mixture-of-Experts (CF-MoE) module constructs expert banks to model high-frequency structural cues and low-frequency thermal contexts. Second, a Mutual-Guided Routing strategy exploits cross-frequency information for joint expert selection. Thus, the network adaptively activates appropriate processing patterns under diverse target morphologies and background complexities. Third, a Semantic-Guided Gating (SGG) mechanism  suppresses intrinsic background noise in skip connections. A Boundary-aware Contrastive (BC) loss enhances pixel-level discriminability in thermally blurred boundary regions. Experiments demonstrate that FAME-Net achieves SOTA performance in detection accuracy and boundary recovery across multiple benchmarks.

\section{Language Models}
\label{LanguageM} 
Language models allow to add contextual information via additional text in addition of the visual information provided by the infrared data obtained by the sensors.

\subsection{Large Language Models (LLMs)}
The previous deep neural networks methods typically rely solely on image modalities which fail to fully capture contextual information and lead to limited detection accuracy and adaptability in complex environments. To address these limitations, Zhang et al. \cite{LLM-1} design a Segment Any Infrared Small Target (SAIST) method which employs textual information with image modalities. SAITS consists of two main parts, namely Scene Recognition Contrastive Language-Image Pretraining (SR-CLIP) and CLIP-guided Segment Anything Model (CG-SAM). First, SR-CLIP provides a set of visual descriptions through object-object similarity and object-scene relevance embedding them into learnable prompts to refine the textual description set. Second, CG-SAM employs the prompts generated by SRCLIP to guide the Mask Decoder in learning prior knowledge of background features, while incorporating infrared imaging equations to improve small target recognition in complex backgrounds and reduce the false alarm rate. Experiments show that SAIST outperforms ISNet \cite{DL-211}, GCINet, UIU-Net \cite{DL-220}, DNA-Net \cite{DL-12}, Dim2Clear \cite{DL-378}, ALCNet \cite{DL-11} and ACMNet on the MIRSTD dataset \cite{LLM-1}. In a similar work, Zhang et al. \cite{LLM-2} propose a Multimodal Vision-Language Segment Anything Model (MIRSAM). Given the differences in noise and structural information between infrared and natural images, Zhang et al. \cite{LLM-2} adapt Segment Anything Model (SAM) by employing a Contourlet Denoising Adapter Module (CDAM). This module allows to suppress noise during feature extraction and encoding. To incorporate textual information, Zhang et al. \cite{LLM-2} use a text encoder of Contrastive Language-Image Pre-Training (CLIP) to convert text into high-dimensional feature vectors. These vectors are employed as prompts to extract relevant details from the features. Experiments on the IR-TXPair \cite{LLM-2} dataset show that MIRSAM outperforms MDvsFA \cite{DL-15}, ACM \cite{DL-10}, ALCNet \cite{DL-11} and ISNet \cite{DL-211}. \\

In 2025, Huang et al. \cite{LLM-3} propose a semantic text method to guide infrared small target detection (Text-IRSTD). First, a fuzzy semantic text prompt allows to accommodate ambiguous target categories. Second, a Progressive Cross-modal Semantic Interaction Decoder (PCSID) is employed to facilitate information fusion between texts and images. Experiments on the  FZDT dataset \cite{LLM-3} show Text-IRSTD achieves better detection performance and target contour recovery than DNA-Net \cite{DL-12}, UIU-Net \cite{DL-220}, RDIAN \cite{DL-33}, DMFNet \cite{DL-599-261} and SCTransNet \cite{DL-36}. In an other work, Chib and Singh \cite{LLM-4} propose a Language-Guided attention Networks (LGNet) derived from the language prior to improve the model's ability presenting an approach that combines textual information with image data. Using the GPT-4 vision model,  Chib and Singh \cite{LLM-4} obtain text descriptions that provide the locations of small targets employing prompt engineering to ensure accuracy.\\

In 2026, Zhu et al. \cite{LLM-5} provide a Text-guided Iterative Polymerization with Prototyping (TIPP) framework for incremental few-shot learning. TIPP exploits text features to obtain modulation factors to iteratively guide the polymerization of spatiotemporal features. Using statistical features, TIPP learn one unique prototype to identify one task. Experiments on the IRDST, DAUB \cite{Dataset-1}, and ITSDT datasets show the effectiveness and superiority of TIPP. 

\subsection{Vision-Language and Vision Foundation Models (VLMs/VFMs)}
The conventional deep neural networks (CNNs, GANS, Tranformers) methods present the following weaknesses \cite{VLM-1,VLM-2,VLM-3,VLM-6}:
\begin{enumerate}
\item The global context and semantic modeling are insufficient because they focus on architectural modifications and local clue amplification. Thus, their ability to characterize global context and high-level semantic relationships is limited with unstable performance in the presence of complex backgrounds. 
\item The trade-off between performance and efficiency is limited because  they employ many parameters generating computational overhead  and making them not deployable for real-time or resource-constrained scenarios. 
\item They focus on the pure-vision features usually by fully-supervised learning requiring extensive high-cost manual annotations. 
\item They almost have not concerned the potentials of multi-modal (e.g., vision and text) learning.
\item Spatial feature Submergence: Hierarchical backbones progressively aggregate information through deep token mixing and large effective receptive fields, leading to missed detections.
\item Temporal Appearance Ambiguity: Transformers and memory-attention-based architectures perform association by matching appearance features across frames, which is effective when objects have distinctive semantic attributes but infrared small targets are nearly textureless.
\item Motion representations have been proven to be one of the most crucial feature kinds for infrared small target detection. Conventional deep neural networks methods can only capture coarse motion representations from temporal domain.
\end{enumerate}
To address the limitations 1 and 2, Yu et al. \cite{VLM-2} propose in 2025 a frozen representations from VFMs designing a Foundation-Driven Efficient Paradigm (FDEP) which adapts to existing encoder-decoder-based methods and increases accuracy. More specifically, a Semantic Alignment Modulation Fusion (SAMF) module achieves dynamic alignment and deep fusion of the global semantic priors from VFMs with task-specific features. To avoid the inference time burden, a Collaborative Optimization-based Implicit Self Distillation (CO-ISD) strategy allows implicit semantic transfer between the main and lightweight branches through parameter sharing and synchronized backpropagation. Experiments show that FDEP improves the accuracy detection of ACM \cite{DL-10}, ALCNet \cite{DL-11}, MLCL-Net \cite{DL-599-77}, ALCL-Net, DNA-Net  \cite{DL-12}, GGL-Net, UIU-Net \cite{DL-220} and MSDA-Net using Holistic SIRST Evaluation (HSE) series accuracy metrics. \\

In 2026, Duan et al. \cite{VLM-1} adresses the limitations 3 and 4 by designing a semi-supervised vision-language (SeViL) framework with adaptive text prompt guiding. SeViL takes text prompts as prior knowledge to adaptively enhance target regions and then filter the low-quality pseudo-labels generated on unlabeled data. An adaptive cross-modal masking strategy allows to align text and vision features, promoting cross-modal deep interactions. Experiments on three public datasets namely DAUB \cite{Dataset-1}, ITSDT-15K and IRDST show that SeViL outperforms other semi-supervised methods and similar performance to fully-supervised methods, with $10\%$ labeled training samples. \\

To address the challenges 5 and 6, Xu et al. \cite{VLM-3} design  a Spatiotemporal Physics-Informed Representation for Infrared Targets (SPIRIT) framework that adapts semantics-oriented VFMs to IRSTD by employing physically constraints into feature representation and attention modules. These modules are lightweight plugins that are integrated into different VFM backbones with minimal changes. More specifically, Xu et al. \cite{VLM-3} employ a Physics-Informed Feature Refinement (PIFR) that is used as a differentiable physical filter inside the encoder for the spatial aspect. PIFR approximates rank-sparsity decomposition in the feature domain, suppressing low-rank background components and enhancing sparse target-like responses. For temporal aspect, a Prior Guided Memory Attention (PGMA) mechanism injects a feasibility prior into memory cross-attention. SPIRIT operates in both single-frame and video cases. When temporal information is unavailable, SPIRIT is reduced to a spatial detector making it flexible for deployment under dynamic operational requirements. Experiments show that SPIRIT outperforms ACM \cite{DL-10}, AGPCNet \cite{DL-21}, DNA-Net \cite{DL-12} and UIU-Net \cite{DL-220}. \\

To address the limitation 7, Shen et al. \cite{VLM-4} design a vision-language framework with Motion Prior Knowledge Learning (MoPKL\footnote{https://github.com/UESTC-nnLab/MoPKL}) which uses homogeneous language descriptions, formatted for moving targets, to directionally guide vision channel learning motion prior knowledge. The motion of infrared small targets is refined by graph attention to obtain more fine motion representations. Experiments on ITSDT-15K and IRDST datasets show that MoPKL outperforms several previous algorithms. In a further work, Shen et al. \cite{VLM-5} develop an an improved version of MoPKL named iMoPKL which also uses homogeneous language descriptions to directionally guide vision channels to learn the motion prior knowledge of targets. First, iMoPKL learns the distribution of target motion reconstruction corresponding to the language description as a type of prior knowledge. Second, the motion of infrared small targets is refined by motion-relation learning to obtain more fine motion representations. Experiments on ITSDT-15K, DAUB-R, and IRDST-H datasets show that iMoPKL outperforms MoPKL.

\section{Graph/Hypergraph Learning Methods}
\label{GraphLearning} 

\subsection{Preliminaries}
Graph Neural Networks\footnote{https://sites.google.com/view/gsp-website/graph-neural-networks/graph-neural-networks}\cite{GM-Theory-1} is a deep learning method that takes graph data as input. Graph Neural Networks can be divided into Convolutional Graph Neural Networks, Recurrent Graph Neural Networks, Graph Auto-Encoders, and Spatial-Temporal Graph Neural Networks. Convolutional. Graph Neural Networks can be decomposed into two branches:  Spectral-based GNNs where graph spectral theory was applied to learn node representations, and spatial-based GNNs based on local-dependence assumption of graph structured data that graph convolution is defined in the spatial domain as aggregating and transforming local information. Thus, spectral-based GNNs extends classical signal processing to graph data, designing graph filters in the frequency domain by borrowing from graph signal analysis, and realizing graph convolution whereas spatial-based GNNs describe the pairwise relations between objects. Transductive and inductive learning schemes are possible \cite{GM-Theory-3}. Transductive learning involves making predictions on partially labeled datasets, using the structure of the data to predict the remaining unlabeled instances. Moreover, transductive learning does not have a clear distinction between the training and testing phases, as it leverages both labeled and unlabeled data simultaneously. Inductive learning involves training machine learning models on labeled datasets to make predictions on new unseen data. This approach has been widely applied in tasks such as image classification, object detection, and semantic segmentation \cite{GM-Theory-2}. \\

Although GNNs show excellent performance in handling graph data, the potential higher-order relations among objects will be lost if there are simply represented by a graph. Hypergraph Neural Networks\footnote{https://sites.google.com/view/gsp-website/graph-neural-networks/hypergraph-neural-networks}\cite{GM-Theory-10} is recognized as a flexible modeling tool for complex and higher-order data, including a vertex set and a hyperedge set. Objects are considered vertices, multiple vertices are connected to form a hyperedge representing higher-order relations. Graph representation and learning has been recently employed in infrared small target detection.

\subsection{Application to Infrared Small Target Detection}
In literature, several works use graph representation and learning; the most representative ones are the following:
\begin{enumerate}
\item \textbf{Graph Learning:} In 2024, Lin et al. \cite{DL-291} design a Cycle Shift Vision Graph Convolution Network (CS-ViG-UNet\footnote{https://linaom1214.github.io/CSViG-UNet}) which uses a cyclic shift sparse graph attention mechanism to solve the issue of reduced expressive power. The CS-ViG module constructs an effective graph structure using image patches to capture feature information relevant to target recognition. Experiments on the SIRST AUG and IRSTD-1K show that CS-ViG-UNet provides F1 scores of 0.8561 and 0.745, respectively.  On the RTX3090 with TensorRT acceleration, CS-ViG-UNet process approximately 357 images of size $256 \times 256 pixels$ per second. In 2025, Shen et al. \cite{GM-1} propose a graph-based context learning approach for infrared small target detection. More specifically, a Graph-based Context Learning Network (GCLNet) integrates a global graph reasoning with local feature learning to perceive context information across multiple scales. To this end, local feature learning blocks are embedded into the encoder to extract detailed textures. At deep layers, the multiple graph reasoning module employs a multi-graph interaction structure to encourage  information transfer, allowing for the optimization of global context learning. Then, the patch-based graph reasoning module divides the low-level features into multiple patches where the context information is explored to capture the saliency of small targets. Experiments show that GCLNet outperforms  state-of-the-art methods, achieving the intersection over union (IoU) of $80.26\%$ and $94.84\%$ on the NUAA-SIRST and NUDT-SIRST datasets, respectively. \\
\item \textbf{Hypergraph Learning:} In 2026, Qi et al. \cite{GM-2} design HyperTea which integrates global and local temporal perspectives to  model high-order spatiotemporal correlations of features. HyperTea comprises three modules: the Global Temporal Enhancement Module (GTEM) which realizes global temporal context enhancement via semantic aggregation and propagation, and the Local Temporal Enhancement Module (LTEM) which  capture local motion patterns between adjacent frames. In addition, a Temporal Alignment Module (TAM) tackle potential cross-scale feature misalignment. HyperTea is the first work in which CNNs, RNNs, and hypergraph neural networks (HGNNs) are integrated for MIRSTD. Experiments on DAUB and IRDST datasets show that HyperTea outperforms ISNet \cite{DL-211}, RPCANet and SSTNet \cite{DL-38} .
\end{enumerate}

\section{Performance Evaluation}
\label{RDC}
\begin{figure*}
\begin{center}
\includegraphics[width=10.5cm]{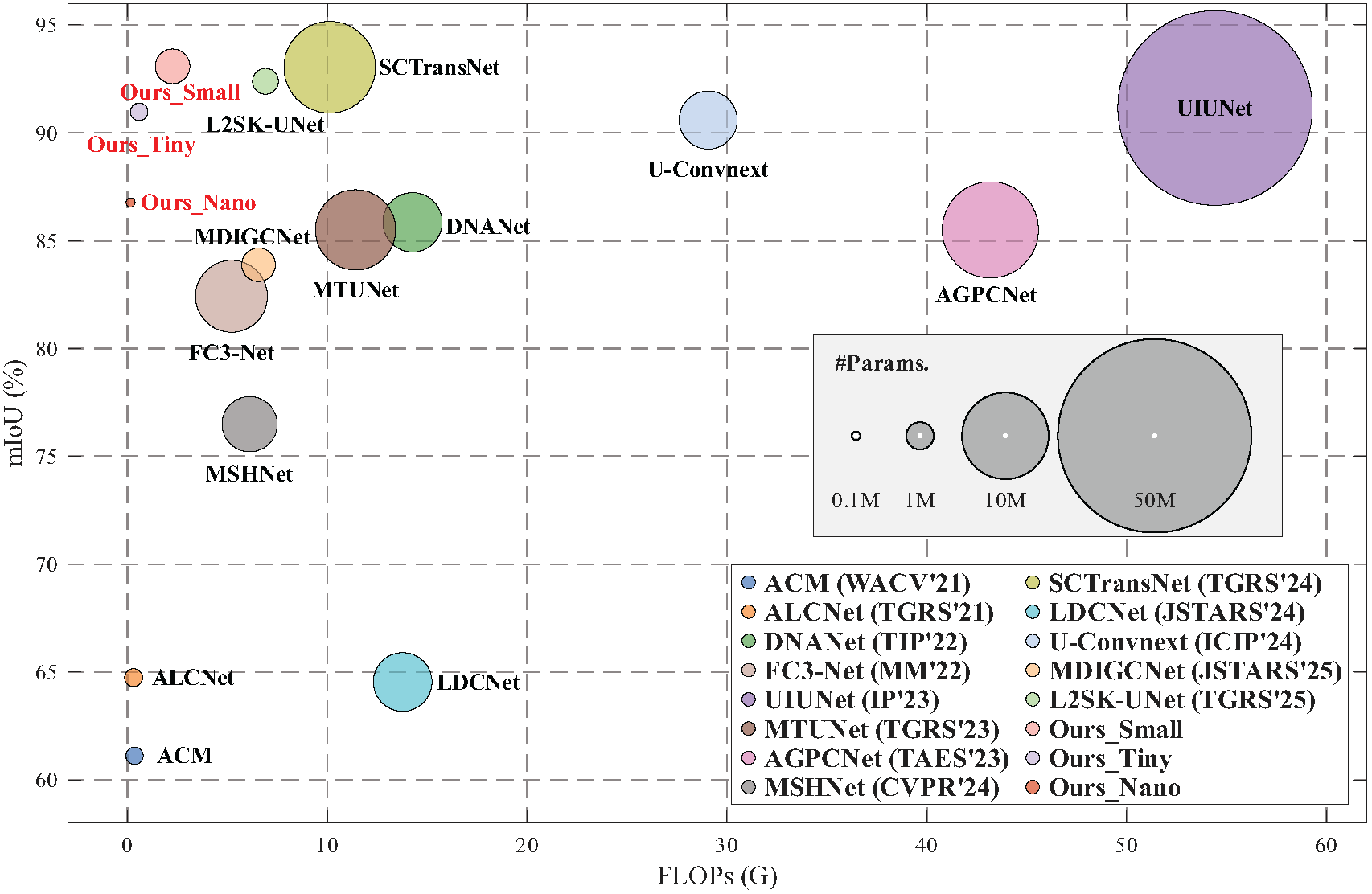}
\caption{Comparison between Performance and the Number of Parameters  for LW-DSTransNet compared to early SOTA Data-driven Algorithms (Image from Huang et al. \cite{DL-599-4}).} 
\label{PE-DM-1}
\end{center}
\end{figure*}

\begin{figure}
\begin{center}
\includegraphics[width=7.5cm]{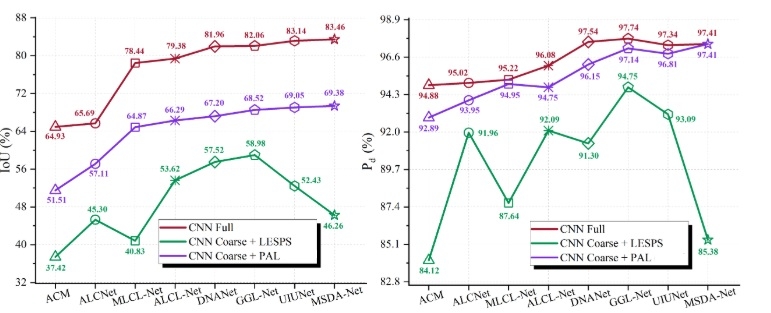}
\caption{Impact of Plug and Play Single Point Supervised Learning Modules on SOTA Data-driven Algorithms (Image from Yu et al. \cite{DL-368}).} 
\label{PE-DM-2}
\end{center}
\end{figure}

\begin{figure}
\begin{center}
\includegraphics[width=7.5cm]{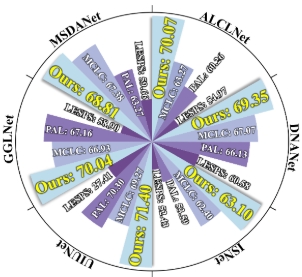}
\caption{Impact of Plug and Play Single Point Supervised Learning Modules on SOTA Data-driven Algorithms (Image from Liu et al. \cite{DL-599-6}).} 
\label{PE-DM-3}
\end{center}
\end{figure}

In complementary of convention protocols reviewed in "Traditional and Recent Approaches for Infrared Small Target Detection : Overview, Challenges and Perspectives (Part 1)", specific protocols are commonly used for deep neural networks supervised models. Indeed, deep neural methods present specific characteristics that need to be evaluated for a fair comparison. 
\begin{itemize}
\item The first one is the compromise between performance and the number of parameters. Figure \ref{PE-DM-1} shows a graphical comparison in which
LW-DSTransNet \cite{DL-599-4} is compared with previous algorithms. 
\item The second one is about the impact of plug and play modules in the performance of the method. For example, Figure \ref{PE-DM-2} and Figure \ref{PE-DM-3} show graphical comparison from  Yu et al. \cite{DL-368}  and Liu et al. \cite{DL-599-6} about the impact of plug and play modules for single point supervised learning modules namely PAL \cite{DL-368}, LESPS \cite{DL-364} and DtoD  (appearing as Ours) \cite{DL-599-6}) on diverse SOTA algorithms, namely ISNet \cite{DL-211}, DNA-Net \cite{DL-12}, UIU-Net \cite{DL-220}, GGLNet and MSDANet.
\end{itemize}

\section{Prospective Research}
We first provide an analysis of the current methods and evaluation protocols. Second, we review the current solved and unsolved challenges. Third, we provide theoretical perspectives for each category of methods.

\subsection{Analysis of Current Methods}
Deep neural network based methods have significantly increased the performance of small target detection by automatically extracting features from large
datasets. Compared to conventional unsupervised methods, they handle complex images better through multilayer feature extraction improving accuracy. However, they present the following weaknesses both in sparse and dense scenes: 
\begin{itemize}
\item They require large annotated datasets increasing data collection costs.
\item In small target detection, critical semantic information may be lost.
\item Their high computational complexity limits their deployability by requiring powerful hardware. 
\end{itemize}
Even if deep learning methods (i.e data-driven methods) excel in many cases, its reliance on data and resources remains a challenge in the presence of unseen data. In contrast, conventional unsupervised methods (i.e model-driven methods) though less effective in complex scenes, offer robustness, interpretability, and efficiency in resource-limited settings. However, both model-driven and data-driven methods face limitations due to the lack of temporal information in single-frame detection, making it difficult to ensure high detection accuracy.  \\

\subsection{Evaluation Protocols}
 About the conventional evaluation protocols, several weaknesses appear to have a fair evaluation and comparison for ranking. First, they rely on fragmented pixel- and target-level specific metrics and thus they fail to provide a comprehensive view of model capabilities. Second, the emphasis on overall performance scores do not allow a precise error analysis required to identify failure modes and improve real-world system performance. Third, the employed dataset are dataset-specific training-testing paradigm and to no allow the understanding of model robustness and generalization across diverse infrared scenarios.

 \subsection{Prospective Solutions}
Solved and unsolved challenges can be investigated either according to single-frame/multi-frame detection and to sparse/dense targets detection and are reviewed in the paper titled "Traditional and Recent Approaches for Infrared Small Target Detection : Overview, Challenges and Perspectives (Part 1)". Here, we present perspectives for deep neural networks supervised models.

\subsubsection{Perspectives for Researcher Solutions}
 Prospective solutions to handle the unsolved challenges can be the use of recent graph-based learning methods requiring less labeled data, different learning strategies and hypergrah methods:
\begin{itemize}
\item \textbf{Graph Learning:} Graph learning is a deep learning method that takes graph data as input. Graph neural networks have demonstrated valuable performance in the close research field of moving object detection in video surveillance in the presence of challenging environments \cite{MOD-Perspectives-1,MOD-Perspectives-3}. 
\item \textbf{Transductive/Inductive Learning:} The existing deep learning based methods employed in infrared small target detection employed a transductive learning scheme. This scheme involves making predictions on partially labeled datasets, using the structure of the data to predict the remaining unlabeled instances. In addition, transductive learning does not have a clear distinction between the training and testing phases, as it leverages both labeled and unlabeled data simultaneously. In contrast, the
inductive learning scheme is a paradigm that involves training machine learning models on labeled datasets to make predictions on new unseen data. This approach has been widely applied in tasks such as image classification, moving object detection \cite{MOD-Perspectives-2}, and semantic segmentation. Thus, inductive learning is more robust in the presence of unseen cases \cite{MOD-Perspectives-10}.  
\item \textbf{Contrastive Graph Learning}:  Self-supervised learning is a key method for training deep learning models when labeled data is scarce or unavailable. While graph machine learning holds great promise across various domains, the design of effective pretext tasks for self-supervised graph representation learning is  challenging. Contrastive learning, a popular approach in graph self-supervised learning, leverages positive and negative pairs to compute a contrastive loss function. Current graph contrastive learning methods often attempt to fully use structural patterns and node similarities and have been applied with success in moving object detection \cite{MOD-Perspectives-4}. In a related (non-graph) direction, Liao and Yin \cite{DL-214} propose in 2026 a Contrastive-aided Distractor Discriminator (CaDD). The aim is to enforce adaptive similarity computation both locally and globally between the real targets and the backgrounds. This approach allows the authors to more precisely discriminate distractors, so as to reduce the false alarm rate. 
\item  \textbf{Hypergraph Learning:} Although GNNs show excellent performance in handling graph data, the potential higher-order relations among objects will be lost if there are simply represented by a graph. Hypergraph Neural Networks recognized as a flexible modeling tool for complex and higher-order data, including a vertex set and a hyperedge set. Objects are considered vertices, multiple vertices are connected to form a hyperedge representing higher-order relations. 
\end{itemize}

\section{Conclusion}
\label{Conclusion}
In this paper, we have presented a review of deep learning methods for infrared small target detection. In this context, we categorized the deep neural networks supervised models. For each categories, the methods have been surveyed in a rationale and historical manner with the corresponding first works, the similar research, their variants/improvements and their last developments. Moreover, we have presented the specific metrics and the ranking protocols employed for performance evaluation. Finally, we provide theoretical perspectives for future research.

\bibliographystyle{IEEEtran}
\bibliography{1-SFD,2-MFD,3-RSTD,4-GraphMethods,5-PSTD,6-RealTime,7-Applications,10-Survey,11-PerformanceEvaluation,12-Datasets,20-BackgroundSubtraction,21-Perspectives,30-Miscellaneous,31-Sensors}

\end{document}